\documentclass{article}
\usepackage[preprint]{NIPS}

\usepackage[utf8]{inputenc} % allow utf-8 input
\usepackage[T1]{fontenc}    % use 8-bit T1 fonts
\usepackage{url}            % simple URL typesetting
\usepackage{booktabs}       % professional-quality tables
\usepackage{amsfonts}       % blackboard math symbols
\usepackage{nicefrac}       % compact symbols for 1/2, etc.
\usepackage{microtype}      % microtypography

\usepackage{graphicx}
\usepackage{tabularx}
\usepackage{booktabs}

\usepackage[many]{tcolorbox}
\usepackage{adjustbox}
\usepackage{dashbox}

\usepackage[framemethod=tikz]{mdframed}
\usepackage{ntheorem}
\usepackage{xcolor}
\usepackage{color,soul}

\newcolumntype{C}{>{\centering\arraybackslash}X}

\usepackage{amsmath}
\usepackage{amssymb}
\usepackage{cuted}
\usepackage[ruled]{algorithm2e}
\usepackage{tabularx}
\usepackage{lipsum}
\usepackage{multirow}
    
\usepackage[inline]{enumitem}

\usepackage{tikz}

\usetikzlibrary{shapes.geometric, arrows,positioning}
\tikzstyle{startstop} = [rectangle, rounded corners, minimum width=2cm, minimum
height=1cm,text centered, draw=black, fill=red!30]
\tikzstyle{process} = [rectangle, minimum width=2cm, minimum height=1cm, text
centered, draw=black, fill=orange!30]
\tikzstyle{decision} = [diamond, minimum width=2cm, minimum height=1cm, text
centered, draw=black, fill=green!30]
\tikzstyle{io} = [rectangle, minimum width=2cm, minimum height=1cm, text
centered, draw=black, fill=blue!30]
\tikzstyle{arrow} = [thick,->,>=stealth]

\usepackage[pagebackref=true, breaklinks=true, colorlinks,
bookmarks=true] {hyperref}

\usepackage{pgfplots}
\pgfplotsset{compat=1.12}

\usepackage{float}
\usepackage{latexsym}
\usepackage{stmaryrd}
\usepackage{mathtools}
\usepackage{marvosym}

\usepackage{overpic}
\providecommand{\newoperator}[3]{%
\newcommand*{#1}{\mathop{#2}#3}}
\newoperator{\argmin}{\mathrm{argmin}}{\limits}
\newoperator{\argmax}{\mathrm{argmax}}{\limits}

\usepackage{pifont}% http://ctan.org/pkg/pifont
\newcommand{\cmark}{\ding{51}}%
\newcommand{\xmark}{\ding{55}}%

\usepackage{booktabs,tabulary,multirow}
\usepackage[caption=false]{subfig}
 
\newcommand{\bd}[1]{\textbf{#1}}
\newcommand{\app}{\raise.17ex\hbox{$\scriptstyle\sim$}}

\newcolumntype{x}[1]{>{\centering\arraybackslash}p{#1pt}}

\newlength\savewidth

\makeatletter\renewcommand\paragraph{\@startsection{paragraph}{4}{\z@}
  {.5em \@plus1ex \@minus.2ex}{-.5em}{\normalfont\normalsize\bfseries}}\makeatother

\usepackage{makecell} % multiple lines in a table cell

\title{A Multi-Level Approach to Waste Object Segmentation}

\author{%
  Tao Wang \\
  Minjiang University\\
  \texttt{twang@mju.edu.cn} \\
  \And
  Yuanzheng Cai\thanks{Corresponding author. Paper appears in \textit{Sensors} \textbf{2020}, \textit{20}(14), 3816. DOI 10.3390/s20143816 } \\
  Minjiang University\\
  \texttt{yuanzheng\_cai@mju.edu.cn} \\
  \And
  Lingyu Liang \\
  South China University of Technology\\
  \texttt{eelyliang@scut.edu.cn} \\
  \And
  Dongyi Ye \\
  Fuzhou University\\
  \texttt{yiedy@fzu.edu.cn} \\
}

\begin{document}

\maketitle

\begin{abstract}
   We address the problem of localizing waste objects from a color image and an optional depth image,
   which is a key perception component for robotic interaction with such objects.
   Specifically, our~method integrates the intensity and depth information at multiple levels of spatial granularity. Firstly,
   a scene-level deep network produces an initial coarse segmentation, based on which we select a few potential object regions
   to zoom in and perform fine segmentation. The~results of the above steps are further integrated into a densely connected conditional
   random field that learns to respect the appearance, depth, and~spatial affinities with pixel-level accuracy.
   In addition, we~create a new RGBD waste object segmentation dataset, MJU-Waste, that~is made public to facilitate future research in this
   area. The~efficacy of our method is validated on both MJU-Waste and the Trash Annotation in Context (TACO) dataset.
\end{abstract}

\section{Introduction}

\begin{figure}[H]
	\begin{center}
		\includegraphics[width=0.9\textwidth]{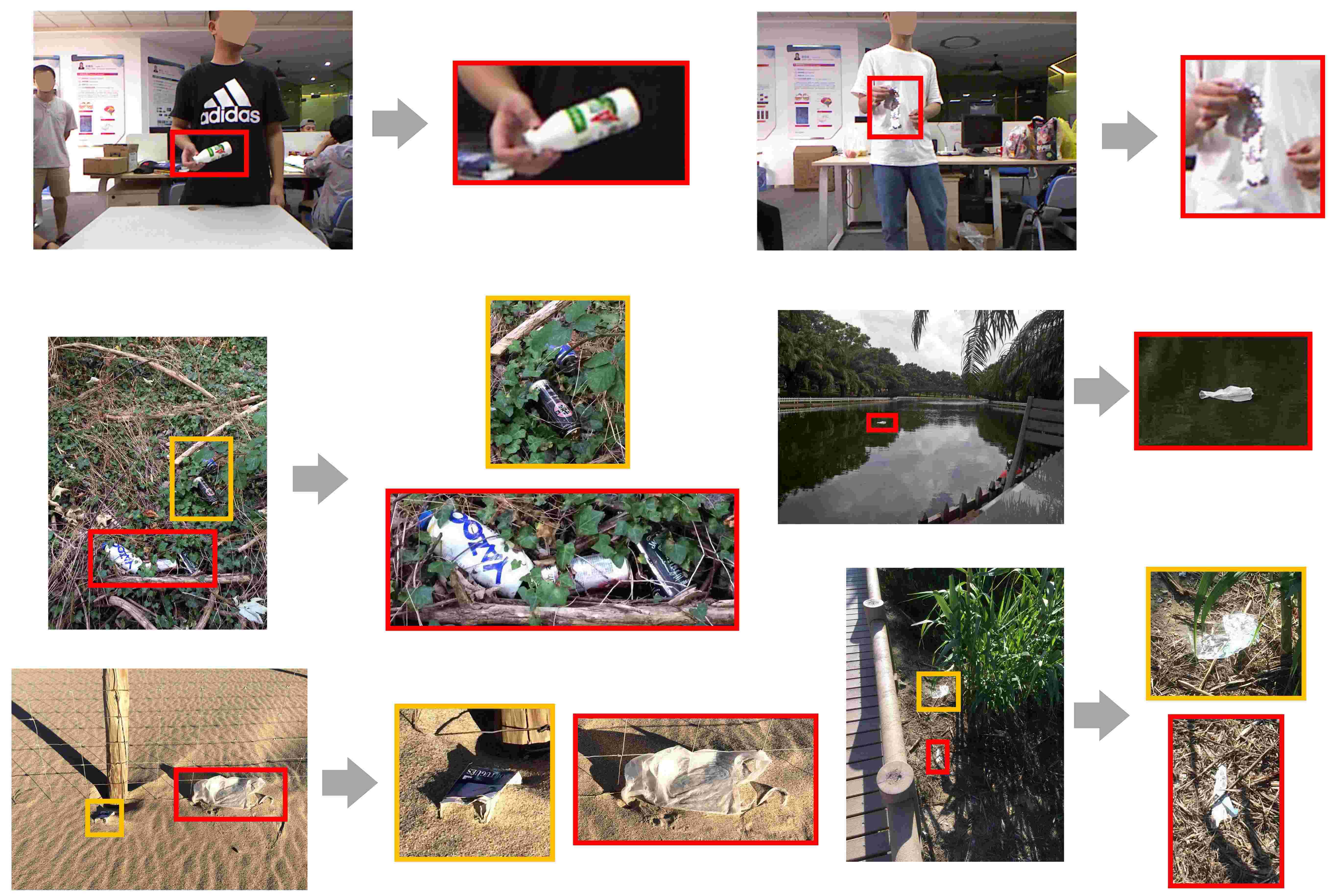}
	\end{center}
	\caption{
			Example images from MJU-Waste and TACO~\cite{proencca2020taco} datasets and their zoomed-in object regions.
			Detecting and localizing waste objects require both scene level and object level reasoning.
			See text for~details.
	}
	\label{fig:motivation}
\end{figure}

Waste objects are commonly found in both indoor and outdoor environments such as household, office or road scenes.
As such, it~is important for a vision-based intelligent robot to localize and interact with them.
However, detecting and segmenting waste objects are much more challenging than most other objects.
For example, waste objects could either be incomplete or damaged, or~both. In~many cases, their presence
could only be inferred from scene-level contexts, e.g., via~reasoning about their contrast to the background and
judging by their intended utilities. On~the other hand, one~key challenge to accurately localizing waste objects is
the extreme scale variation resulting from the variable physical sizes and the dynamic perspectives,
as shown in Figure~\ref{fig:motivation}.
Due to the large number of small objects, it~is difficult even for most humans to
accurately delineate waste object boundaries without zooming in to see the appearance details clearly.
For the human vision system, however, attention can either be shifted to cover a wide area of the visual field,
or narrowed to a tiny region as when we scrutinize a small area for details (e.g.,~\cite{eriksen1986visual,shulman1987spatial,pashler1999psychology,palmer1999vision}).
Presented with an image, we~can immediately recognize the meaning of the scene and the global structure,
which allow us to easily spot objects of interest. We~can consequently attend to those object regions to perform
fine-grained delineation. Inspired by how the human vision system works, we~solve the waste object segmentation
problem in a similar manner by integrating visual cues from multiple levels of spatial granularity.

\begin{figure}[t]
	\begin{center}
		\includegraphics[width=0.98\textwidth]{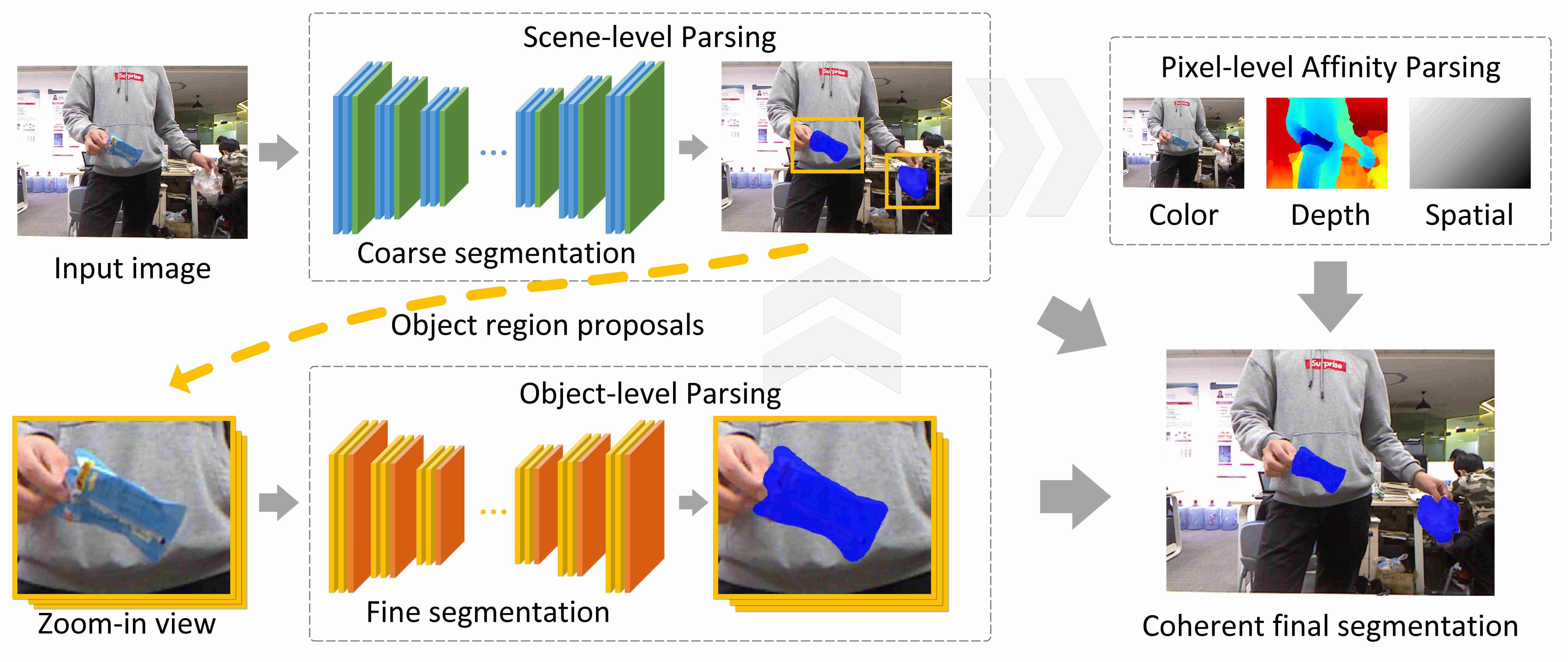}
	\end{center}
	\caption{
			Overview of the proposed method. Given an input image, we~approach the waste object segmentation problem at three levels:
			(i) scene-level parsing for an initial coarse segmentation, (ii)~object-level parsing to recover fine details for each object region
			proposal, and~(iii) pixel-level refinement based on color, depth, and~spatial affinities. Together, joint inference at all
			these levels produces coherent final segmentation results.
	}
	\label{fig:hilevel}
\end{figure}

The general idea of exploiting objectness has long been proven effective for a wide range of vision-based
applications~\cite{alexe2012measuring,uijlings2013selective,zitnick2014edge,cheng2014bing}.
In particular, several works have already demonstrated that objectness reasoning can positively impact semantic
segmentation~\cite{wang2016objectness,xia2016zoom,alexe2010classcut,vezhnevets2011weakly}.
However, in~this work we propose a simple yet effective strategy for waste object
proposal that neither require pretrained objectness models nor additional object or part annotations.
Our primary goal is to address the extreme scale variation which is much less common in generic objects.
In order to obtain accurate and coherent segmentation results, our~method performs joint inference at three levels.
Firstly, we~obtain a coarse segmentation at the scene-level to capture the global context and to propose potential object regions.
We note that our simple object region proposal strategy captures the objectness priors reasonably well in practice.
This is followed by an object-level segmentation to recover fine structural details for each object region proposal.
In particular, adopting two separate models at the scene and object levels respectively allows us to disentangle the learning of
the global image contexts from the learning of the fine object boundary details.
Finally, we~perform joint inference to integrate results from both the scene and object levels, as~well as making
pixel-level refinements based on color, depth, and~spatial affinities.
The main steps are summarized and illustrated in Figure~\ref{fig:hilevel}.
We obtain significantly superior results
with our method, greatly surpassing a number of strong semantic segmentation baselines.

Recent years witnessed a huge success of deep learning in a wide spectrum of vision-based perception tasks~\cite{he2016deep,zhang2020resnest,ren2015faster,long2015fully}.
In this work, we~would also like to harness the powerful learning capabilities of convolutional neural network (CNN) models to address
the waste object segmentation problem. Most~state-of-the-art CNN-based segmentation models exploit the spatial-preserving properties of
fully convolutional networks~\cite{long2015fully} to directly learn feature representations that could translate into class probability maps
either at the scene level (i.e., semantic segmentation) or the object level (i.e.,~instance segmentation). One~of the key limitations
when it comes to applying these general-purpose models directly for waste object segmentation is that they are unable to handle
the extreme object scale variation due to the delicate contention between global semantics and accurate localization
under a fixed feature resolution, and~the resulting segmentation can be inaccurate for the abundant small objects with complex
shape details.
Based upon this observation, we~propose to learn a multi-level model that allows us to adaptively zoom into
object regions to recover fine structural details, while retaining a scene-level model to capture the long-range context
and to provide object proposals.
Furthermore, such~a layered model can be jointly reasoned with pixel-level refinements
under a unified Conditional Random Field (CRF)~\cite{koller2009probabilistic} model.

The main contributions of our work are three-fold. Firstly, we~propose a deep-learning based waste object segmentation
framework that integrates scene-level and object-level reasoning.
In particular, our~method does not require additional object-level annotations. By~virtue of a simple
object region proposal method, we~are able to
learn separate scene-level and object-level segmentation models that allow us to achieve
accurate localization while preserving the strong global contextual semantics.
Secondly, we~develop a strategy based on densely connected CRF~\cite{krahenbuhl2011efficient} to perform joint inference at the scene, object,
and pixel levels to produce a highly accurate and coherent final segmentation. In~addition to the scene and object level parsing, our
CRF model further refines the segmentation results with appearance, depth, and~spatial affinity pairwise terms.
Importantly, this~CRF model is also amenable to a filtering-based efficient inference.
Finally, we~collected and annotated a new {RGBD}~\cite{han2013enhanced} dataset, MJU-Waste, for~waste object segmentation.
We believe our dataset is the first public RGBD dataset for this task.
Furthermore, we~evaluate our method on the TACO dataset~\cite{proencca2020taco}, which is another public
waste object segmentation benchmark.
To the best of our knowledge, our~work is among the first in the literature to address waste object segmentation on public datasets.
Experiments on both datasets verify that our method can be used as a general framework to improve the performance of a wide range of
deep models such as FCN~\cite{long2015fully}, PSPNet~\cite{zhao2017pyramid}, CCNet~\cite{huang2019ccnet} and DeepLab~\cite{chen2017rethinking}.

We note that the focus of this work is to obtain accurate waste object boundary delineation.
Another closely related and also very challenging task is waste object detection and classification. Ultimately, we~would like to solve for
waste instance segmentation with fine-grained class information. However, existing datasets do not provide a large number of object classes with sufficient training data.
In addition, differentiating waste instances under a single class label is also challenging.
For~example, the~best Average Precision (AP) obtained in~\cite{proencca2020taco} are in the $20$s for the TACO-1 classless litter detection task where the goal is to
detect and segment litter items with a single class label.
Therefore, in~this paper we adopt a research methodology under which we gradually move
toward richer models while maintaining a high level of performance. In~this regard, we~formulate our problem as a
two-class (waste vs. background) semantic segmentation one.
This allows us to obtain high quality segmentation results as we demonstrate with our experiments.

In the remainder of this paper, Section~\ref{sec:related} briefly reviews the literature on waste object segmentation and
related tasks, as~well as recent progress in semantic segmentation.
We then describe details of our method in Section~\ref{sec:approach}.
Afterwards, Section~\ref{sec:experiments} presents findings from our experimental evaluation, followed by closing remarks in Section~\ref{sec:conclusion}.

\section{Related Work}
\label{sec:related}
\vspace{-3pt}
\subsection{Waste Object Segmentation and Related Tasks}

The ability to automatically detect, localize and classify waste objects is of wide interest in the computer and robotic vision community.
However, there are relatively limited works in the literature that address the specific task of waste object segmentation.
We believe this is partially due to the poor availability of public waste segmentation datasets
until very recently.
Therefore, in~this paper we propose the MJU-Waste dataset with $2475$ RGBD images each annotated with a pixelwise waste object mask.
To facilitate future research, we~make our dataset publicly available.
To the best of our knowledge, this~is the only public dataset of this kind in addition to the
TACO dataset~\cite{proencca2020taco} of $1500$ color images.
Below we briefly review some recent works on waste object classification, detection and segmentation which are closely related ours.

Yang and Thung~\cite{yang2016classification} addressed the waste classification problem and compared the performance of shallow and deep models.
In their work, they~collected the TrashNet dataset of $2500$ images of single pieces of waste.
Based on their data, Bircano{\u{g}}lu et~al.~\cite{bircanouglu2018recyclenet} and Aral et~al.~\cite{aral2018classification} performed detailed
comparisons among various deep architectures.
Additionally, Awe~et~al.~\cite{awe2017smart} created a synthetic dataset for waste object detection based on Faster RCNN~\cite{ren2015faster}.
Similarly, Chu~et~al.~\cite{chu2018multilayer} proposed a hybrid CNN approach for waste classification with a dataset of $5000$ waste objects.
Vo et~al.~\cite{vo2019novel} created another dataset VN-trash with $5904$ images for deep transfer learning.
Furthermore, Ramalingam et~al.~\cite{ramalingam2018cascaded} presented a debris classification model for floor-cleaning robots with a cascade CNN and an SVM.
Yin et~al.~\cite{yin2020table} proposed a lightweight CNN for food litter detection in table cleaning tasks.
Rad~et~al.~\cite{rad2017computer} presented an approach similar to OverFeat~\cite{sermanet2013overfeat} for litter object detection.
Another similar approach based on Faster RCNN~\cite{ren2015faster} is presented by Wang and Zhang~\cite{wang2018autonomous}.
Contrary to the above works, we~address the waste object segmentation problem that requires accurate delineation of object~boundaries.

In terms of methods that involves a segmentation component,
Bai et~al.~\cite{bai2018deep} designed a robot for picking up garbage on the grass with a two-stage perception approach.
Firstly, they~used SegNet~\cite{badrinarayanan2017segnet} for ground segmentation to allow the robot to move toward waste objects.
After a close-range image is acquired, ResNet~\cite{he2016deep} is used for object classification.
Here the segmentation module is used for background modeling of the grassland only, hence no object segmentation is performed.
Deepa et~al.~\cite{deepa2017estimation} presented a garbage coverage segmentation method in water terrain based on color transformation and K-means.
In addition, Mittal et~al.~\cite{mittal2016spotgarbage} proposed an approach based on the Fully Convolutional Network (FCN)~\cite{long2015fully} for coarse garbage segmentation.
Their method is based on extracting image patches and combining their predictions, and~therefore cannot capture the finer object boundary details.
Zeng et~al.~\cite{zeng2019multi} proposed a multi-scale CNN based garbage detection method from airborne hyperspectral data. In~their method,
a binary segmentation map is generated as the input to selective search~\cite{uijlings2013selective} for the purpose of obtaining bounding box-based region proposals.
All~these above works do not address the specific task of waste object segmentation for robotic interaction.

Perhaps being the closest to our work, Zhang et~al.~\cite{qiu2017three} proposed an object segmentation method for waste disposal lines based on RGBD sensors.
Their method begins with background subtraction on the 3D point cloud, and~then attempts to find
an optimal projection plane for subsequent object segmentation.
Another work~\cite{wang2017rgb} from the same group proposed a relabeling method for ambiguous regions after the background is subtracted.
Unlike their methods, we~take a data-driven approach to address the problem in a much more challenging scenario.
Specifically, our~method does not assume a particular background model and is able to segment waste objects
in both hand-held and in-the-wild~scenarios.

Another work that is conceptually similar to ours is from Grard et~al.~\cite{grard2019object}. They~explored an interesting interactive setting for
object segmentation from a cluttered background. A~user is asked to click on an object to extract, and~their model uses a dual-objective FCN
trained on synthetic depth images to produce the object mask. In~our work, however, we~aim at a more challenging scenario that does not require
human interaction.
\vspace{-3pt}
\subsection{Semantic Segmentation}

The task of assigning a class label to every pixel in an image is a long-established
fundamental problem in computer vision~\cite{he2004multiscale,shotton2006textonboost,ladicky2009associative,gould2009decomposing,kumar2010efficiently,munoz2010stacked,tighe2010superparsing,liu2011nonparametric,lempitsky2011pylon}.
Since the pioneering work of Long, Shelhamer and Darrell~\cite{long2015fully},
researchers proposed a large number of architectures and techniques to improve upon the 
FCN to capture multi-scale context~\cite{liu2015parsenet,mostajabi2015feedforward,yu2015multi,lin2016efficient,zhang2018context,ding2018context,lin2018multi} or to improve results at object
boundaries~\cite{zheng2015conditional,schwing2015fully,liu2015semantic,jampani2016learning,chandra2016fast,pohlen2017full,gadde2016superpixel,li2017foveanet}.
Other recent works have explored encoder-decoder structures~\cite{noh2015learning,ronneberger2015u,badrinarayanan2017segnet} and
contextual dependencies based on the self-attention mechanism~\cite{yuan2018ocnet,zhao2018psanet,huang2019ccnet,fu2019dual}.
For example, some~of the recent influential works include PSPNet~\cite{zhao2017pyramid} which proposed the pyramid pooling module to combine representations from multiple scales
and RefineNet~\cite{lin2017refinenet} in which a multi-path refinement network is proposed for high-resolution semantic segmentation.
Furthermore, Gated SCNN~\cite{takikawa2019gated} proposed a two-stream structure that explicitly enhances shape prediction.
We note that the performance of semantic segmentation algorithms is clearly related to the backbone architecture being used.
For example, the~recently proposed ResNeSt~\cite{zhang2020resnest} model based on Split-Attention blocks provided large performance improvements
to a number of vision tasks including semantic segmentation.

We note that our work is similar to DeepLab~\cite{chen2017deeplab} whose main contributions include the atrous spatial pyramid pooling (ASPP)
module to capture the multi-scale context and the use of dense CRF~\cite{krahenbuhl2011efficient} to improve results at object boundaries.
In their follow up work~\cite{chen2017rethinking,chen2018encoder},
they improved the ASPP module by adding image pooling and a decoder structure.
Our method differs from the DeepLab series in two important aspects. Firstly, our~method is tailored to the task of waste object segmentation
and introduces layered deep models that perform scene-level parsing and object-level parsing respectively. Secondly, the
dense CRF model in our work integrates information from both the scene and object level parsing results, as~well as
pixel-level affinities which can additionally encode local geometric information via input depth images. In~fact, we~demonstrate through our experiments that
the proposed method is a general framework which can be applied in conjunction with DeepLab and other strong semantic segmentation baselines
such as PSPNet~\cite{zhao2017pyramid} and CCNet~\cite{huang2019ccnet} to improve their results by a clear margin on the waste object segmentation task.

There have been a few works that explored semantic segmentation with RGBD data~\cite{gupta2014learning,li2016lstm,eigen2015predicting,hu2019acnet}.
For~example, FuseNet~\cite{hazirbas2016fusenet} proposed a two-stream encoder that extracts features from both color and depth images in an encoder-decoder
type of network. Qi~et~al.~\cite{qi20173d} constructed a graph neural network based on spatial affinities inferred from depth.
Contrary to existing works, we~use depth affinities as a means to refining waste object segmentation results. Our~use of depth information is flexible in that
the model can cope with situations where the depth modality is present or absent without re-training.

Lastly, there have been a few works that tackle objectness aware semantic segmentation~\cite{wang2016objectness,xia2016zoom}.
Apart from addressing the problem in the novel waste object segmentation domain, our~method uses a simple yet effective strategy
for object region proposal that does not require any additional object or part annotations.

\section{Our Approach}
\label{sec:approach}

In this section, let~us formally introduce the waste object segmentation problem
and the proposed approach. We~begin with the definition of the problem and notations.
Given an input color image and optionally an additional depth image, our~model outputs a pixelwise labeling map,
as shown in Figure~\ref{fig:hilevel}.
Mathematically, denote the input color image as $\mathbf{I} \in \mathbb{R}^{H \times W \times 3}$, the~optional
depth image as $\mathbf{D} \in \mathbb{R}^{H \times W}$, and~the semantic label set as $\mathcal{C}=\{1,2,\dots,C\}$,
{where $C$ is the number of classes.}
Our goal is to produce a structured semantic labeling $\mathbf{x} \in \mathcal{C}^{H \times W}$.  We note that in deep models,
the labeling of $\mathbf{x}$ at image coordinate $(i,j)$, $x_{\textit{ij}}$, is~usually obtained via multi-class softmax scores on a spatial-preserving
convolutional feature map $\mathbf{C} \in \mathbb{R}^{H \times W \times C}$, i.e.,
$x_{\textit{ij}} = \argmax_{k'} \frac{\exp(\mathbf{C}^{i,j,k'})}{\sum_{k=1}^{C}\exp(\mathbf{C}^{i,j,k})}$.
In practice, it~is common that the convolutional feature map $\mathbf{C}$ is downsampled w.r.t. the~original
image resolution, but~we can always assume that the resolution can be restored with interpolation.

\subsection{Layered Deep Models}

\begin{figure}[t!]
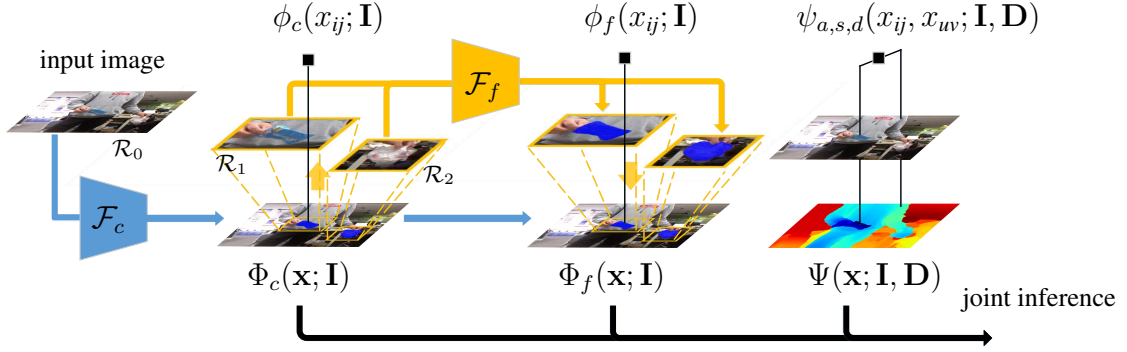

	\begin{overpic}[scale=.051]%
		{images/model.pdf}
		\put(9,12.5){ \large{$ \mathcal{F}_c $} }
		\put(45,25.5){ \large{$ \mathcal{F}_f $} }
		\put(26.5,32){ \large{$ \phi_c(x_{\textit{ij}};\mathbf{I}) $} }
		\put(56.5,32){ \large{$ \phi_f(x_{\textit{ij}};\mathbf{I}) $} }
		\put(77,32){ \large{$ \psi_{a,s,d}(x_{\textit{ij}},x_{\textit{uv}};\mathbf{I,D}) $} }
		\put(24,7){ \large{$ \Phi_c({\mathbf{x}};\mathbf{I}) $} }
		\put(54,7){ \large{$ \Phi_f({\mathbf{x}};\mathbf{I}) $} }
		\put(78,7){ \large{$ \Psi({\mathbf{x}};\mathbf{I,D}) $} }
		\put(11,19.5){ \footnotesize{$ \mathcal{R}_0 $} }
		\put(21,18){ \footnotesize{$ \mathcal{R}_1 $} }
		\put(41,17){ \footnotesize{$ \mathcal{R}_2 $} }
		\put(4,28){ input image }		
		\put(93,5){ joint inference }
	\end{overpic}
		%\vspace{1mm}
		\caption
         {
			The graphical representation of our CRF. $\mathcal{F}_c$ and $\mathcal{F}_f$
			represent the feature embedding functions for the coarse and the fine
			segmentation networks. Our~model consists of the scene-level unary term
			$ \Phi_c({\mathbf{x}};\mathbf{I}) $, the~object-level unary term
			$ \Phi_f({\mathbf{x}};\mathbf{I}) $, and~the pixel-level pairwise term
			$ \Psi({\mathbf{x}};\mathbf{I,D}) $.
         }
      \label{fig:model}
\end{figure}

\textls[-12]{In this work, we~apply deep models at both the scene and the object levels. For~this purpose, let~us define a number of image regions in which we obtain deeply trained feature representations.
Firstly, let~$\mathcal{R}_0 = \{(i,j)_{i \in \{1 \dots H\}, j~\in \{1 \dots W\}}\}$ be the set of all
spatial coordinates on the image plane, or~the entire image region. This~is the region in which we perform
scene-level parsing (i.e., coarse segmentation). In~addition, we~perform object-level parsing (i.e., fine
segmentation) on a set of non-overlapping object region proposals. We~denote each of these additional regions
as \mbox{$\mathcal{R}_l=\{(i,j)_{i \in \{1 \dots H_l\}, j~\in \{1 \dots W_l\}}\}, 1 \leq l \leq L$}.
Details on generating these regions are discussed in Section~\ref{sec:proposals}.
We apply our coarse segmentation feature embedding network $\mathcal{F}_c$ and the fine segmentation feature
embedding network $\mathcal{F}_f$ to the appropriate image regions as follows:}
\begin{align}
	\mathbf{C}_0 &= \mathcal{F}_c(\mathcal{R}_0 \Rightarrow \mathbf{I}), \notag \\
	\mathbf{C}_l &= \mathcal{F}_f(\mathcal{R}_l \Rightarrow \mathbf{I}), 1 \leq l \leq L
\end{align}

\noindent where $\mathcal{R}_l \Rightarrow \mathbf{I}$ denotes cropping the region $\mathcal{R}_l$ from image $\mathbf{I}$.
Here $\mathbf{C_0} \in \mathbb{R}^{H \times W \times C}$ and $\mathbf{C}_l \in \mathbb{R}^{H_l \times W_l \times C}$, and
we note that these feature maps are upsampled where necessary.
{In addition, the~spatial dimension may be image and region specific for both $\mathcal{R}_0$ and $\mathcal{R}_l$, which poses a practical problem
for batch-based training. To~address this issue, during CNN training we~resize all image regions so that they have a common shorter side length, followed by
randomly cropping a fixed-sized patch as part of the data augmentation procedure.
We refer the readers to Section}~\ref{sec:impl} {for details.}
In Figure~\ref{fig:model}, the~processes shown in blue and yellow illustrate the steps described in this section.

\subsection{Coherent Segmentation with CRF}

Given the layered deep models, we~now introduce our graphical model for predicting coherent
waste object segmentation results. Specifically, the~overall energy function of our CRF model consists of three main
components:

\begin{align}
	E(\mathbf{x}, \mathbf{I}, \mathbf{D}) = \Phi_c(\mathbf{x}; \mathbf{I}) + \alpha \cdot \Phi_f(\mathbf{x}; \mathbf{I}) + \Psi(\mathbf{x}; \mathbf{I}, \mathbf{D})
	\label{eqn:overall}
\end{align}

\noindent where $\Phi_c(\mathbf{x}; \mathbf{I})$ represents the scene-level coarse segmentation potentials,
$\Phi_f(\mathbf{x}; \mathbf{I})$ denotes the object-level fine segmentation potentials, and
$\Psi(\mathbf{x}; \mathbf{I}, \mathbf{D})$ is the pairwise potentials that respect the color, depth, and~spatial
affinities in the input images.
$\alpha$ is the weight for the relative importance among the two unary terms.
The graphical representation of our CRF model is shown in Figure~\ref{fig:model}.
We describe the details of these three terms below.

\noindent \textbf{Scene-level unary term.} The scene-level coarse segmentation unary term is given
by $\Phi_c(\mathbf{x};\mathbf{I}) = \sum_{(i,j) \in \mathcal{R}_0}\phi_c(x_{\textit{ij}};\mathbf{I}) = \sum_{(i,j) \in \mathcal{R}_0}-\log(P_c(x_{\textit{ij}};\mathbf{I}))$ where
$P_c(x_{\textit{ij}};\mathbf{I})$ is a pixelwise softmax on the feature map $\mathbf{C}_0$ as follows:
\begin{align}
	P_c(x_{\textit{ij}};\mathbf{I})=\frac{\exp(\mathbf{C}_0^{i,j,k'})}{\sum_{k=1}^{C}\exp({\mathbf{C}_0^{i,j,k}})} \llbracket x_{\textit{ij}}=k' \rrbracket
	\label{eqn:unary_coarse}
\end{align}

\noindent where $\llbracket \cdot \rrbracket$ denotes the indicator function. This~term produces a coarse segmentation map
based on the long-range contexts from the input image. Importantly, we~use the output of this term to generate our object
region proposals, as~discussed in Section~\ref{sec:proposals}.

\noindent \textbf{Object-level unary term.} The object-level fine segmentation unary term is given
by $\Phi_f(\mathbf{x}; \mathbf{I}) = \sum_{(i,j) \in \mathcal{R}_0}\phi_f(x_{\textit{ij}};\mathbf{I})$ where
$\phi_f(x_{\textit{ij}};\mathbf{I})$ is defined as:
\begin{align}
	\phi_f(x_{\textit{ij}};I)=
	\begin{cases}
		-\log(P_f(x_\textit{ij},l;\mathbf{I})), & \text{if $ (i,j) \in \mathcal{R}_l, 1 \leq l \leq L$,}
        \\
        \phi_c(x_{\textit{ij}};\mathbf{I}), & \text{otherwise}
	\end{cases}
\end{align}

\noindent The formulation above states that if a pixel location $(i,j)$ belongs to one of the $L$ object region
proposals, a~negative log-probability obtained via fine segmentation is adopted. Otherwise, the~object-level
unary term falls back to the scene-level unary potentials. Here~the probability $P_f(x_\textit{ij},l;\mathbf{I})$
given by the fine segmentation model is obtained as follows:
\begin{align}
	P_f(x_{\textit{ij}},l;\mathbf{I})=\frac{\exp(\mathbf{C}_l^{\mathcal{T}^1_l(i),\mathcal{T}^2_l(j),k'})}{\sum_{k=1}^{C}\exp({\mathbf{C}_l^{\mathcal{T}^1_l(i),\mathcal{T}^2_l(j),k}})} \llbracket x_{\textit{ij}}=k' \rrbracket
	\label{eqn:unary_fine}
\end{align}

\noindent where $\mathcal{T}^1_l(\cdot)$ and $\mathcal{T}^2_l(\cdot)$ are translation functions that map the image coordinates to that of
the $l$-th object proposal region, and~$\mathbf{C}_l$ is the output feature embedding from the fine segmentation model
for the $l$-th object proposal region.
We note that the object-level unary potentials typically recover more fine
details along object boundaries, as~opposed to the scene-level unary potentials. In~general, it~would become
too computationally expensive to compute scene-level potentials at a comparable resolution for the entire image.
In most cases, computing the object-level unary term on less than $3$ object region proposals are sufficient,
see Section~\ref{sec:proposals} for details.
Additionally, our~object-level potentials are
obtained via a separate deep model that allows us to decouple the learning of long-range contexts
from the learning of fine structural details.

\noindent \textbf{Pixel-level pairwise term.} Although the object-level unary potentials provide finer
segmentation details, accurate boundary details could still be lost for some irregularly shaped waste objects.
This~poses a practical
challenge for detail-preserving global inference.

Following~\cite{krahenbuhl2011efficient}, we~address this challenge by introducing a pairwise term that is a linear combination of Gaussian
kernels in a joint feature space that includes color, depth, and~spatial coordinates. This~allows us to produce coherent object
segmentation results that respect the appearance, depth, and~spatial affinities in the original image resolution.
More importantly, this~form of the pairwise term allows for efficient global inference~\cite{krahenbuhl2011efficient}.
Specifically, our~pairwise term $\Psi(\mathbf{x}; \mathbf{I}, \mathbf{D})$ includes
an appearance term $\psi_a(x_{\textit{ij}},x_{\textit{uv}}; \mathbf{I})$,
a spatial smoothing term $\psi_s(x_{\textit{ij}},x_{\textit{uv}})$ and a depth term $\psi_d(x_{\textit{ij}},x_{\textit{uv}}; \mathbf{D})$:
\begin{align}
	\Psi(\mathbf{x}; \mathbf{I}, \mathbf{D}) = &\sum_{\mathclap{\substack{(i,j) \in \mathcal{R}_0\\ 
													 (u,v) \in \mathcal{R}_0}}}
													 \llbracket x_{\textit{ij}} \neq x_{\textit{uv}} \rrbracket \Big[\psi_a(x_{\textit{ij}},x_{\textit{uv}};\mathbf{I})
													 +\psi_s(x_{\textit{ij}},x_{\textit{uv}}) + \psi_d(x_{\textit{ij}},x_{\textit{uv}},\mathbf{D})\Big]
\end{align}

\noindent where $\llbracket x_{\textit{ij}} \neq x_{\textit{uv}} \rrbracket$ is the Potts label compatibility function.
The appearance term and the smoothing term follow~\cite{krahenbuhl2011efficient} and take the following form:
\begin{align}
\psi_a(x_{\textit{ij}},x_{\textit{uv}}; \mathbf{I}) = w^{(a)}\exp\Big(-\frac{|p_{\textit{ij}}-p_{\textit{uv}}|^2}{2\theta_\alpha^2}-\frac{|I_{\textit{ij}}-I_{\textit{uv}}|^2}{2\theta_\beta^2}\Big)
\end{align}
\begin{align}
\psi_s(x_{\textit{ij}},x_{\textit{uv}}) = w^{(s)}\exp\Big(-\frac{|p_{\textit{ij}}-p_{\textit{uv}}|^2}{2\theta_\gamma^2}\Big)
\end{align}

\noindent where $I_{\textit{ij}}$ and $p_{\textit{ij}}$ are the image appearance and position features at the pixel location $(i,j)$.
In addition, when~a input depth image $\mathbf{D}$ is available, we~are able to enforce an additional pairwise term induced by
geometric affinities:
\begin{align}
\psi_d(x_{\textit{ij}},x_{\textit{uv}}; \mathbf{D}) = w^{(d)}\exp\Big(-\frac{|p_{\textit{ij}}-p_{\textit{uv}}|^2}{2\theta_\delta^2}-\frac{|D_{\textit{ij}}-D_{\textit{uv}}|^2}{2\theta_\epsilon^2}\Big)
\label{eqn:depth-pairwise}
\end{align}

\noindent where $D_{\textit{ij}}$ is the depth reading at the pixel location $(i,j)$.
We note that in practice, any~missing values in $\mathbf{D}$ are filled in with a median filter~\cite{lai2011large} beforehand,
see Section~\ref{sec:mjudata} for details.
{In addition, Equation}~(\ref{eqn:depth-pairwise}) {can be conveniently added or removed depending on
the depth data availability.}
{We~simply discard Equation}~(\ref{eqn:depth-pairwise}) {when training
models for the TACO dataset which only contains color images.}

\subsection{Generating Object Region Proposals}
\label{sec:proposals}

\begin{figure}[H]
	\begin{center}
		\includegraphics[width=0.8\textwidth]{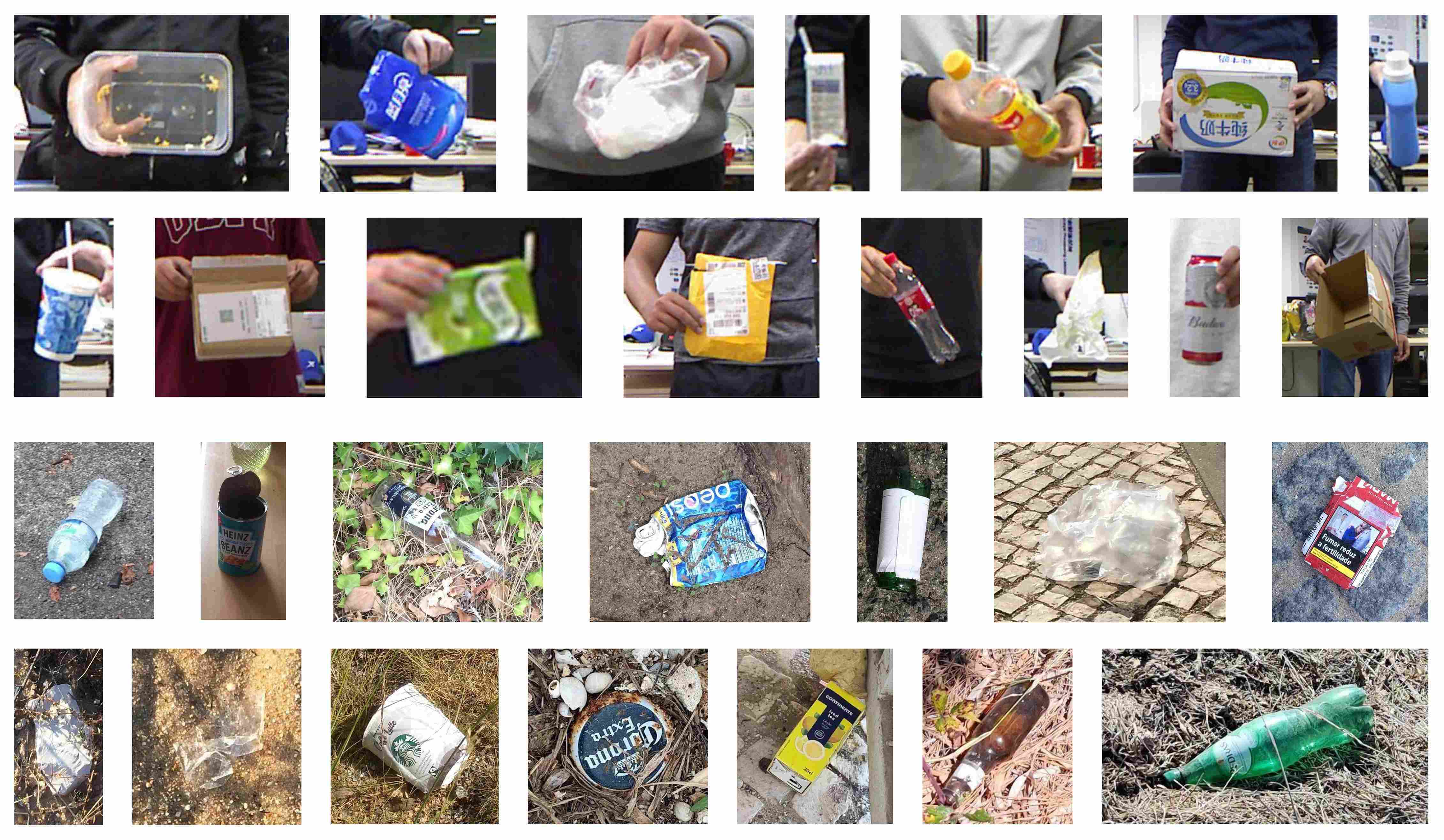}
	\end{center}
	\caption{
			Example object region proposals. The~first two rows show the object region proposals
			from the MJU-Waste dataset. The~remaining two rows show the object region proposals
			from the TACO~dataset.
	}
	\label{fig:proposalex}
\end{figure}

In this work, we~follow a simple strategy to generate object region proposals
$\mathcal{R}_l=\{(i,j)_{i \in \{1 \dots H_l\}, j~\in \{1 \dots W_l\}}\}, 1 \leq l \leq L$.
In particular, the~output from the scene-level coarse segmentation model is a good
indication of the waste object locations. See~Figure~\ref{fig:model} for an example.
We begin with extracting the connected components
in the foreground class labelings of $\mathbf{x} \in \mathcal{C}^{H \times W}$ from the 
maximum a posterior (MAP) estimate of the scene-level unary term
$\Phi_c(\mathbf{x}; \mathbf{I})$. For~each connected component, a~tight bounding box
$\mathcal{R}_l^t$ is extracted. This~is followed by extending $\mathcal{R}_l^t$ by $30\%$
in four directions (i.e., N,S,W,E), subject to the image boundary truncation. Finally, we~merge
overlapping regions and remove those below or above certain size thresholds (details in
Section~\ref{sec:impl}) to obtain a
concise set of final object region proposals $\mathcal{R}_l, 1 \leq l \leq L$.
Example object region proposals obtained using this procedure are shown in Figure~\ref{fig:proposalex},
and we note that any similar implementation should also work~satisfactorily.

Most images from the MJU-Waste dataset contain only one hand-held waste object per image.
For~DeepLabv3 with a ResNet-50 backbone, for~example, only~$2.4\%$ of all images from MJU-Waste
produce $2$ or more object proposals. For~the TACO dataset, $24.0\%$ of all images produce
$2$ or more object proposals. However, only~$8.1\%$ and $0.6\%$ of all images produce more than
$3$ and $5$ object proposals,~respectively.

\subsection{Model Inference}

Following~\cite{krahenbuhl2011efficient}, we~use the mean field approximation of the
joint probability distribution $P(\mathbf{x})$ that computes a
factorized distribution $Q(\mathbf{x})$ which minimizes the
{KL-divergence}~\cite{kullback1951information,bishop2006pattern} $\mathit{KL}(Q||P)$.
For our model, this~yields the following message passing-based iterative update equation:
\begin{align}
Q_{\textit{ij}}(x_{\textit{ij}}=k)=&\frac{1}{Z_\textit{ij}}\exp\Big\{
-\phi_c(x_{\textit{ij}})-\alpha \cdot \phi_f(x_{\textit{ij}})\notag \\
&-\sum_{k' \in \mathcal{C}}\llbracket k,k' \rrbracket
\sum_{\textit{uv} \neq \textit{ij}}\left[ \psi_a(x_{\textit{ij}},x_{\textit{uv}}) + \psi_s(x_{\textit{ij}},x_{\textit{uv}})
+ \psi_d(x_{\textit{ij}},x_{\textit{uv}}) \right] Q_{\textit{uv}}(k')
\Big\}
\end{align}

\noindent where the input color image $\mathbf{I}$ and the depth image $\mathbf{D}$ are omitted for notation simplicity.
In practice, we~use the efficient message passing algorithm proposed in~\cite{krahenbuhl2011efficient}.
The number of iterations is set to $10$ in all experiments.

\subsection{Model Learning}

Let us now move on to discuss details pertaining to the learning of our model. Specifically, we~learn the
parameters of our model by piecewise training. First, the~coarse segmentation feature embedding network
$\mathcal{F}_c$ is trained with standard cross-entropy (CE) loss on the predicted coarse segmentation.
Based on the coarse segmentation for the training images, we~extract object region proposals with the
method discussed in Section~\ref{sec:proposals}. This~allows us to then train the fine segmentation
feature embedding network $\mathcal{F}_f$ using the cropped object regions in a similar manner.
Next, we~learn the weight and the kernel parameters of our CRF model.
We initialize them to the default values used in~\cite{krahenbuhl2011efficient}
and then use grid search to finetune their values on a held-out validation set.
We note that our model is not too sensitive to most of the parameters. On~each dataset, we~use fixed
values of these parameters for all CNN architectures. See~Section~\ref{sec:impl} for details.

\section{Experimental Evaluation}
\label{sec:experiments}

In this section, we~compare the proposed method with state-of-the-art semantic segmentation baselines.
We focus on two challenging
scenarios for waste object localization: the hand-held setting (for applications such as service robot interactions or smart
trash bins) and waste objects ``in the wild''.
In our experiments, we~found that one of the common challenges for both scenarios is
the extreme scale variation causing standard segmentation algorithms to underperform. Our~proposed method, however,
greatly improves the segmentation performance in these adverse scenarios.
Specifically, we~evaluate our method on the following two datasets:

\begin{itemize}[leftmargin=*,labelsep=5.5mm]
\item \textbf{MJU-Waste Dataset.}~In this work,
we created a new benchmark for waste object segmentation.
The dataset is available from~\url{https://github.com/realwecan/mju-waste/}.
To the best of our knowledge, MJU-Waste is the largest public benchmark available for waste object segmentation,
with $1485$ images for training, $248$ for validation and $742$ for testing.
For each color image, we~provide the co-registered depth image captured using an RGBD camera.
We manually labeled each of the image. More~details about our dataset are presented in Section~\ref{sec:mjudata}.

\item \textbf{TACO Dataset.}~The
Trash Annotations in COntext (TACO) dataset~\cite{proencca2020taco} is another public benchmark for waste object segmentation.
Images are collected from mainly outdoor environments such as woods, roads and beaches.
The dataset is available from~\url{http://tacodataset.org/}.
Individual images in this dataset are either under the CC BY 4.0 license or the ODBL (c) OpenLitterMap \& Contributors license.
See~\url{http://tacodataset.org/} for details.
The~current version
of the dataset contains $1500$ images, and~a split with $1200$ images for training, $150$ for validation and
$150$ for testing is available from the authors. In~all experiments that follow, we~use this split from the~authors.

\end{itemize}

{We summarize the key statistics of the two datasets in Table}~\ref{tab:datastat}.
{Once again, we~emphasize that one of the key characteristics of waste objects is that
the number of objects per class can be highly imbalanced (e.g., in~the case of TACO}~\cite{proencca2020taco}{).}
{In order to obtain sufficient data to train a strong segmentation algorithm,
we use a single class label for all waste objects, and~our problem
is therefore defined as a binary pixelwise prediction one (i.e., waste vs. background).}
For the quantitative evaluation that follows, we~report the performance of baseline methods and the proposed
method by four criteria: Intersection over Union (IoU) for the waste object class,
mean IoU (mIoU), pixel Precision (Prec) for the waste object class,
and Mean pixel precision (Mean). Let~TP, FP~and FN denote the total number of true positive, false positive and
false negative pixels, respectively. The~four
criteria used are defined as follows:

\begin{itemize}[leftmargin=*,labelsep=5.5mm]

\item Intersection over Union (IoU) for the $c$-th class is the intersection of the prediction and ground-truth
regions of the $c$-th class over the union of them, defined as:
\begin{align}
	\text{IoU}_c = \frac{\text{TP}_c}{\text{TP}_c+\text{FP}_c+\text{FN}_c}
\end{align}

\item mean IoU (mIoU) is the average IoU of all $C$ classes:
\begin{align}
	\text{mIoU} = \frac{1}{C} \sum_{c=1}^{C} \frac{\text{TP}_c}{\text{TP}_c+\text{FP}_c+\text{FN}_c}
\end{align}

\item Pixel Precision (Prec) for the $c$-th class is the percentage of correctly classified pixels of
all predictions of the $c$-th class:
\begin{align}
	\text{Prec}_c = \frac{\text{TP}_c}{\text{TP}_c+\text{FP}_c}
\end{align}

\item Mean pixel precision (Mean) is the average class-wise pixel precision:
\begin{align}
	\text{Mean} = \frac{1}{C} \sum_{c=1}^{C} \frac{\text{TP}_c}{\text{TP}_c+\text{FP}_c}
\end{align}
\end{itemize}

We note that the image labelings are typically dominated by the background class,
therefore IoU and Prec reported on the waste objects
only are more sensitive than mIoU and Mean which consider both waste objects and the background.

\begin{table}[H]
	\centering
	       \caption
         {
            {Key statistics of the two datasets used in our experimental evaluation.
            Data splits are the number of training $/$ validation $/$ test images.
            The image size varies in TACO so we report the average image size here.
            Objects p.i. is the number of objects per image.
            Currently, MJU-Waste uses a single class label for all waste objects
            (in addition to the background class). For~TACO, there are $60$ categories
            which belong to $28$ super (top) categories.}
         }
\label{tab:datastat}
		\begin{tabular}{cc c c c c c}
	        \toprule 
             & \textbf{Modalities} & \textbf{Images} & \textbf{Data Split} & \textbf{Image Size} & \textbf{Objects p.i.} & \textbf{Classes }\\
             \midrule
%             \multicolumn{10}{l}{} \\[-0.9em] % Spacer
			 MJU-Waste & RGBD & $2475$ & $1485/248/742$ & $640~\times~480$ & $1.02$ & single \\
			 TACO & RGB & $1500$ & $1200/150/150$ & $3223~\times~2825$ & $3.19$ & $60~(28)$ \\
			 \bottomrule
         \end{tabular}

\end{table}
\subsection{The MJU-Waste Dataset}
\label{sec:mjudata}

Before we move on to report our findings from the experiments, let~us more formally introduce the MJU-Waste dataset.
We created this dataset by collecting waste items
from a university campus, bringing them back to a lab, and~then take pictures of people holding waste items in their hands.
All~images in the dataset are captured using a Microsoft Kinect RGBD camera~\cite{han2013enhanced}.
The current version of our dataset, MJU-Waste V1, contains $2475$ co-registered RGB and depth image pairs.
Specifically, we~randomly split the images into a training set, a~validation set and a test set of $1485$, $248$ and $742$~images,
respectively.

\begin{figure}[H]
  \begin{tabular}{ll} % mother table
	  \centering
	\hspace{0mm}
    \begin{minipage}{1\textwidth}
      \begin{tabular}{cccc}
        \vspace{-3mm} \makecell{Color} & \hspace{-2mm}\makecell{\makecell{Ground-\\truth}}
        & \hspace{-3mm}\makecell{Raw\\Depth} & \hspace{-3mm}\makecell{Depth\\Processed}\vspace{3mm}\\  
        \includegraphics[width=0.11\textwidth]{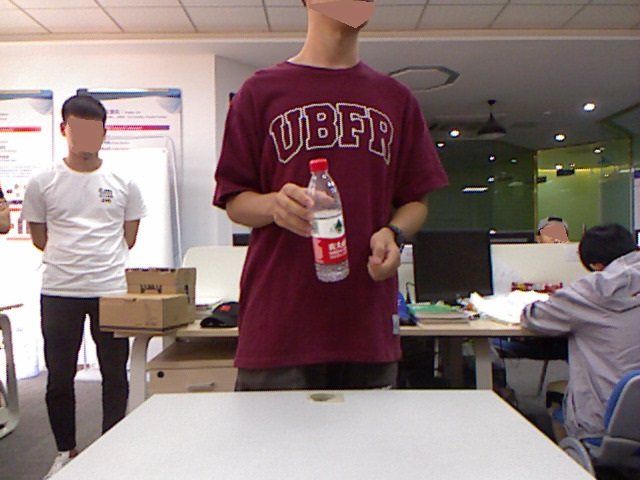} &
        {\hspace{-3.8mm}\includegraphics[width=0.11\textwidth]{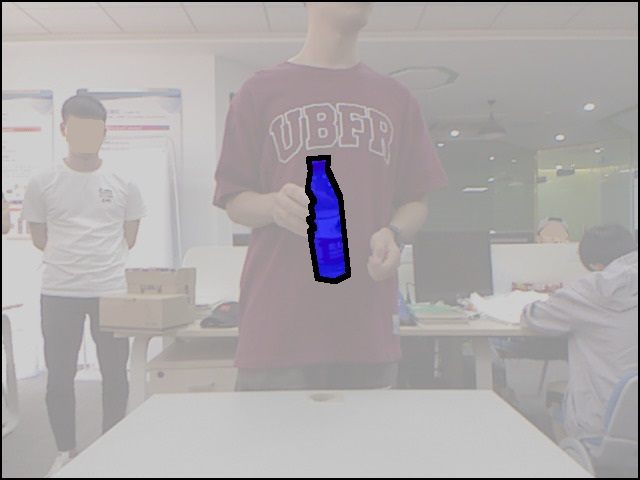}} &
        {\hspace{-4mm}\includegraphics[width=0.11\textwidth]{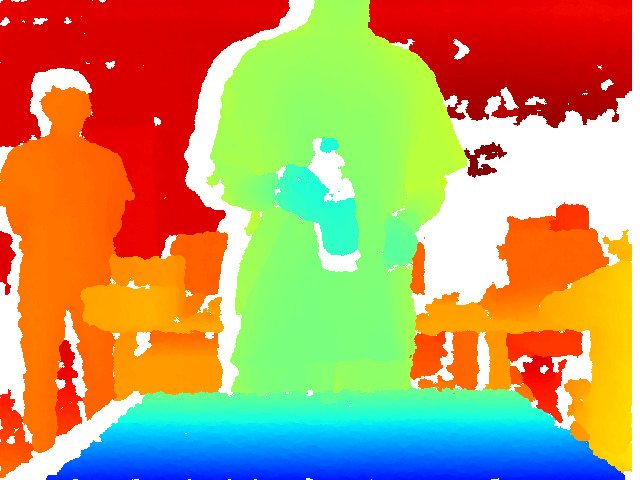}} &
        {\hspace{-4mm}\includegraphics[width=0.11\textwidth]{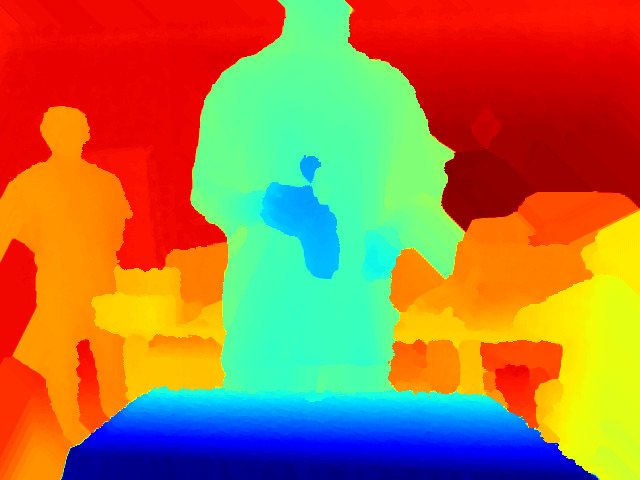}} \\
        \includegraphics[width=0.11\textwidth]{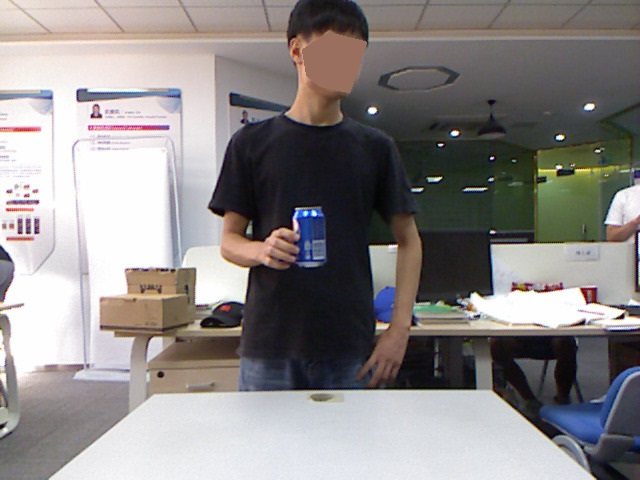} &
        {\hspace{-3.8mm}\includegraphics[width=0.11\textwidth]{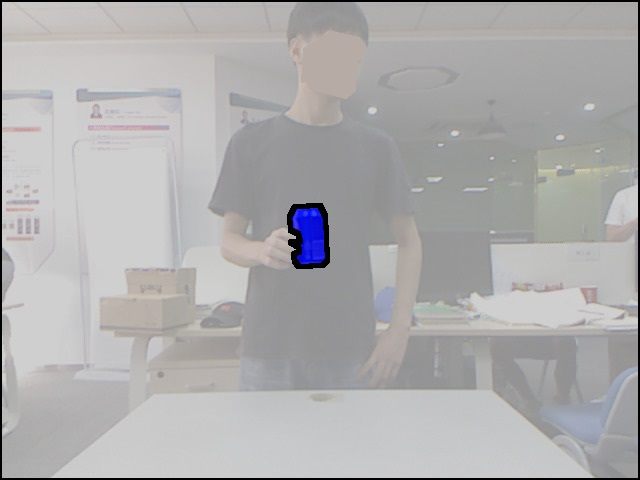}} &
        {\hspace{-4mm}\includegraphics[width=0.11\textwidth]{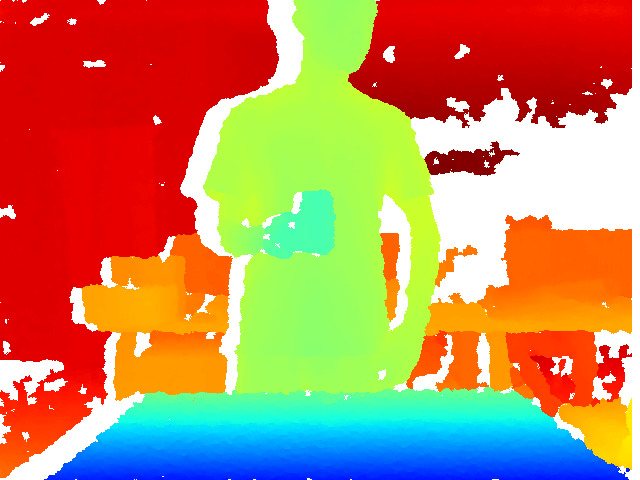}} &
        {\hspace{-4mm}\includegraphics[width=0.11\textwidth]{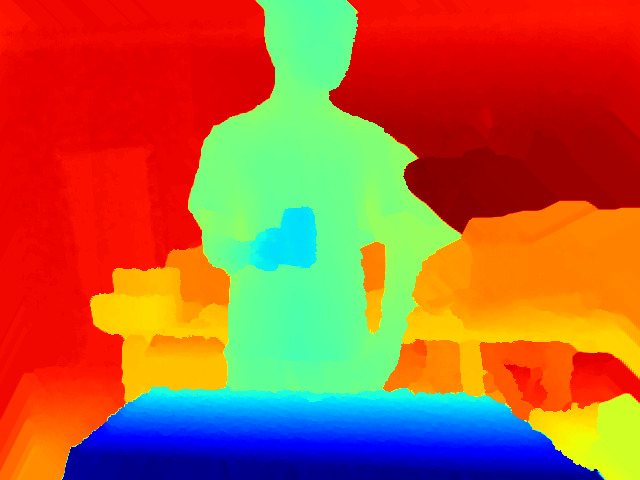}} \\
        \includegraphics[width=0.11\textwidth]{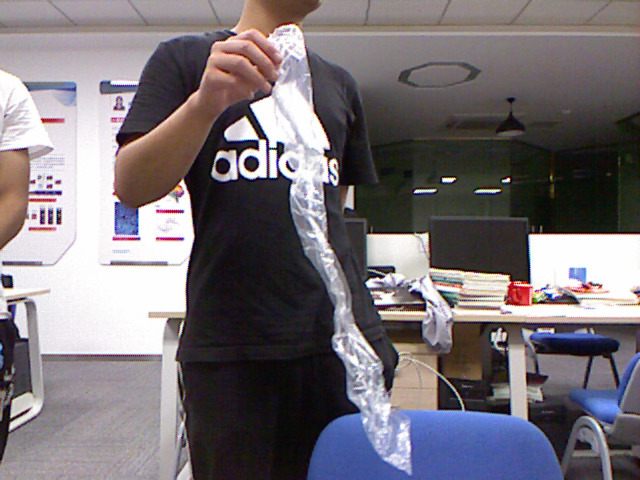} &
        {\hspace{-3.8mm}\includegraphics[width=0.11\textwidth]{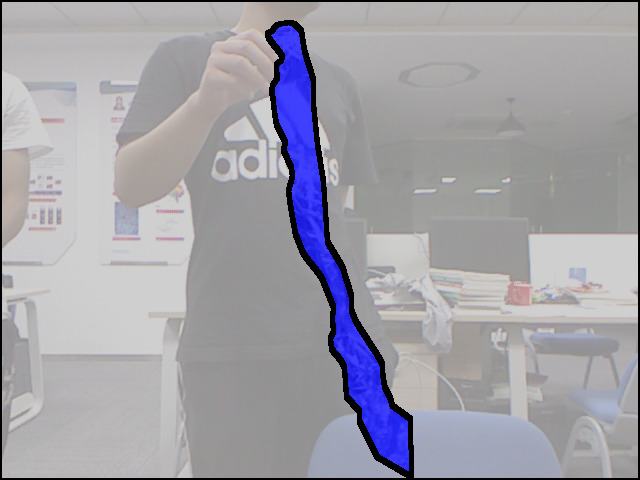}} &
        {\hspace{-4mm}\includegraphics[width=0.11\textwidth]{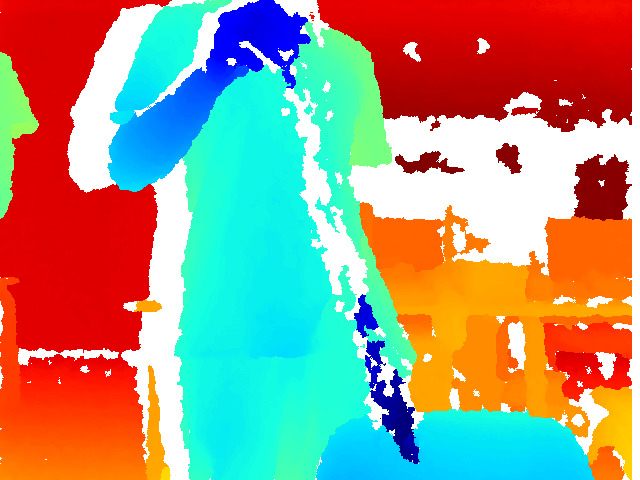}} &
        {\hspace{-4mm}\includegraphics[width=0.11\textwidth]{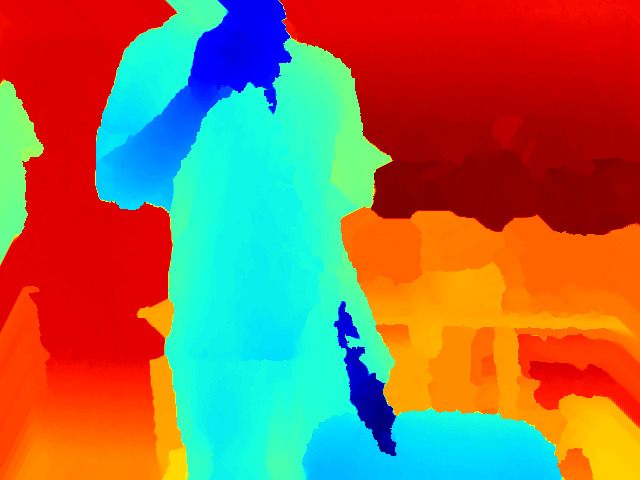}} \\                       
      \end{tabular}
    \end{minipage} & \hspace{-78mm}
    
    \begin{minipage}{1\textwidth}
      \begin{tabular}{cccc}
        \vspace{-3mm} \makecell{Color} & \hspace{-2mm}\makecell{\makecell{Ground-\\truth}}
        & \hspace{-3mm}\makecell{Raw\\Depth} & \hspace{-3mm}\makecell{Depth\\Processed}\vspace{3mm}\\  
        \includegraphics[width=0.11\textwidth]{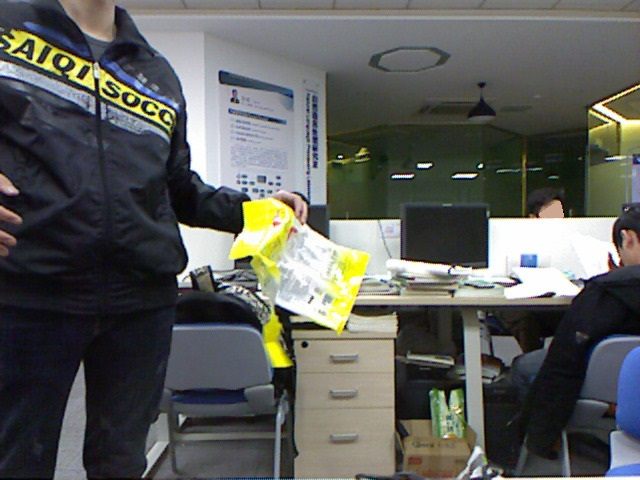} &
        {\hspace{-3.8mm}\includegraphics[width=0.11\textwidth]{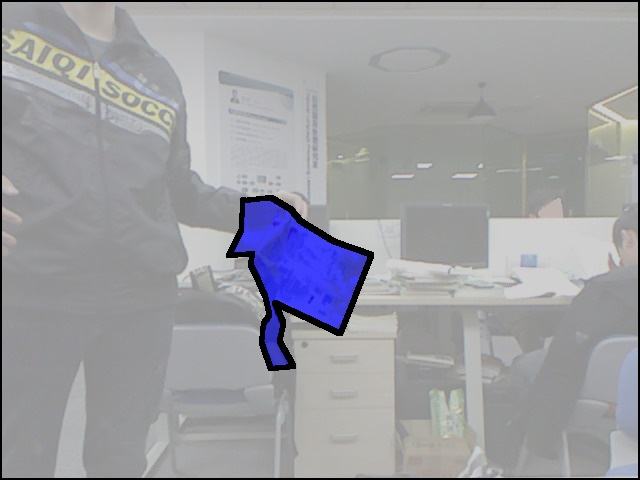}} &
        {\hspace{-4mm}\includegraphics[width=0.11\textwidth]{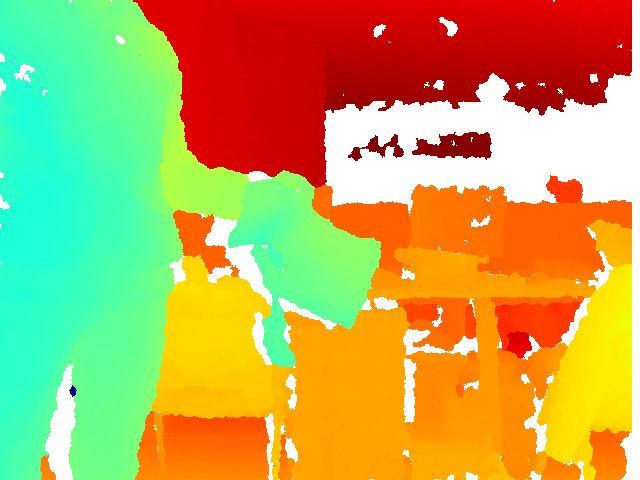}} &
        {\hspace{-4mm}\includegraphics[width=0.11\textwidth]{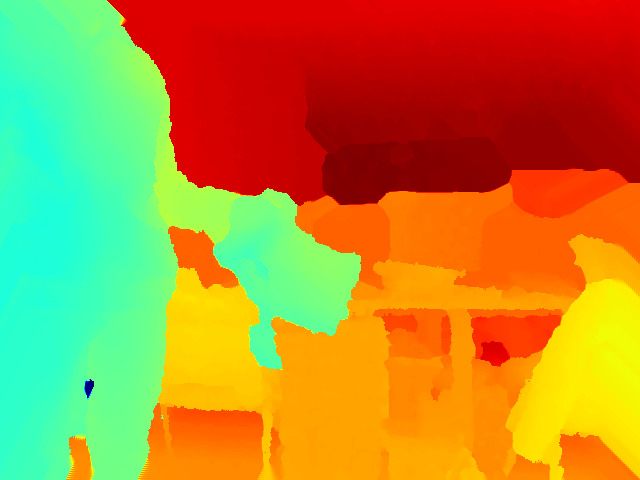}} \\
        \includegraphics[width=0.11\textwidth]{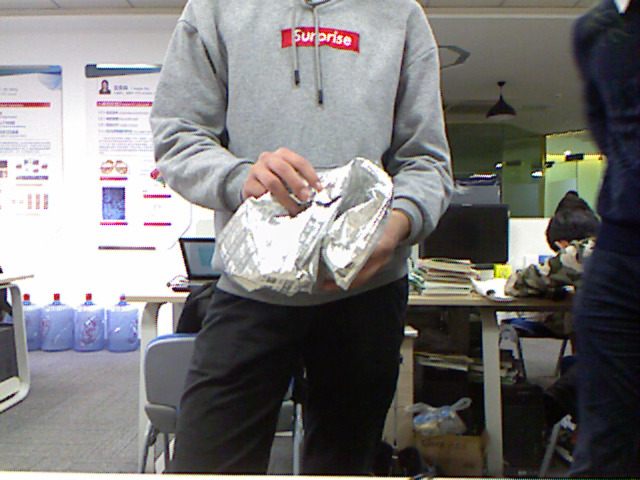} &
        {\hspace{-3.8mm}\includegraphics[width=0.11\textwidth]{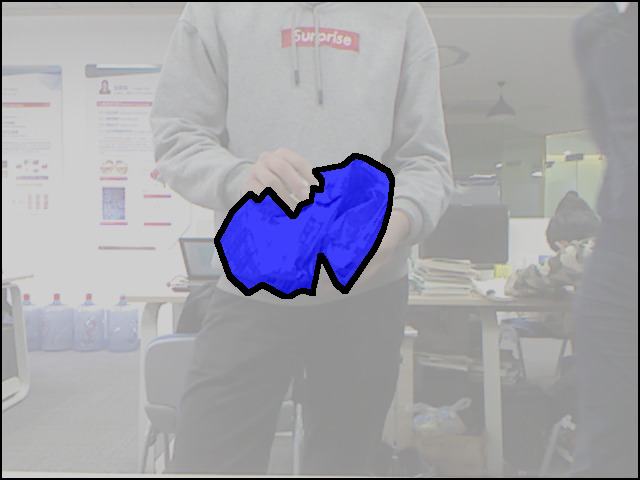}} &
        {\hspace{-4mm}\includegraphics[width=0.11\textwidth]{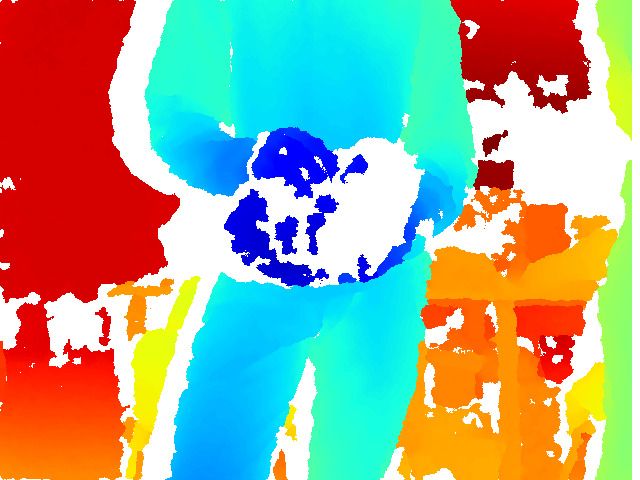}} &
        {\hspace{-4mm}\includegraphics[width=0.11\textwidth]{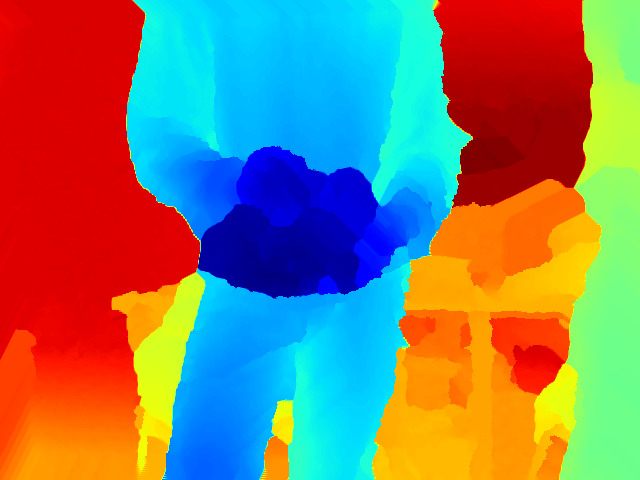}} \\
        \includegraphics[width=0.11\textwidth]{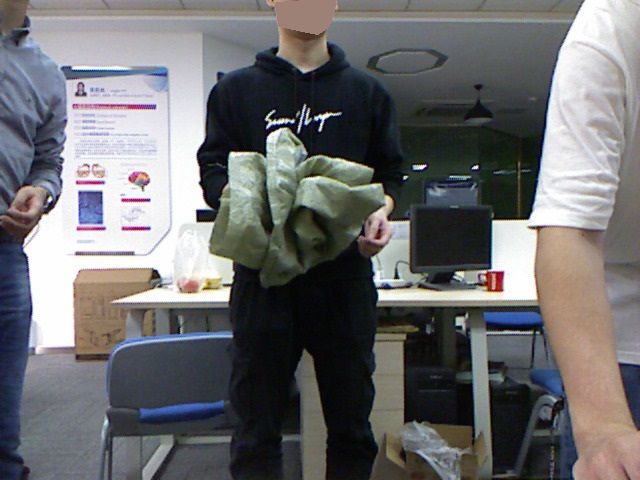} &
        {\hspace{-3.8mm}\includegraphics[width=0.11\textwidth]{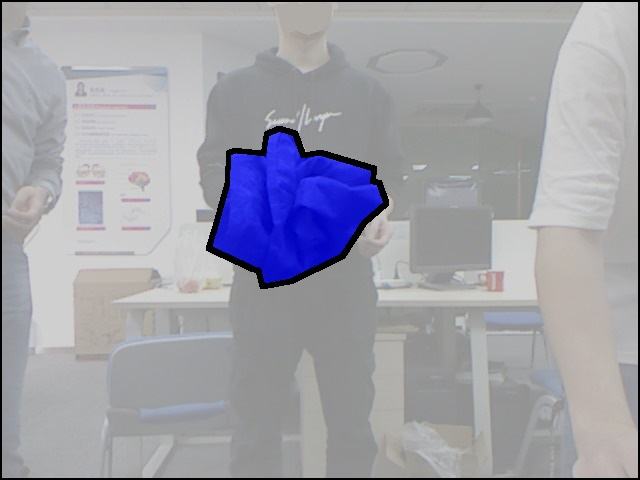}} &
        {\hspace{-4mm}\includegraphics[width=0.11\textwidth]{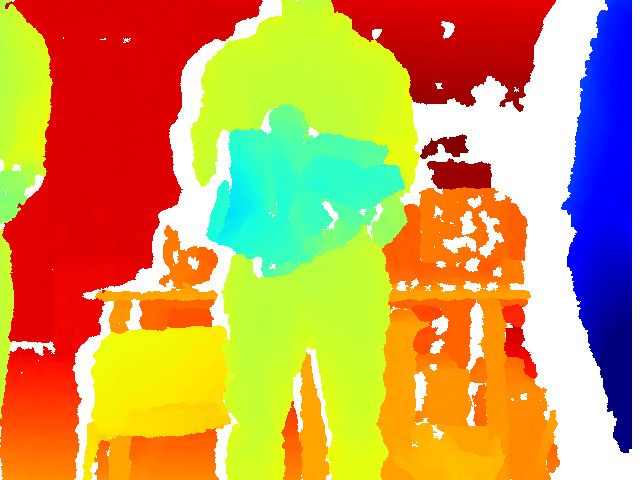}} &
        {\hspace{-4mm}\includegraphics[width=0.11\textwidth]{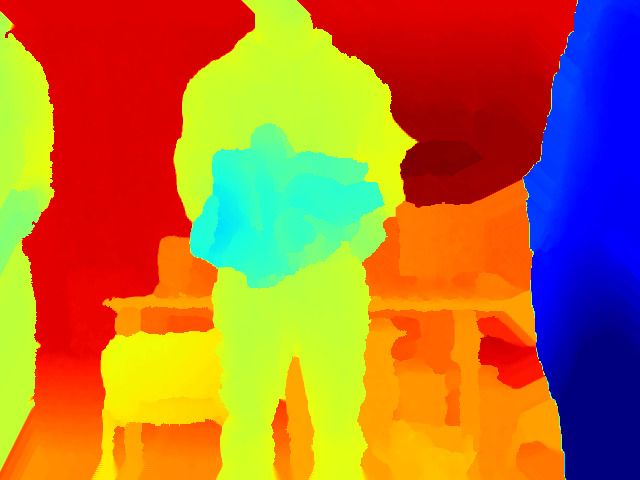}} \\                
      \end{tabular}
    \end{minipage}    

  \end{tabular}
	  
\vspace{1mm}
		\caption
         {
			Example color frames, ground-truth annotations, and~depth frames from the MJU-Waste dataset.
			Ground-truth masks are shown in blue.
			Missing values in the raw depth frames are shown in white.
			These values are filled in with a median filter following~\cite{lai2011large}.
         }
	\label{fig:mjuex}
\end{figure}

Due to sensor limitations, the~depth frames contain missing values at reflective surfaces,
occlusion boundaries, and~distant regions. We~use a median filter~\cite{lai2011large} to fill in the missing values in order to obtain high quality depth images.
Each image in MJU-Waste is annotated with a pixelwise mask of waste objects.
Example color frames, ground-truth annotations, and~depth frames are shown in Figure~\ref{fig:mjuex}.
In addition to semantic segmentation ground-truths, object instance masks are also~available.

\subsection{Implementation Details}
\label{sec:impl}

Here we report the key implementation details of our experiments, as~follows:

\begin{itemize}[leftmargin=*,labelsep=5.5mm]
\item \textbf{Segmentation networks $\mathcal{F}_c$ and $\mathcal{F}_f$.} Following~\cite{chen2017deeplab,zhao2017pyramid}, we~use the polynomial
learning rate policy with the initial learning rate set to $0.001$ and the power factor set to $0.9$.
The total number of iterations are set to $50$ epochs on both datasets with a batch size of $4$ images.
In all experiments, we~use the ImageNet-pretrained backbones~\cite{krizhevsky2012imagenet} and a standard SGD optimizer with
momentum and weight decay factors set to $0.9$ and $0.0001$, respectively. To~avoid overfitting, standard data
augmentation techniques including random mirroring, resizing (with a resize factor between $0.5$ and $2$), cropping and random
Gaussian blur~\cite{zhao2017pyramid} are used.
{The base (and cropped) image sizes for $\mathcal{F}_c$ and $\mathcal{F}_f$ are set to $520 (480)$ and $260 (240)$ pixels during training, respectively.}

\item \textbf{Object region proposals.} To maintain a concise set of object region proposals, we~empirically set the minimum and maximum
number of pixels $N_{\text{min}}$ and $N_{\text{max}}$ in an object region proposal. For~MJU-Waste, $N_{\text{min}}$ and $N_{\text{max}}$
are set to $900$ and 40,000, respectively. For~TACO, $N_{\text{min}}$ and $N_{\text{max}}$ are set to 25,000 and 250,000,
due to the larger image sizes. Object region proposals that are either too small or too big will simply be discarded.

\item \textbf{CRF parameters.} We initialize the CRF parameters with the default values in~\cite{krahenbuhl2011efficient} and follow
a simple grid search strategy to find the optimal values of CRF parameters in each term. For~reference, the~CRF parameters used
in our experiments are listed in Table~\ref{tab:crfparams}. We~note that our model is somewhat robust to the exact values of these
parameters, and~for each dataset we use the same parameters for all segmentation models.

\item \textbf{Training details and codes.} {In this work, we~use a publicly available
implementation to train the segmentation networks $\mathcal{F}_c$ and $\mathcal{F}_f$.
{The CNN training codes are available from:~\url{https://github.com/Tramac/awesome-semantic-segmentation-pytorch/}}.
We~use the default training settings unless otherwise specified earlier
in this section. For~CRF inference we use another public
implementation. {The CRF inference codes are available from:~\url{https://github.com/lucasb-eyer/pydensecrf/}.} The~complete set
of CRF parameters are summarized in Table}~\ref{tab:crfparams}.
\end{itemize}
\vspace{-10pt}

\begin{table}[H]
	\centering
	         \caption
         {
            CRF parameters used in our experiments. Depth terms are not applicable to the TACO dataset.
         }
\label{tab:crfparams}

		\begin{tabular}{c  c c c c c c c c c}
	        \toprule 
             & \boldmath{$\alpha$} & \boldmath{$w^{(a)}$} & \boldmath{$w^{(s)}$} & \boldmath{$w^{(d)}$} & \boldmath{$\theta_{\alpha}$}
             & \boldmath{$\theta_{\beta}$} & \boldmath{$\theta_{\gamma}$} & \boldmath{$\theta_{\delta}$} & \boldmath{$\theta_{\epsilon}$} \\
             \midrule
%             \multicolumn{10}{l}{} \\[-0.9em] % Spacer
			 MJU-Waste & $1$ & $3$ & $1$ & $1$ & $20$ & $20$ & $1$ & $10$ & $20$ \\
			 TACO & $1$ & $3$ & $1$ & - & $100$ & $20$ & $10$ & - & - \\
			 \bottomrule
         \end{tabular}
       %  \vspace{2mm}

\end{table}
\subsection{Results on the MJU-Waste Dataset}
\label{sec:res-mju}

The quantitative performance evaluation results we obtained on the test set of MJU-Waste are summarized in Table~\ref{tab:mjuexp}.
Methods using our proposed multi-level model have ``ML'' in their names. For~this dataset, we~report the performance of
the following baseline methods:

\begin{itemize}[leftmargin=*,labelsep=5.5mm]
\item \textbf{FCN-8s~\cite{long2015fully}.} {FCN is a seminal work in CNN-based semantic segmentation. In~particular,
FCN proposes to transform fully connected layers into convolutional layers that enables a classification net to output a
probabilistic heatmap of object layouts. In~our experiments, we~use the network architecture as proposed in}~\cite{long2015fully},
{which adopts a VGG16}~\cite{simonyan2014very} {backbone. In~terms of the skip connections, we~choose the FCN-8s variant as it retains
more precise location information by fusing features from the early $pool3$ and $pool4$ layers.}

\item \textbf{PSPNet~\cite{zhao2017pyramid}.} {PSPNet proposes the pyramid pooling module for multi-scale context aggregation.
Specifically, we~choose the ResNet-101}~\cite{he2016deep} {backbone variant for a good tradeoff between model complexity and performance.
The pyramid pooling module concatenates the features from the last layer of the $conv4$ block with the same features applied with
$1~\times~1$, $2~\times~2$, $3~\times~3$ and $6~\times~6$ average pooling and upsampling to harvest multi-scale contexts.}

\item \textbf{CCNet~\cite{huang2019ccnet}.} {CCNet presents an attention-based context aggregation method for semantic segmentation.
We also choose the ResNet-101 backbone for this method. Therefore, the~overall architecture is similar to
PSPNet except that we use the Recurrent Criss Cross Attention (RCCA) module for context modeling. Specifically, given the $conv4$ features,
the RCCA module obtains a self-attention map to aggregate the context information in horizontal and vertical directions.
Similarly, the~resultant features are concatenated with the $conv4$ features for downstream~segmentation.}

\item \textbf{DeepLabv3~\cite{chen2017rethinking}.} {DeepLabv3 proposes the Atrous Spatial Pyramid Pooling (ASPP) module for capturing
the long-range contexts. Specifically, ASPP~proposes the parallel dilated convolutions with varying atrous rates to encode features
from different sized receptive fields. The~atrous rates used in our experiments are $12$, $24$ and $36$. In~addition, we~experimented with
both ResNet-50 and ResNet-101 backbones on the MJU-Waste dataset to explore the performance impact of different backbone architectures.}
\end{itemize}

\begin{table}[H]
	\centering
	      \caption
         {
            Performance comparisons on the test set of MJU-Waste. For~each method, we~report the IoU for waste objects (IoU),
            mean IoU (mIoU), pixel Precision for waste objects (Prec) and Mean pixel precision (Mean). See~Section~\ref{sec:res-mju} for details.
         }
\label{tab:mjuexp}

		\begin{tabular}{l l  c c c c}
	         \toprule
             \makecell[bl]{\textbf{Dataset:} \\\textbf{MJU-Waste (test)}} & \makecell[bl]{\textbf{Backbone}} & \textbf{IoU} & \textbf{mIoU} & \textbf{Prec} & \textbf{Mean} \\
             \midrule
%             \multicolumn{10}{l}{} \\[-0.9em] % Spacer
             \multicolumn{6}{l}{Baseline Approaches} \\
             \hline
             &&&&&\\[-2ex]
%             \multicolumn{10}{l}{} \\[-0.9em] % Spacer
             FCN-8s~\cite{long2015fully} & \scriptsize VGG-16 & 75.28 & 87.35 & 85.95 & 92.83 \\          
             PSPNet~\cite{zhao2017pyramid} & \scriptsize ResNet-101 & 78.62 & 89.06 & 86.42 & 93.11 \\
             CCNet~\cite{huang2019ccnet} & \scriptsize ResNet-101 & 83.44 & 91.54 & \bf{92.92} & \bf{96.35} \\
             DeepLabv3~\cite{chen2017rethinking} & \scriptsize ResNet-50 & 79.92 & 89.73 & 86.30 & 93.06 \\
             DeepLabv3~\cite{chen2017rethinking} & \scriptsize ResNet-101 & \bf{84.11} & \bf{91.88} & 89.69 & 94.77 \\     
%             \multicolumn{10}{l}{} \\[-0.9em] % Spacer
             \hline
             &&&&&\\[-2ex]
 %            \multicolumn{10}{l}{} \\[-0.9em] % Spacer             
             \multicolumn{6}{l}{Proposed Multi-Level (ML) Model} \\[.1em]
             \hline
             &&&&&\\[-2ex]
%             \multicolumn{10}{l}{} \\[-0.9em] % Spacer     
             FCN-8s-ML & \scriptsize VGG-16 & 82.29 & 90.95 & 91.75 & 95.76 \\
              & & \scriptsize(+7.01) & \scriptsize(+3.60) & \scriptsize(+5.80) & \scriptsize(+2.93) \\
             PSPNet-ML & \scriptsize ResNet-101 & 81.81 & 90.70 & 89.65 & 94.73 \\
              & & \scriptsize(+3.19) & \scriptsize(+1.64) & \scriptsize(+3.23) & \scriptsize(+1.62) \\
             CCNet-ML & \scriptsize ResNet-101 & 86.63 & 93.17 & \bf{96.05} & \bf{97.92} \\
              & & \scriptsize(+3.19) & \scriptsize(+1.63) & \scriptsize(+3.13) & \scriptsize(+1.57) \\             
             DeepLabv3-ML & \scriptsize ResNet-50 & 84.35 & 92.00 & 91.73 & 95.78 \\
              & & \scriptsize(+4.43) & \scriptsize(+2.27) & \scriptsize(+5.43) & \scriptsize(+2.72) \\             
             DeepLabv3-ML & \scriptsize ResNet-101 & \bf{87.84} & \bf{93.79} & 94.43 & 97.14 \\             
              & & \scriptsize(+3.73) & \scriptsize(+1.91) & \scriptsize(+4.74) & \scriptsize(+2.37) \\                          [.5ex]
           \noalign{\hrule height 1.0pt}  
         \end{tabular}
      %   \vspace{3mm}
   
\end{table}

\begin{figure}[H]
	  \centering
      \begin{tabular}{cccccc}
        \scriptsize{Image} & \hspace{-4mm} \scriptsize{Ground-truth} & \hspace{-4mm} \scriptsize{FCN-8s} & \hspace{-4mm} \scriptsize{FCN-8s-ML} & \hspace{-4mm} \scriptsize{DeepLabv3} & \hspace{-4mm} \scriptsize{DeepLabv3-ML} \\
        \includegraphics[width=0.16\textwidth]{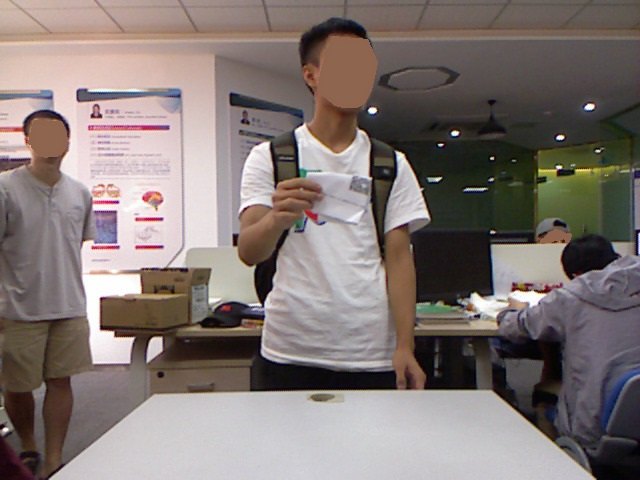} &
        {\hspace{-4mm}\includegraphics[width=0.16\textwidth]{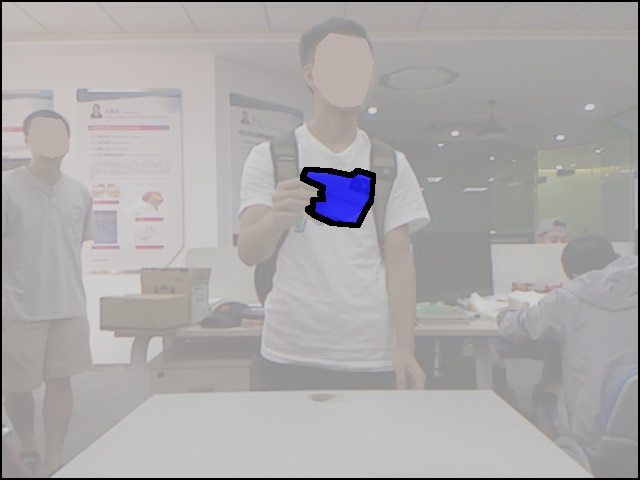}} &
        {\hspace{-4mm}\includegraphics[width=0.16\textwidth]{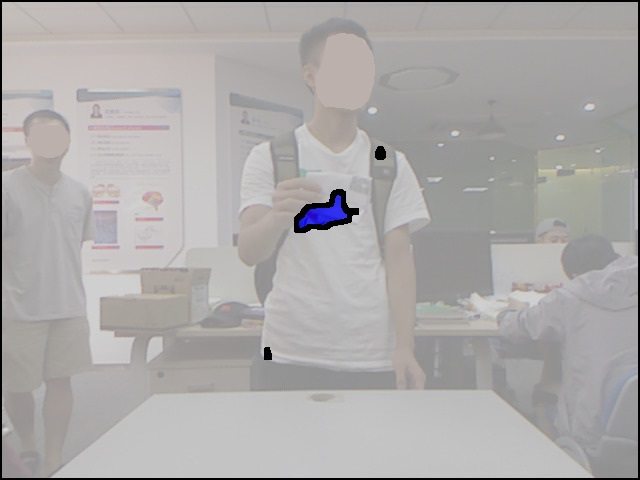}} &
        {\hspace{-4mm}\includegraphics[width=0.16\textwidth]{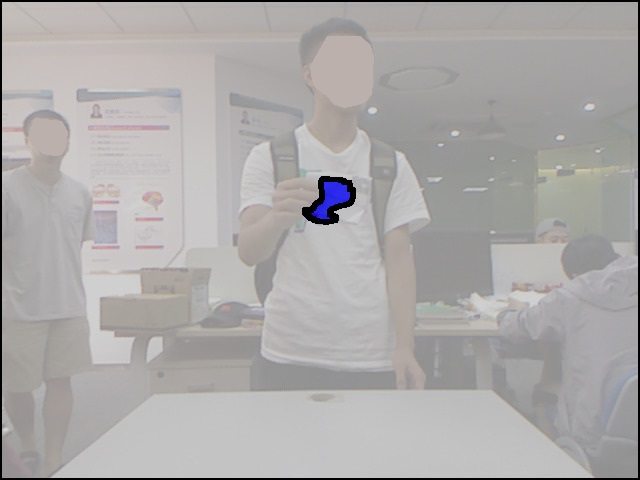}} &
        {\hspace{-4mm}\includegraphics[width=0.16\textwidth]{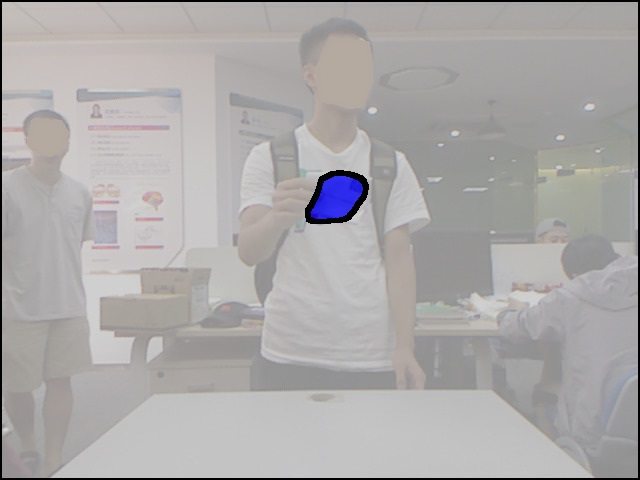}} &        
        {\hspace{-4mm}\includegraphics[width=0.16\textwidth]{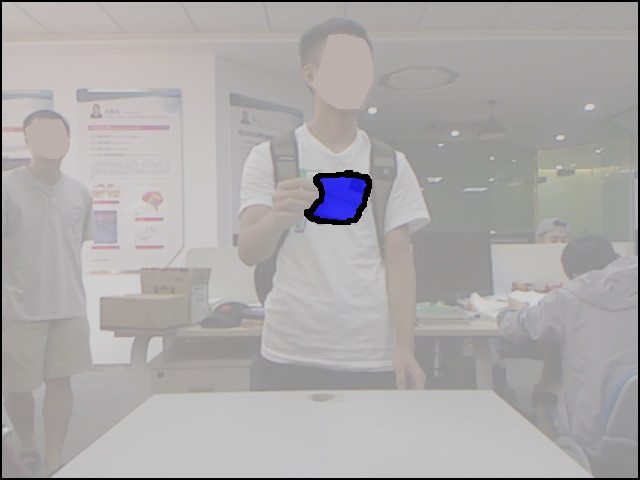}} \vspace{1mm}\\                
        \includegraphics[width=0.16\textwidth]{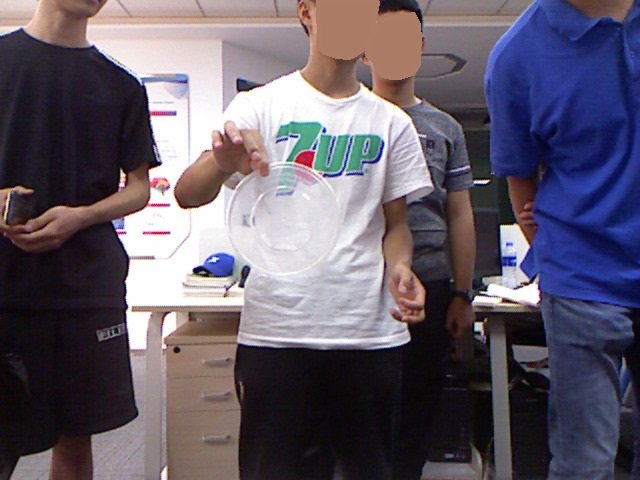} &
        {\hspace{-4mm}\includegraphics[width=0.16\textwidth]{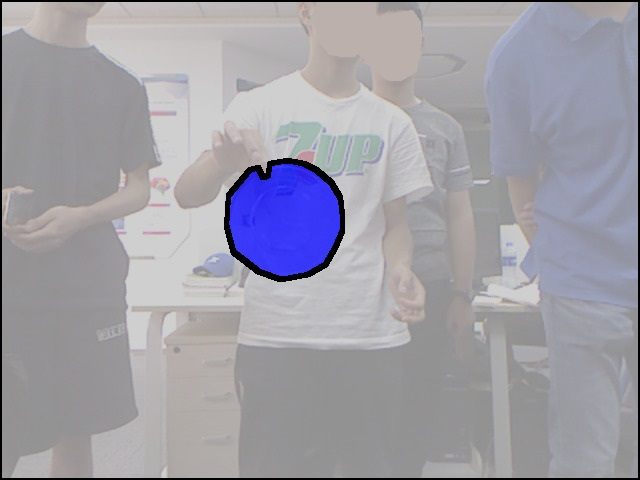}} &
        {\hspace{-4mm}\includegraphics[width=0.16\textwidth]{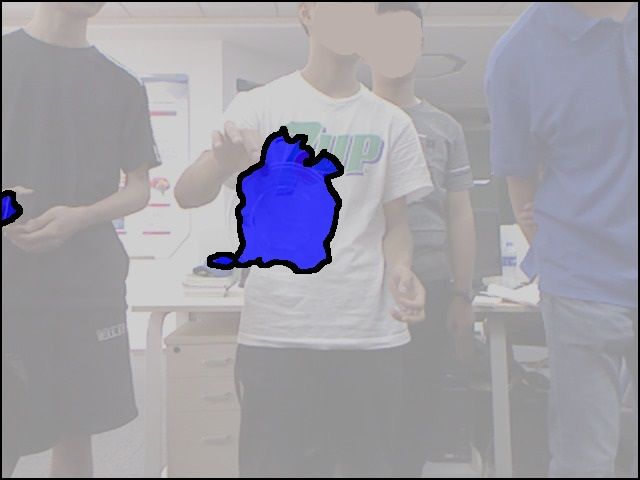}} &
        {\hspace{-4mm}\includegraphics[width=0.16\textwidth]{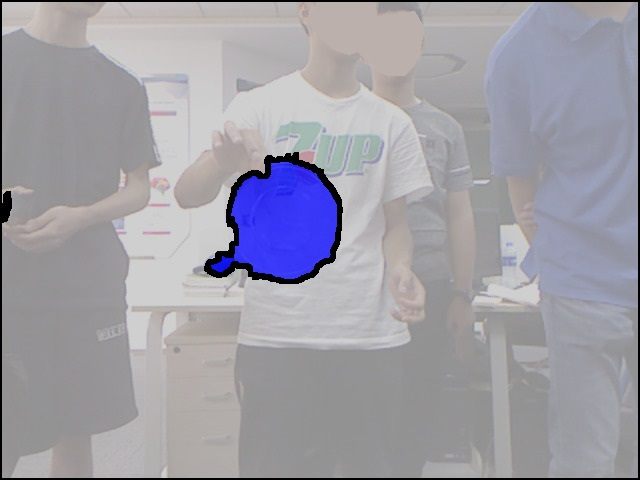}} &
        {\hspace{-4mm}\includegraphics[width=0.16\textwidth]{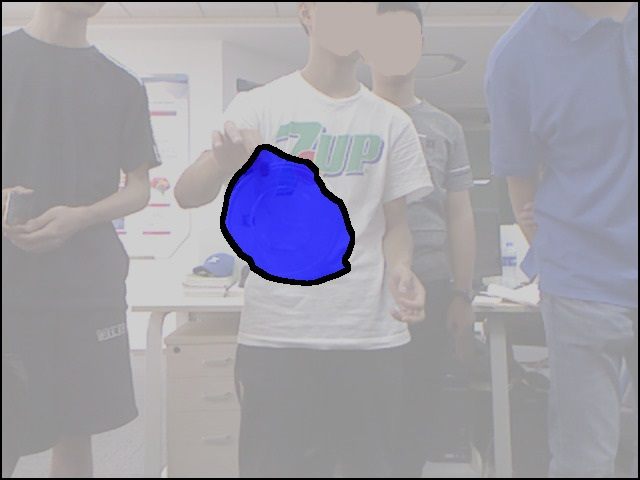}} &        
        {\hspace{-4mm}\includegraphics[width=0.16\textwidth]{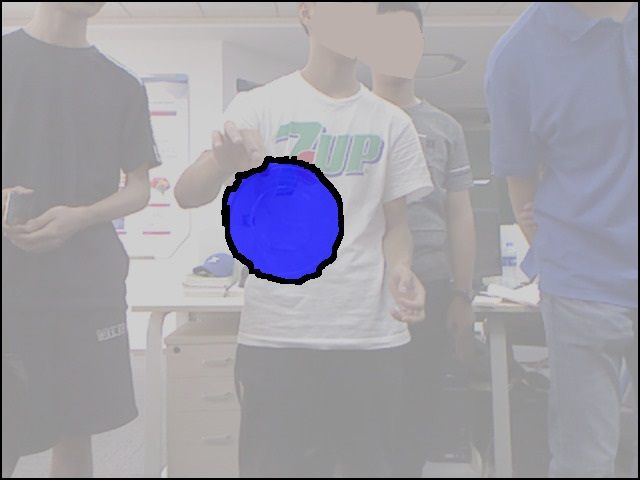}} \vspace{1mm}\\        
        \includegraphics[width=0.16\textwidth]{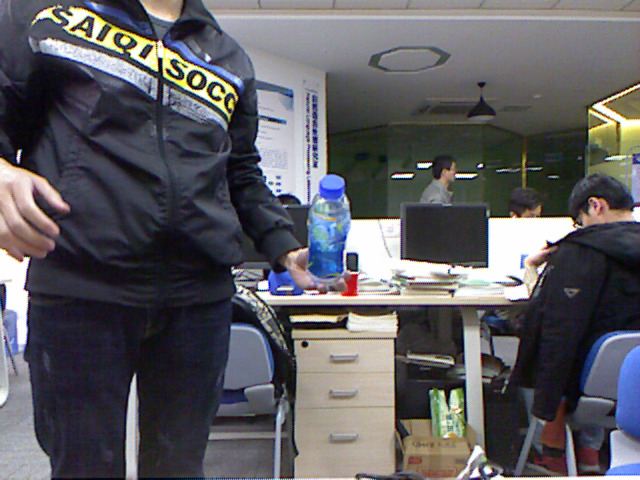} &
        {\hspace{-4mm}\includegraphics[width=0.16\textwidth]{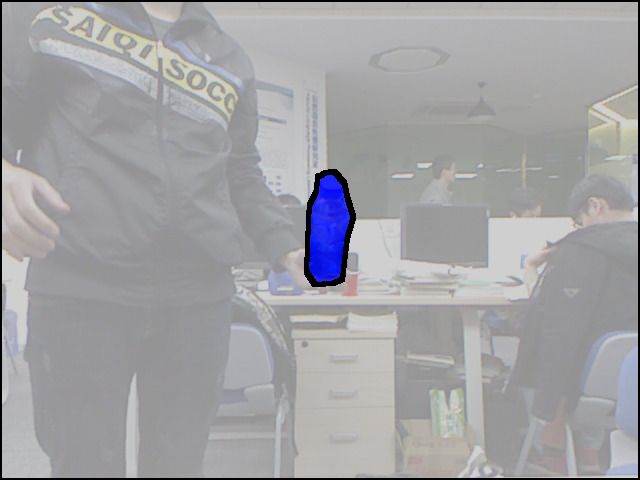}} &
        {\hspace{-4mm}\includegraphics[width=0.16\textwidth]{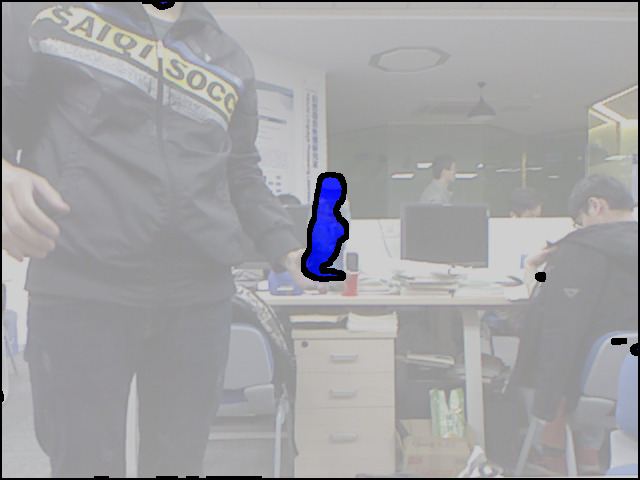}} &
        {\hspace{-4mm}\includegraphics[width=0.16\textwidth]{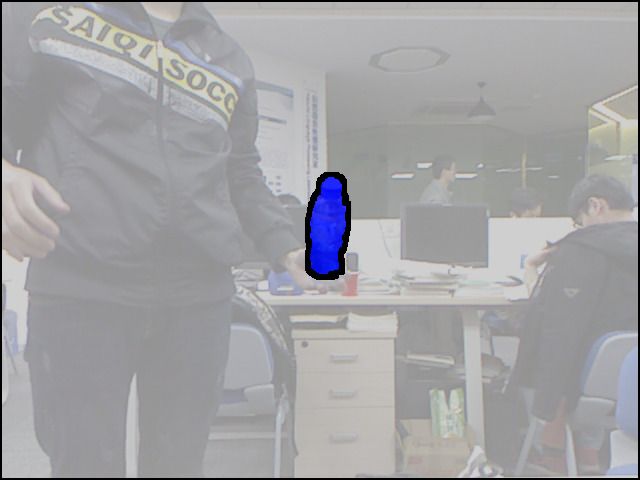}} &
        {\hspace{-4mm}\includegraphics[width=0.16\textwidth]{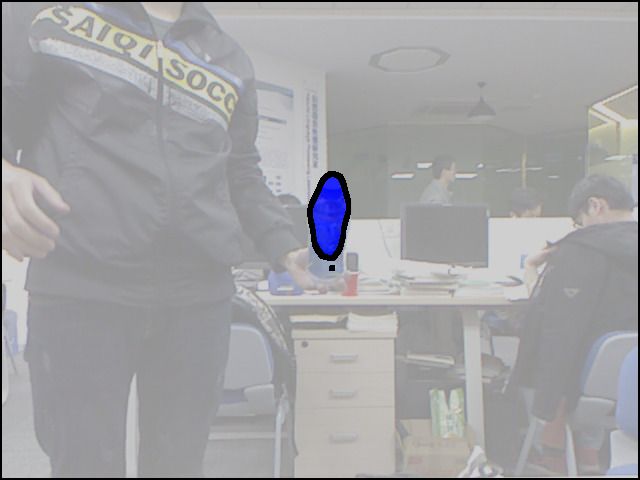}} &        
        {\hspace{-4mm}\includegraphics[width=0.16\textwidth]{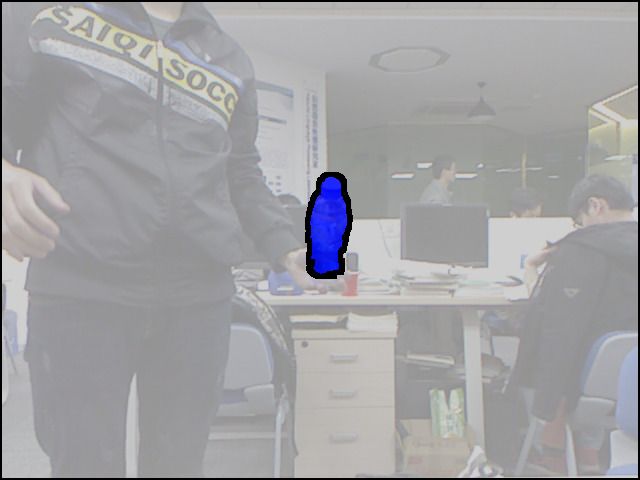}} \vspace{1mm}\\ 
        \includegraphics[width=0.16\textwidth]{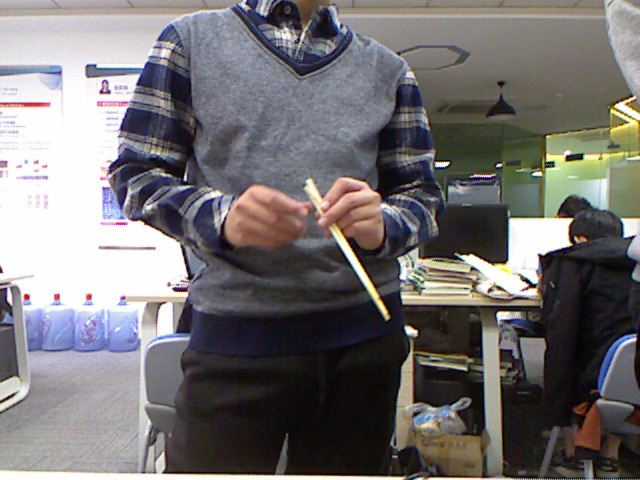} &
        {\hspace{-4mm}\includegraphics[width=0.16\textwidth]{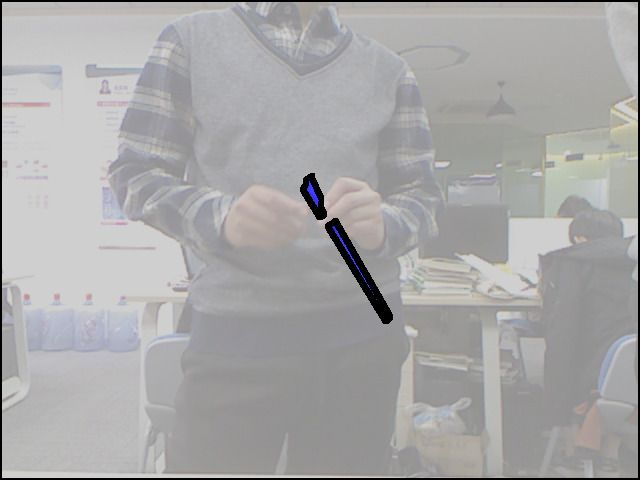}} &
        {\hspace{-4mm}\includegraphics[width=0.16\textwidth]{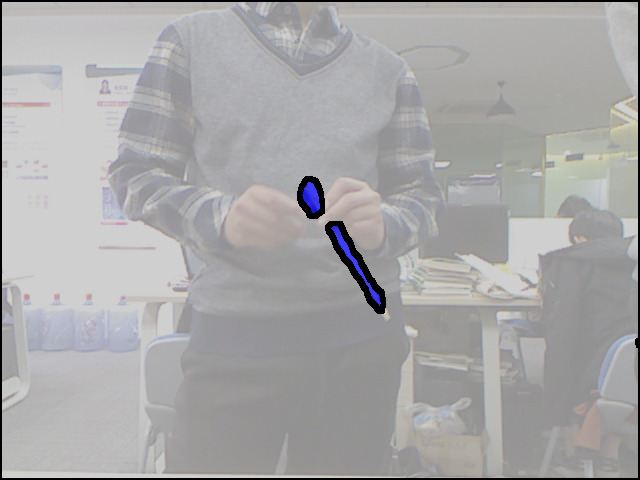}} &
        {\hspace{-4mm}\includegraphics[width=0.16\textwidth]{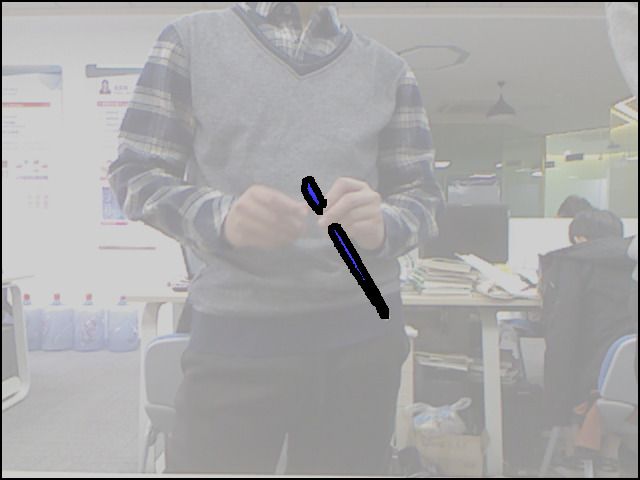}} &
        {\hspace{-4mm}\includegraphics[width=0.16\textwidth]{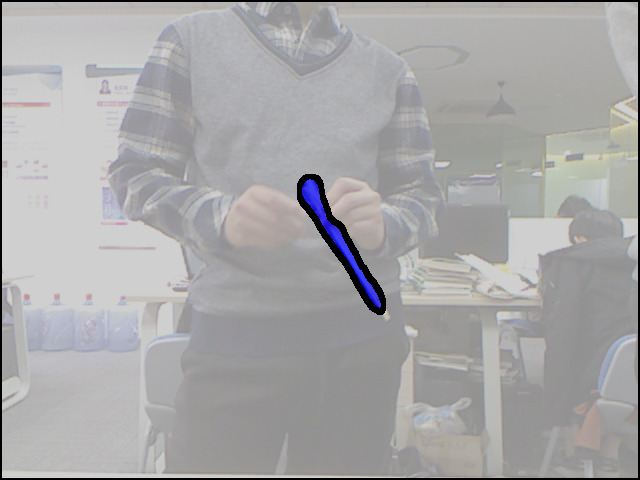}} &        
        {\hspace{-4mm}\includegraphics[width=0.16\textwidth]{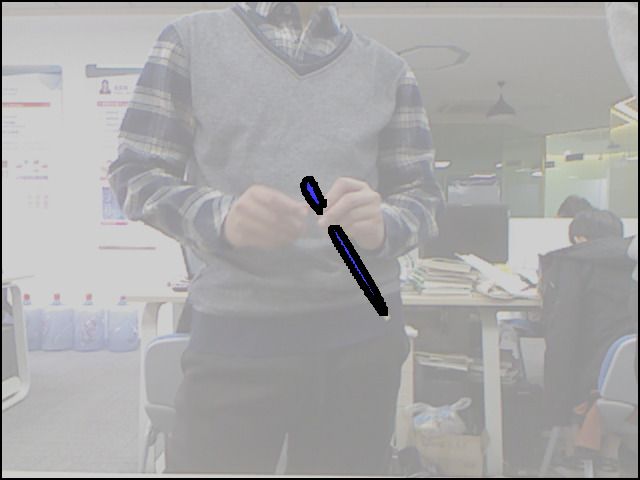}} \vspace{1mm}\\
      \end{tabular}
		\caption
         {
			Example segmentation results on the MJU-Waste test set. Input images and ground-truth annotations are shown
			in the first two columns. Baseline methods are FCN-8s (VGG-16) and DeepLabv3 (ResNet-50).
			Our proposed methods (FCN-8s-ML and DeepLabv3-ML) more accurately recover object boundaries.
			Best viewed electronically, zoomed in.
         }
      %\vspace{3mm}
      \label{fig:segs-mju}
\end{figure}

{We refer interested readers to the public implementation discussed in Section}~\ref{sec:impl} {for the network details of the above baselines.}
For each baseline method, we~additionally implement our
proposed multi-level modules and then present a direct performance comparison in terms of IoU, mIoU, Prec~and Mean improvements.
We show that our method provides a general framework under which a number of strong semantic segmentation baselines could be further improved.
For example, FCN-8s benefits the most from a multi-level approach (i.e., $+7.01$ points of IoU improvement),
partially due to the relatively low baseline performance.
Even for the best-performing baseline, DeepLabv3 with a ResNet-101 backbone, our~multi-level model further improves its performance
by $+3.73$ IoU points. We~note that such a large quantitative improvement can also be visually significant. In~Figure~\ref{fig:segs-mju},
we~present qualitative comparisons between FCN-8s, DeepLabv3 and their multi-level counterparts. It~is clear that our approach
helps to remove false positives in some non-object regions. More~importantly, it~is evident that
multi-level models more precisely follow object boundaries.

In Table~\ref{tab:ablations}, we~additionally perform ablation studies on the validation set of
MJU-Waste. Specifically, we~compare the performance of the following variants of our method:

\begin{itemize}[leftmargin=*,labelsep=5.5mm]

\item \textbf{Baseline.} DeepLabv3 baseline with a ResNet-50 backbone.

\item \textbf{Object only.} The above baseline with additional object-level reasoning. This~method is implemented by retaining only the two unary terms of
Equation~(\ref{eqn:overall}). All~pixel-level pairwise terms are turned off. This~will test if the object-level reasoning will contribute
to the baseline~performance.

\item \textbf{Object and appearance.} The baseline with object-level reasoning plus the appearance and the spatial smoothing pairwise terms. The~depth pairwise terms are turned off. This~will test if the additional pixel affinity information (without depth, however) is useful. It~also verifies the efficacy of the depth pairwise terms.

\item \textbf{Appearance and depth.} The baseline with all pixel-level pairwise terms but without the object-level unary term. This~will test if an object-level
fine segmentation network is necessary, as~well as the performance contribution of the pixel-level pairwise terms alone.

\item \textbf{Full model.} Our full model with all components proposed in Section~\ref{sec:approach}.

\end{itemize}

\begin{table}[H]
	\centering
    \caption
         {
            Results from our ablation studies carried out on the validation set of MJU-Waste.
            The baseline method is DeepLabv3 with a ResNet-50 backbone. We~add different
            components proposed in Section~\ref{sec:approach} individually to test their
            performance impact. See~Section~\ref{sec:res-mju} for details.
         }
\label{tab:ablations}
\begin{tabular}{ccccx{22}x{22}x{22}x{22}}
\toprule
 \makecell[bl]{\textbf{Dataset:} \\\textbf{MJU-Waste (val)}}
 & \textbf{Object?} &  \textbf{Appearance?} &  \textbf{Depth?}
 & \textbf{IoU} & \textbf{mIoU} & \textbf{Prec} & \textbf{Mean} \\
\midrule
 Baseline
  & \xmark & \xmark & \xmark & 80.86 & 90.24 & 87.49 & 93.67 \\
\midrule
 \multirow{3}{*}{+~components}
  & \cmark & \xmark & \xmark & 81.43 & 90.53 & 88.06 & 93.96 \\
  & \cmark & \cmark & \xmark & 85.44 & 92.58 & 91.79 & 95.83 \\
  & \xmark & \cmark & \cmark & 83.45 & 91.57 & 91.84 & 95.83 \\
\midrule
 Full model & \cmark & \cmark & \cmark & \bd{86.07} & \bd{92.90} & \bd{92.77} & \bd{96.32}\\
 \bottomrule
\end{tabular}

        % \vspace{2mm}

\end{table}

Results are clear that the full model performs the best, producing superior performance by all four criteria. This~validates that
the various components proposed in our method all positively impact the final results.

\begin{table}[H]
 %   \vspace{3mm}
	%\renewcommand{\arraystretch}{1.3}
	\centering
	   \caption
         {
            {Average per-image inference time on MJU-Waste. The~baseline method is DeepLabv3 with a ResNet-50 backbone,
            which corresponds to the scene-level inference time. Additional object and pixel level inference incurs
            extra computational costs. System specs: i9-9900KS CPU, 64GB DDR4 RAM, RTX~2080Ti GPU. Test~batch size
            set to $1$ with FP32 precision. See~Section}~\ref{sec:res-mju} {for details.}
         }
\label{tab:time}
		\begin{tabular}{c c c c c}
	        \toprule 
             \textbf{MJU-Waste (val)} & \textbf{Scene-Level} & \textbf{Object-Level }& \textbf{Pixel-Level} & \textbf{Total} \\
             \midrule
%             \multicolumn{10}{l}{} \\[-0.9em] % Spacer
			 inference time (ms) & $52$ & $352$ & $398$ & $802$ \\
			 \bottomrule
         \end{tabular}
      
%\vspace{3mm}
\end{table}

\begin{figure}[H]
	  \centering
      \begin{tabular}{cccccc}
        \scriptsize{Image} & \hspace{-4mm} \scriptsize{Prediction} & \hspace{-4mm} \scriptsize{Image} & \hspace{-4mm} \scriptsize{Prediction} & \hspace{-4mm} \scriptsize{Image} & \hspace{-4mm} \scriptsize{Prediction} \\
		\includegraphics[width=0.16\textwidth]{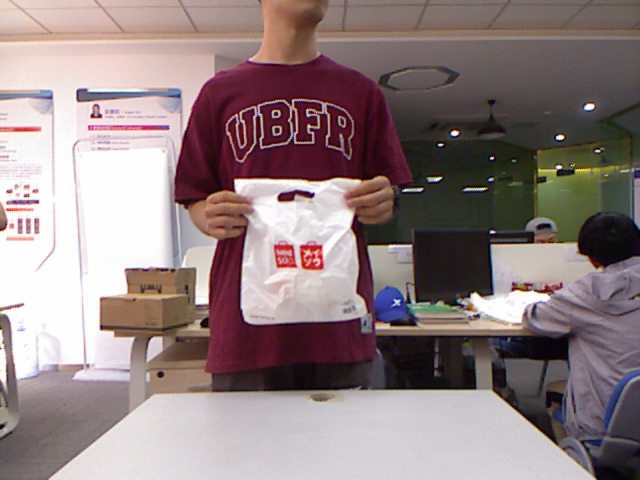} &
        {\hspace{-4mm}\includegraphics[width=0.16\textwidth]{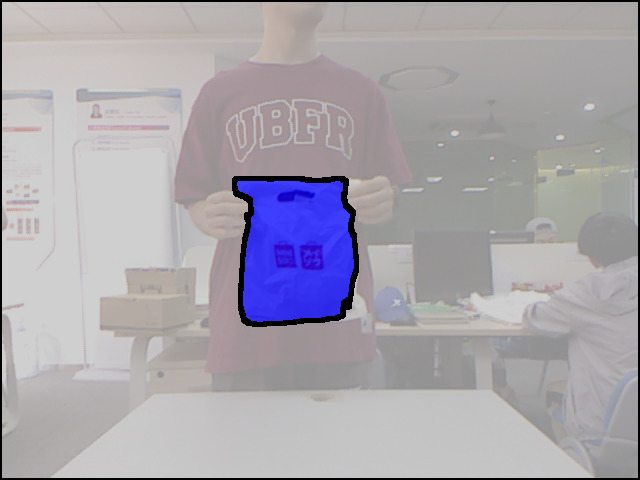}} &
        {\hspace{-4mm}\includegraphics[width=0.16\textwidth]{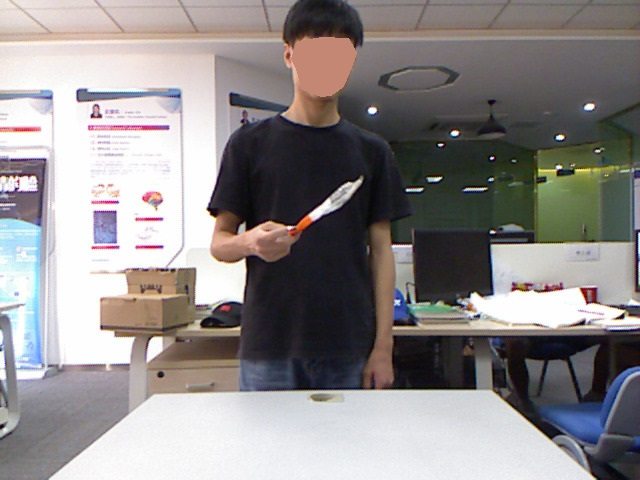}} &
        {\hspace{-4mm}\includegraphics[width=0.16\textwidth]{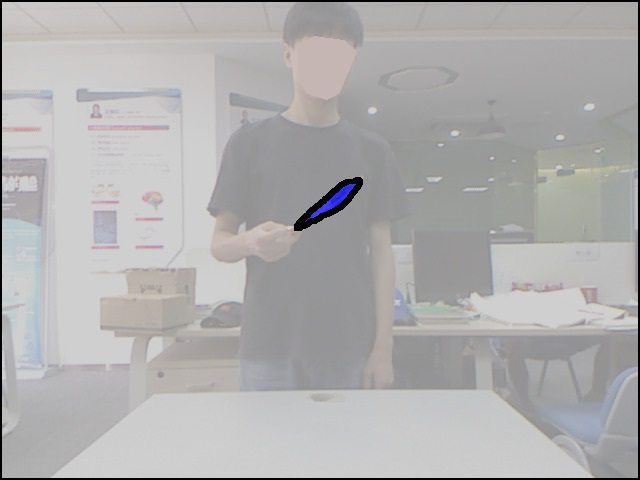}} &
        {\hspace{-4mm}\includegraphics[width=0.16\textwidth]{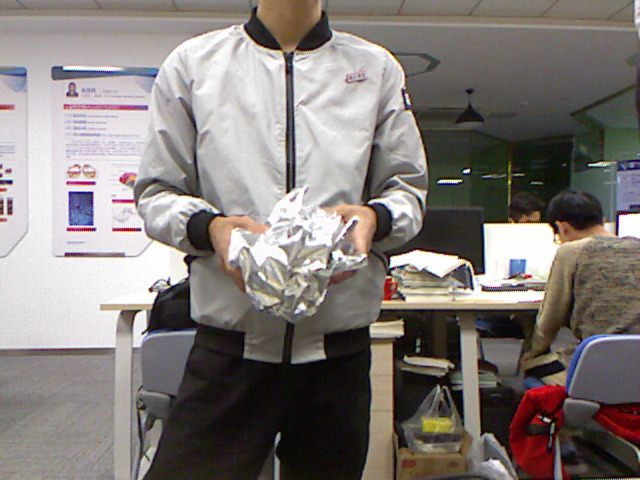}} &        
        {\hspace{-4mm}\includegraphics[width=0.16\textwidth]{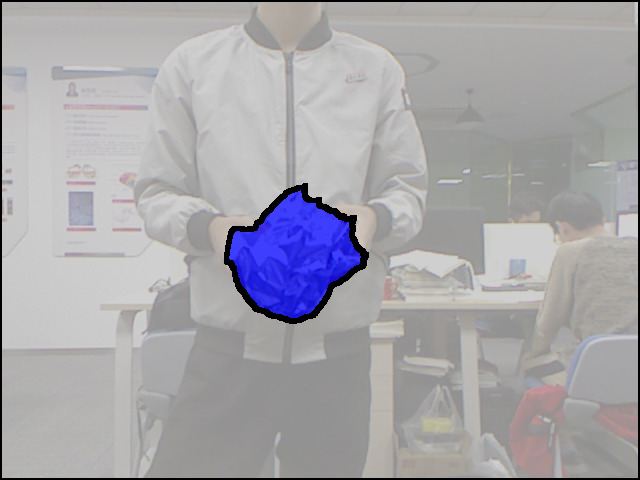}} \vspace{1mm}\\     
		\includegraphics[width=0.16\textwidth]{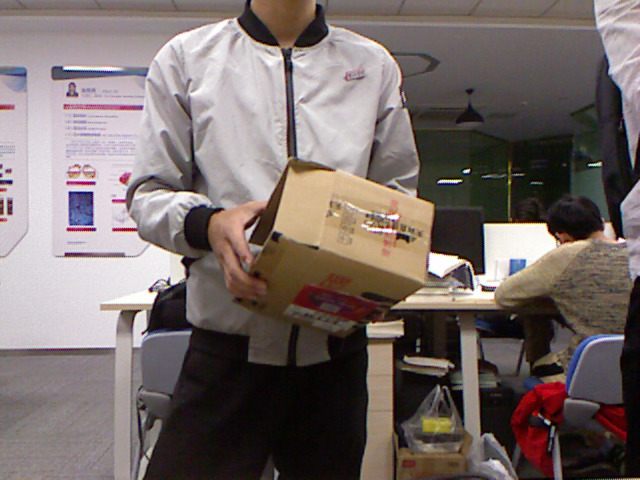} &
        {\hspace{-4mm}\includegraphics[width=0.16\textwidth]{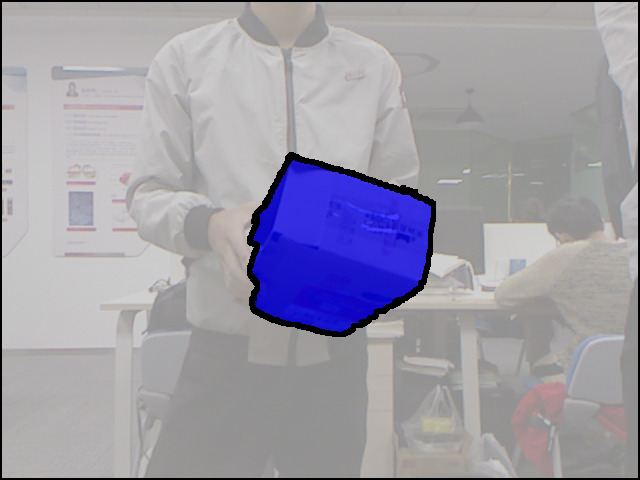}} &
        {\hspace{-4mm}\includegraphics[width=0.16\textwidth]{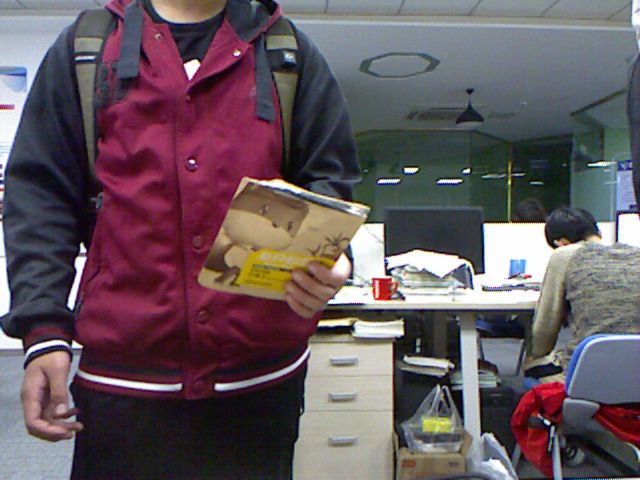}} &
        {\hspace{-4mm}\includegraphics[width=0.16\textwidth]{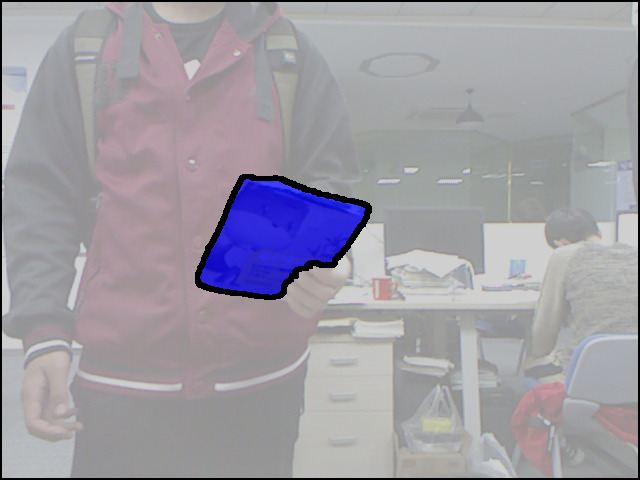}} &
        {\hspace{-4mm}\includegraphics[width=0.16\textwidth]{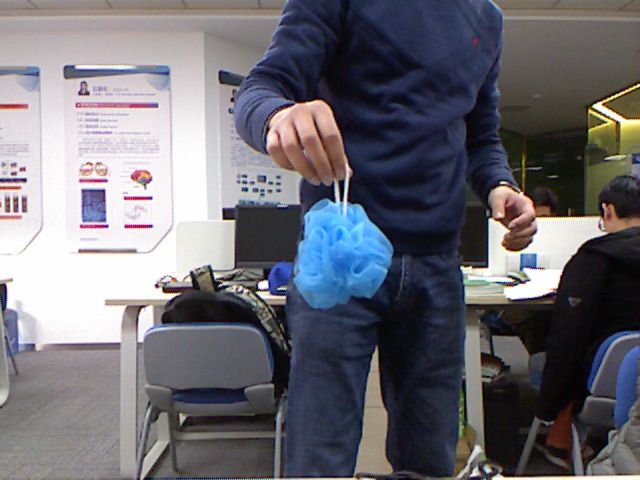}} &        
        {\hspace{-4mm}\includegraphics[width=0.16\textwidth]{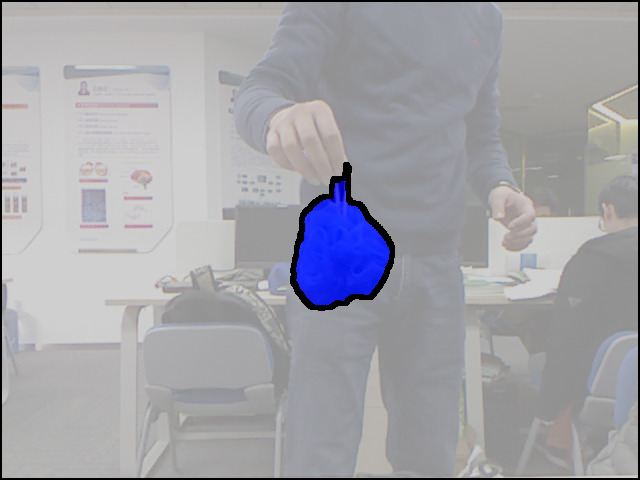}} \vspace{1mm}\\                                                                  
		\includegraphics[width=0.16\textwidth]{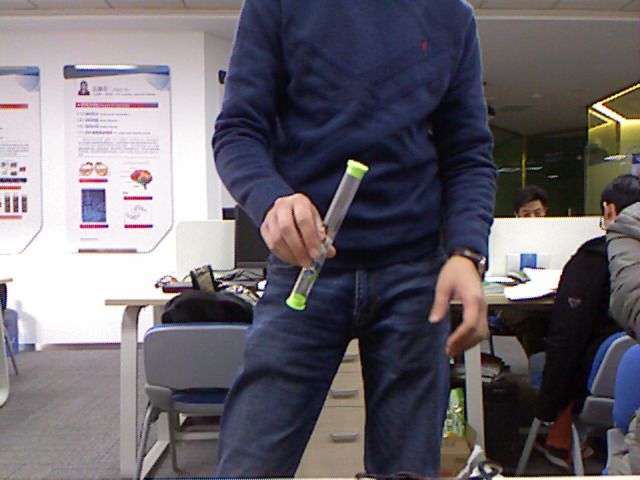} &
        {\hspace{-4mm}\includegraphics[width=0.16\textwidth]{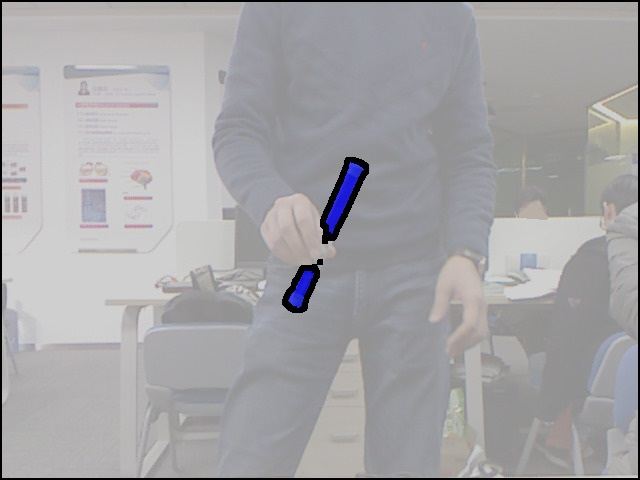}} &
        {\hspace{-4mm}\includegraphics[width=0.16\textwidth]{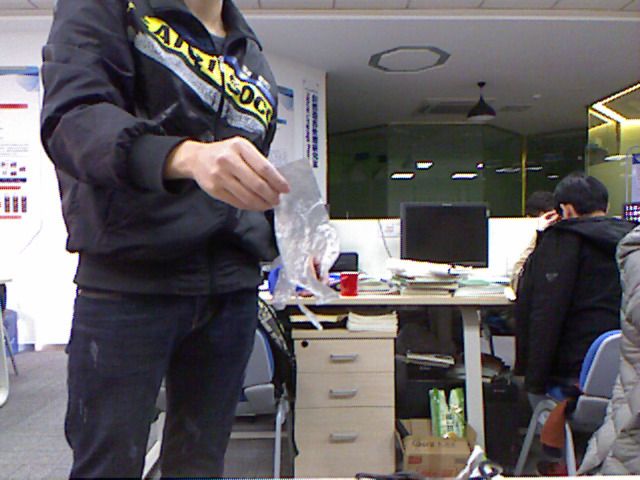}} &
        {\hspace{-4mm}\includegraphics[width=0.16\textwidth]{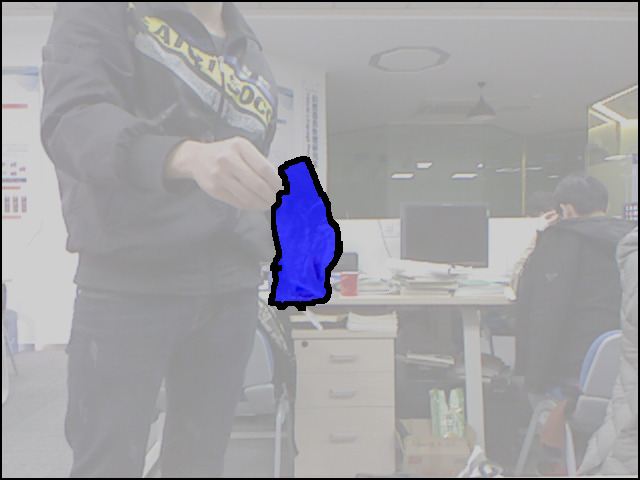}} &
        {\hspace{-4mm}\includegraphics[width=0.16\textwidth]{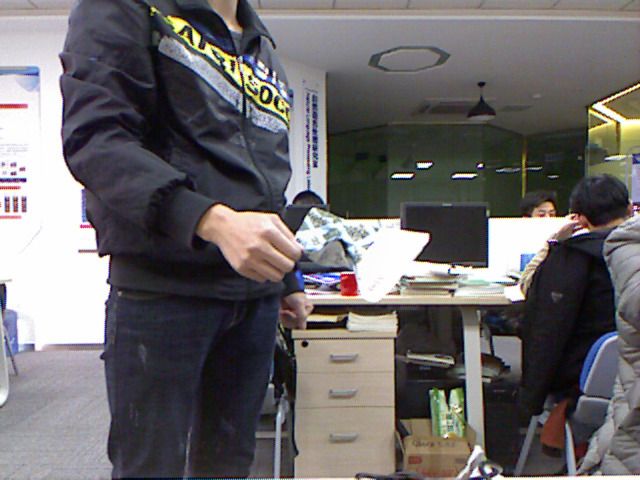}} &        
        {\hspace{-4mm}\includegraphics[width=0.16\textwidth]{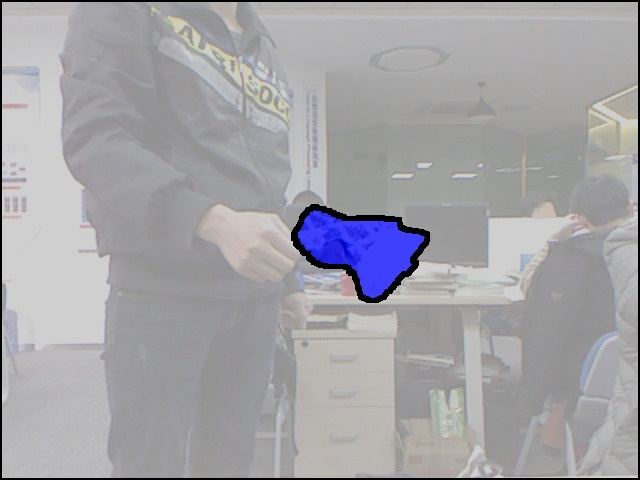}} \vspace{1mm}\\                                                                          
		\includegraphics[width=0.16\textwidth]{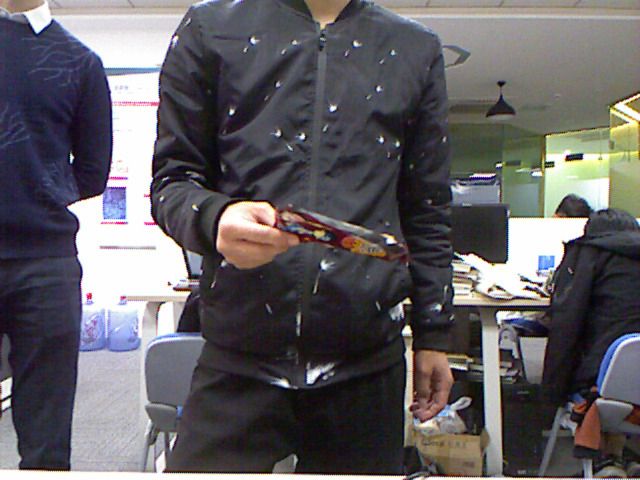} &
        {\hspace{-4mm}\includegraphics[width=0.16\textwidth]{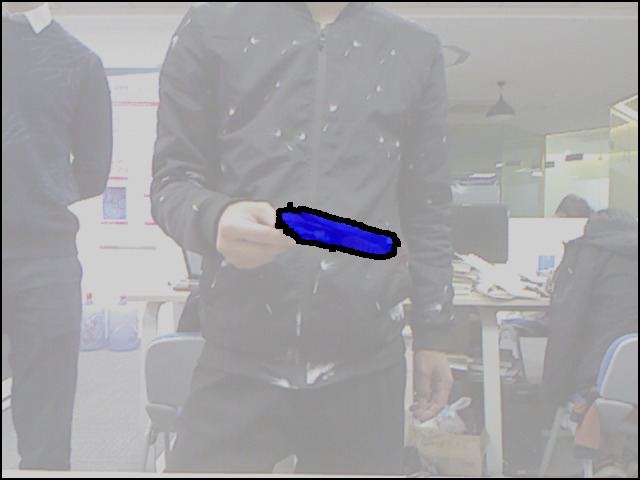}} &
        {\hspace{-4mm}\includegraphics[width=0.16\textwidth]{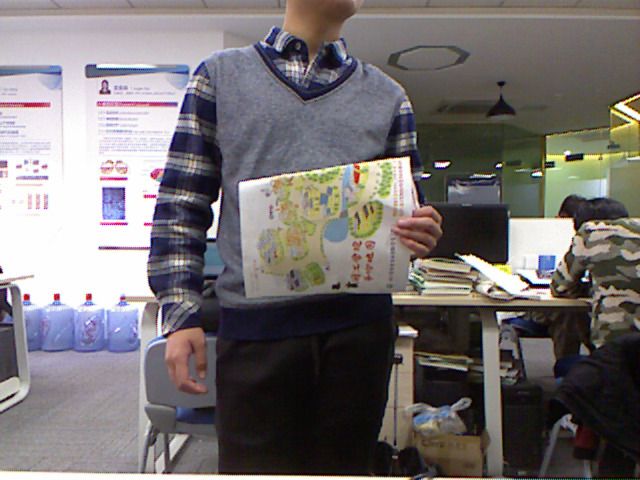}} &
        {\hspace{-4mm}\includegraphics[width=0.16\textwidth]{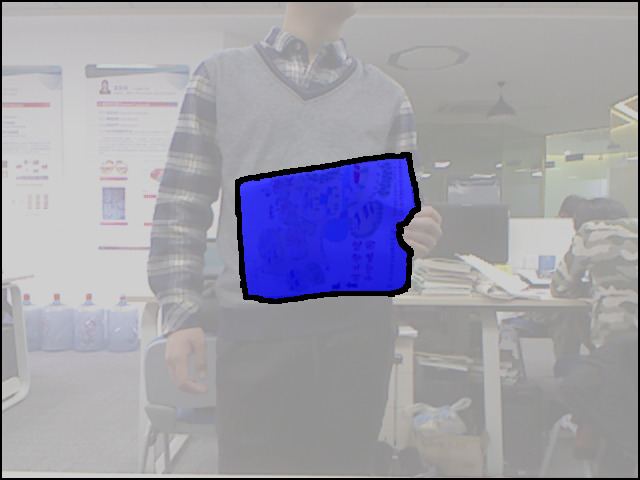}} &
        {\hspace{-4mm}\includegraphics[width=0.16\textwidth]{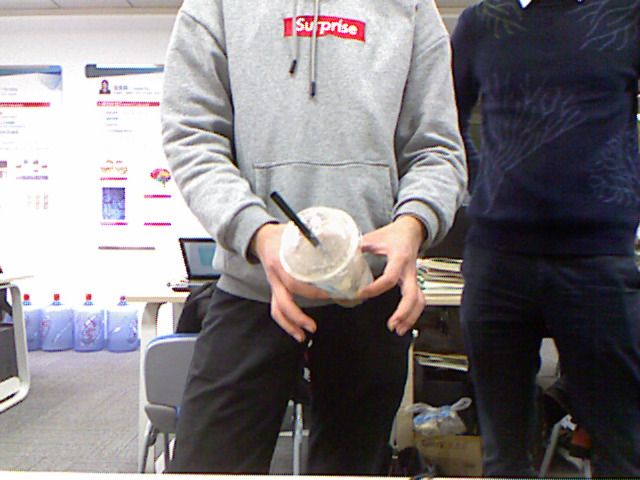}} &        
        {\hspace{-4mm}\includegraphics[width=0.16\textwidth]{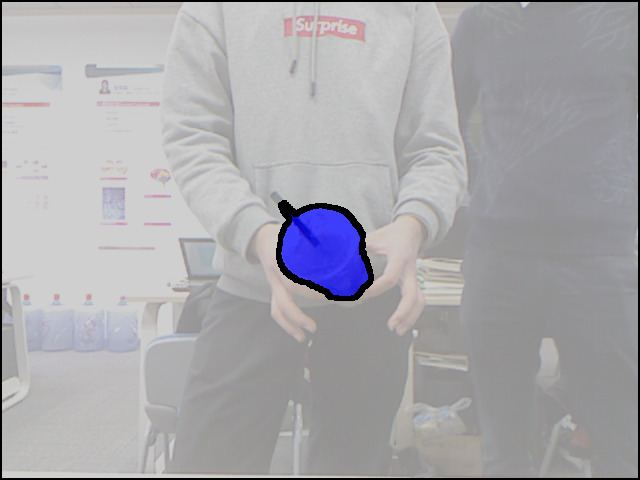}} \vspace{1mm}\\                                                                                 
		\includegraphics[width=0.16\textwidth]{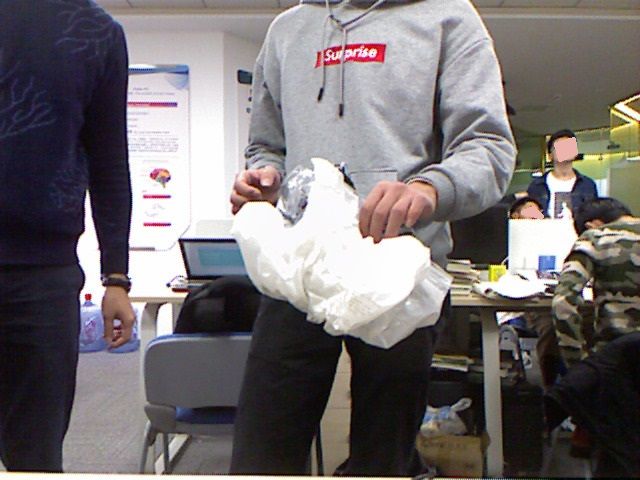} &
        {\hspace{-4mm}\includegraphics[width=0.16\textwidth]{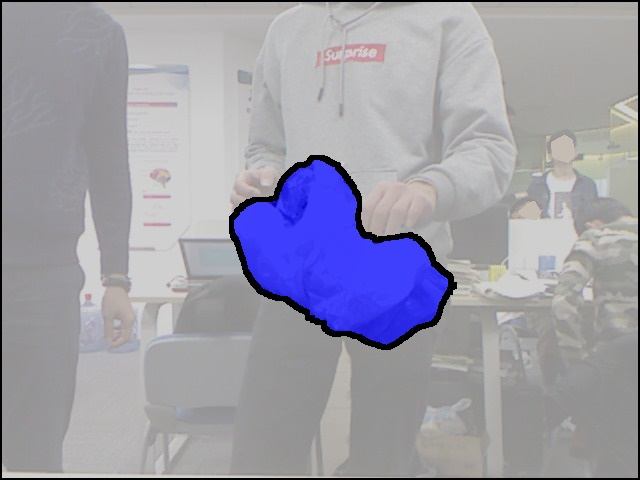}} &
        {\hspace{-4mm}\includegraphics[width=0.16\textwidth]{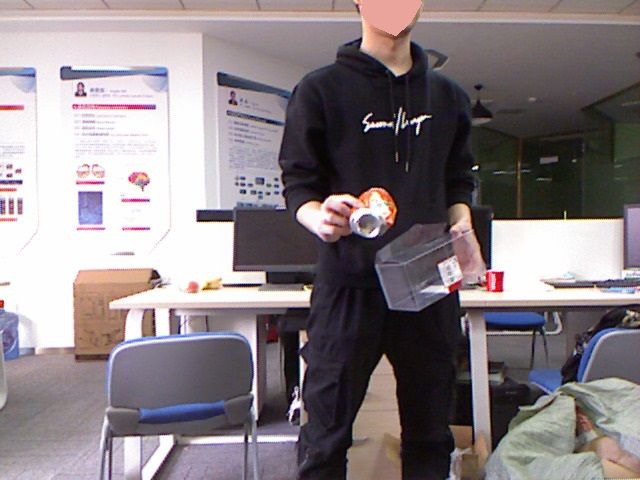}} &
        {\hspace{-4mm}\includegraphics[width=0.16\textwidth]{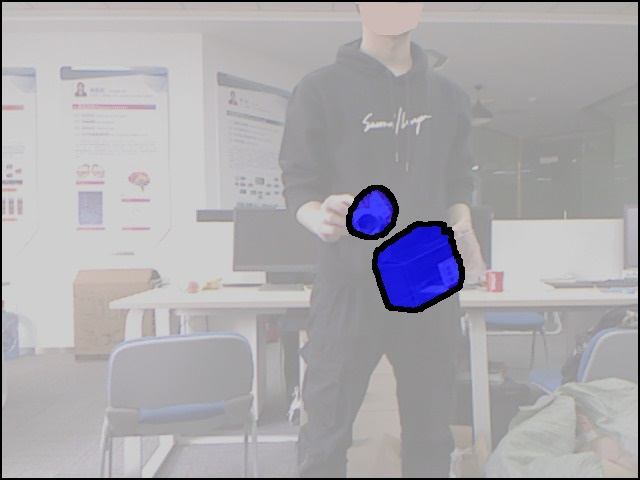}} &
        {\hspace{-4mm}\includegraphics[width=0.16\textwidth]{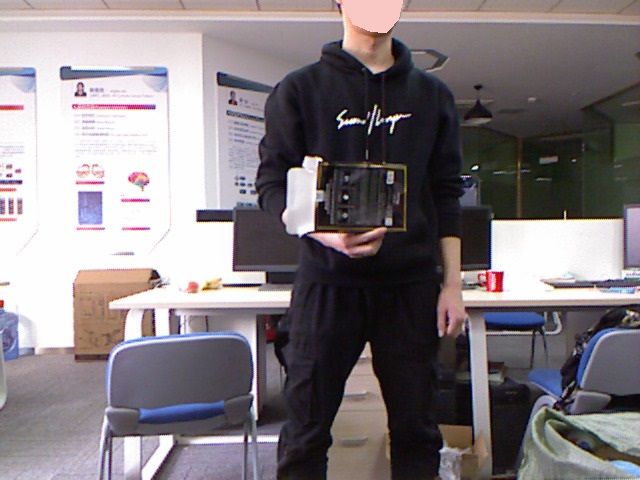}} &        
        {\hspace{-4mm}\includegraphics[width=0.16\textwidth]{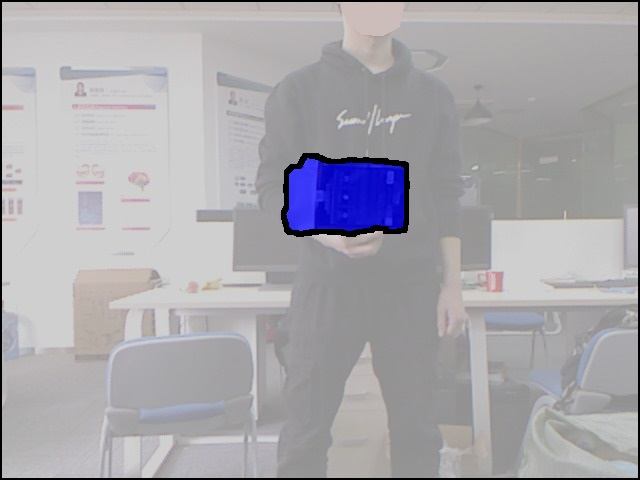}} \vspace{1mm}\\
		\includegraphics[width=0.16\textwidth]{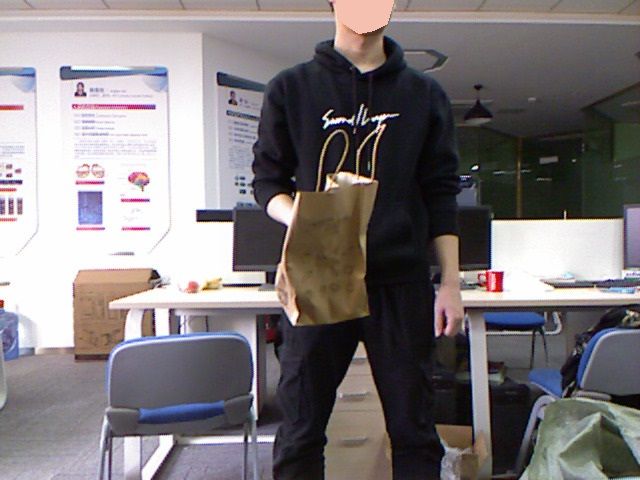} &
        {\hspace{-4mm}\includegraphics[width=0.16\textwidth]{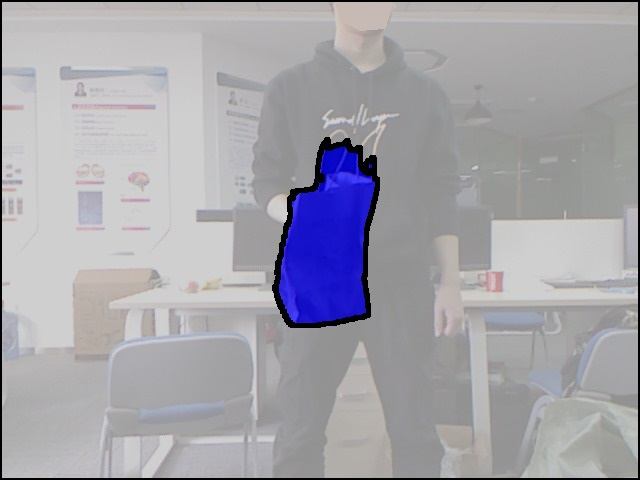}} &
        {\hspace{-4mm}\includegraphics[width=0.16\textwidth]{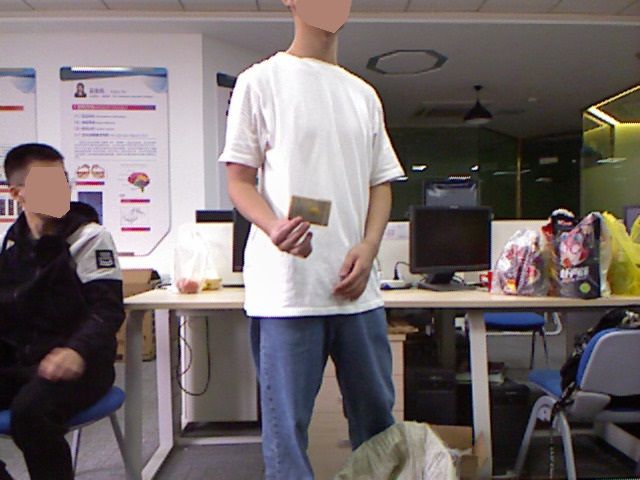}} &
        {\hspace{-4mm}\includegraphics[width=0.16\textwidth]{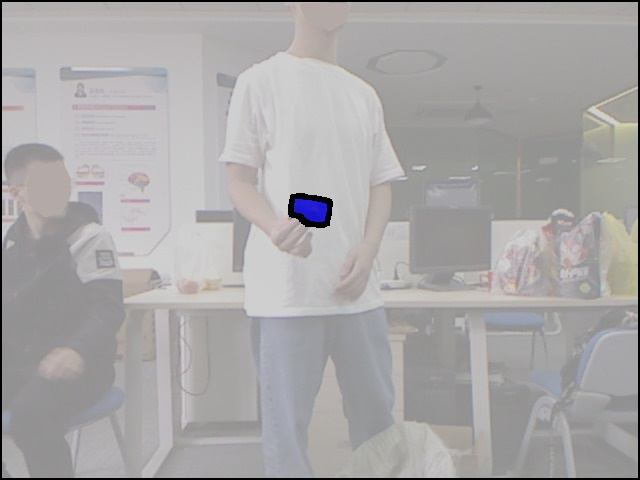}} &
        {\hspace{-4mm}\includegraphics[width=0.16\textwidth]{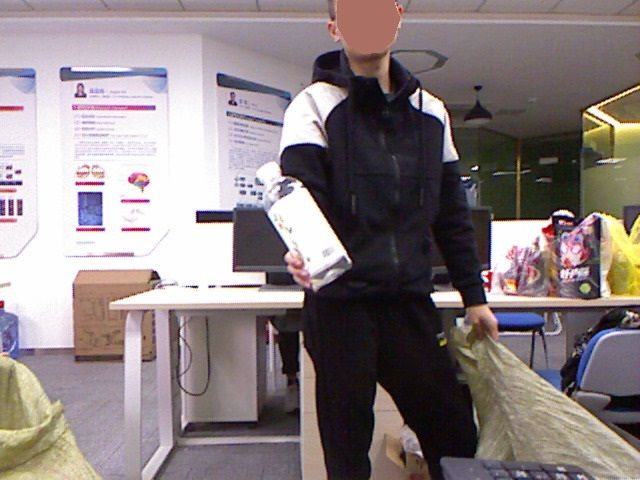}} &        
        {\hspace{-4mm}\includegraphics[width=0.16\textwidth]{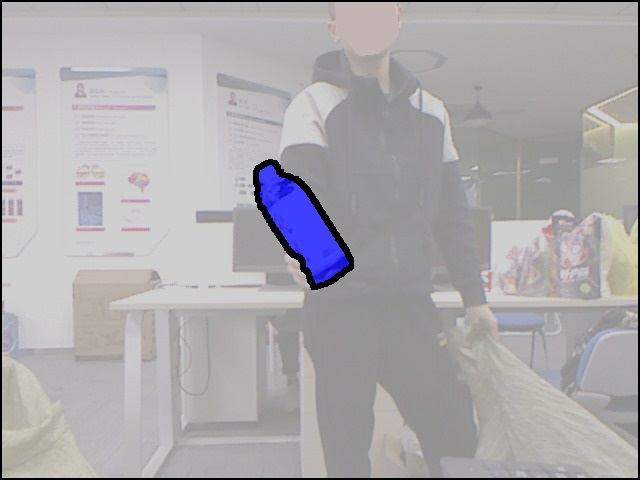}} \vspace{1mm}\\   
		\includegraphics[width=0.16\textwidth]{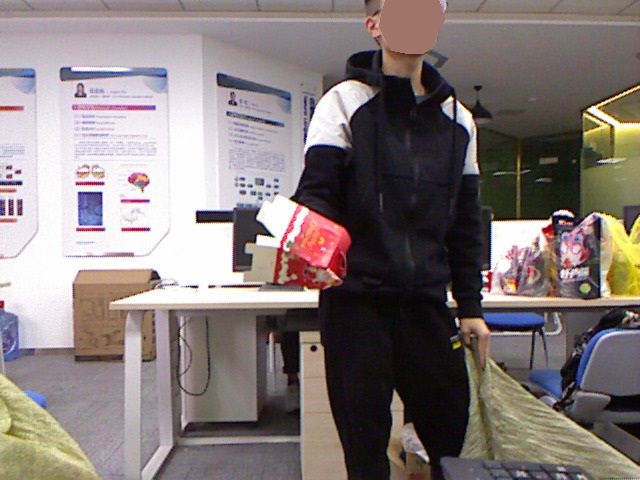} &
        {\hspace{-4mm}\includegraphics[width=0.16\textwidth]{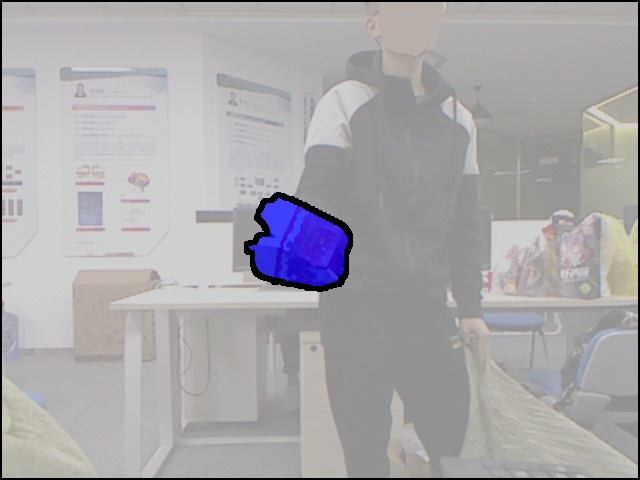}} &
        {\hspace{-4mm}\includegraphics[width=0.16\textwidth]{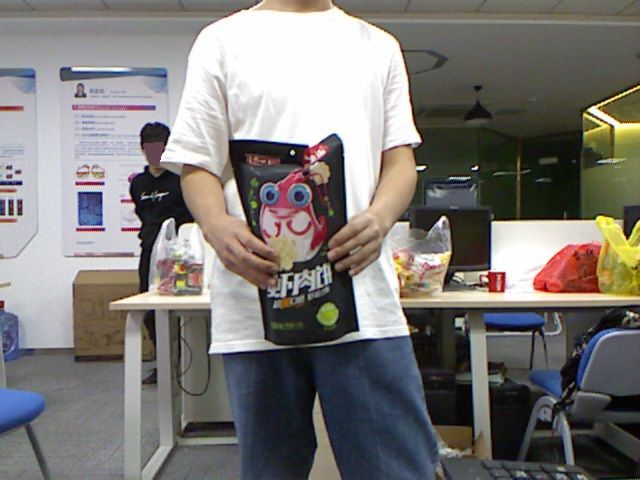}} &
        {\hspace{-4mm}\includegraphics[width=0.16\textwidth]{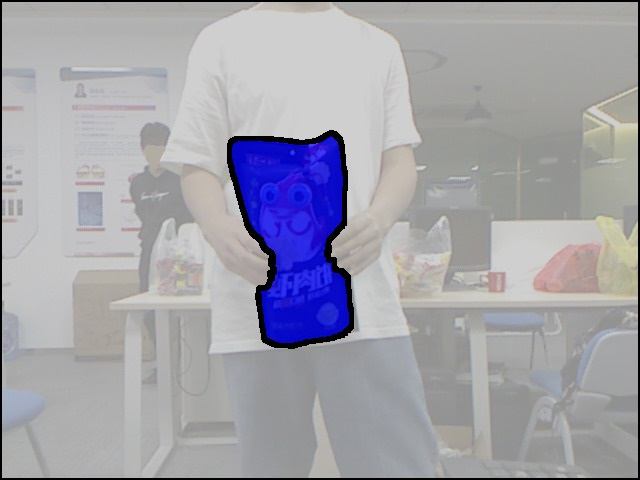}} &
        {\hspace{-4mm}\includegraphics[width=0.16\textwidth]{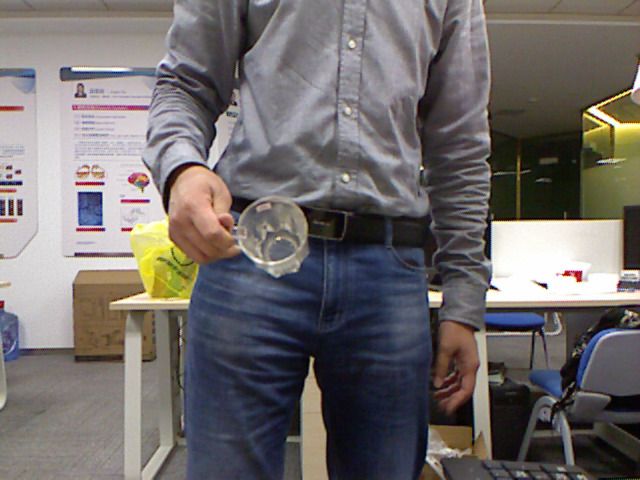}} &        
        {\hspace{-4mm}\includegraphics[width=0.16\textwidth]{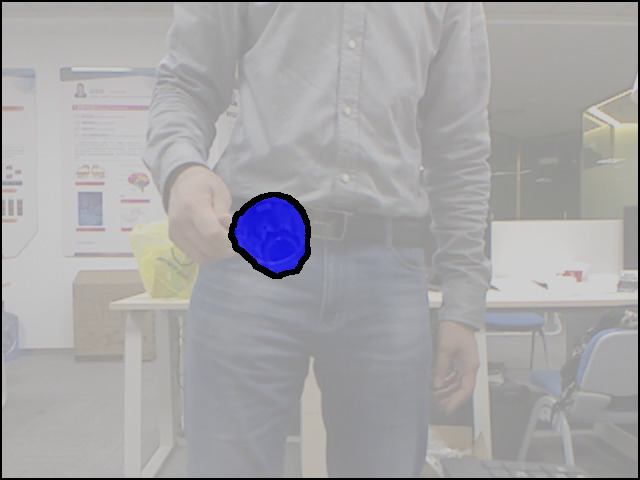}} \vspace{1mm}\\                     
      \end{tabular}
		\caption
         {
			Segmentation results on MJU-Waste (test). Method is DeepLabv3-ML (ResNet-50).
         }
      \label{fig:more-mju}
%\vspace{3mm}
\end{figure}

{In terms of the computational efficiency, we~report a breakdown of the average per-image inference time in Table}~\ref{tab:time}.
{The baseline method corresponds to the scene-level inference only; additional object and pixel level inference incurs extra
computational costs.
These runtime statistics are obtained with an i9 desktop CPU and a single RTX 2080Ti GPU.
Our full model with DeepLabv3 and ResNet-50 runs at approximately $0.8$ second per image.
Specifically, the~computational costs for object-level inference are mainly a result from the object region
proposals and the forward pass of the object region CNN. The~pixel-level inference time, on~the other hand, is~mostly the result
from the iterative mean-field approximation.
It should be noted that the inference times reported here are obtained based on public implementations as mentioned in Section}~\ref{sec:impl},
{without any specific optimization.}

More example results obtained on the test set of MJU-Waste with our full model are shown in Figure~\ref{fig:more-mju}.
{Although the images in MJU-Waste are captured indoors so that the illumination variations are less significant, there
are large variations in the clothing colors and, in~some cases, the~color contrast between the waste objects and the clothes
is small. In~addition, the~orientation of the objects also exhibits large variations. For~example, the~objects can
be held with either one or both hands. During the data collection, we~simply ask the participants to hold objects however they like.
Despite these challenges, our~model is able to reliably recover the fine boundary details in most cases.}

\subsection{Results on the TACO Dataset}
\label{sec:res-taco}

\begin{figure}[H]
	  \centering
      \begin{tabular}{cccccc}
        \scriptsize{Image} & \hspace{-4mm} \scriptsize{Ground-truth} & \hspace{-4mm} \scriptsize{FCN-8s} & \hspace{-4mm} \scriptsize{FCN-8s-ML} & \hspace{-4mm} \scriptsize{DeepLabv3} & \hspace{-4mm} \scriptsize{DeepLabv3-ML} \\
        \includegraphics[width=0.16\textwidth]{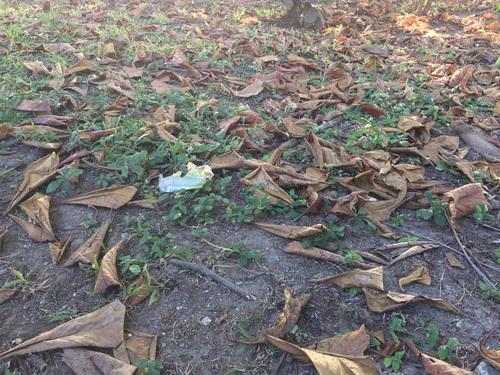} &
        {\hspace{-4mm}\includegraphics[width=0.16\textwidth]{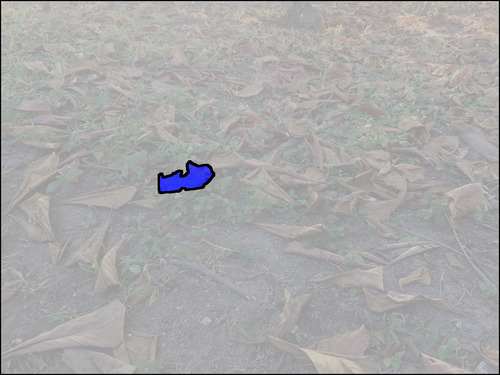}} &
        {\hspace{-4mm}\includegraphics[width=0.16\textwidth]{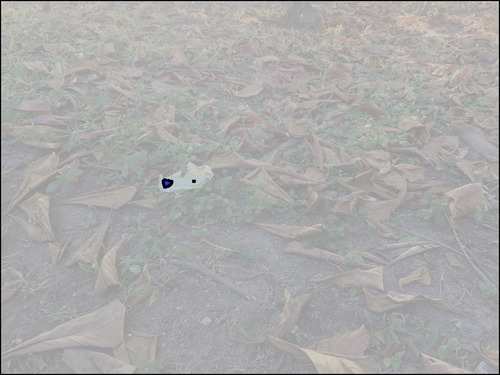}} &
        {\hspace{-4mm}\includegraphics[width=0.16\textwidth]{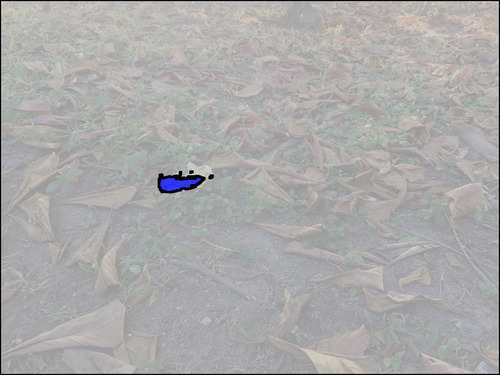}} &
        {\hspace{-4mm}\includegraphics[width=0.16\textwidth]{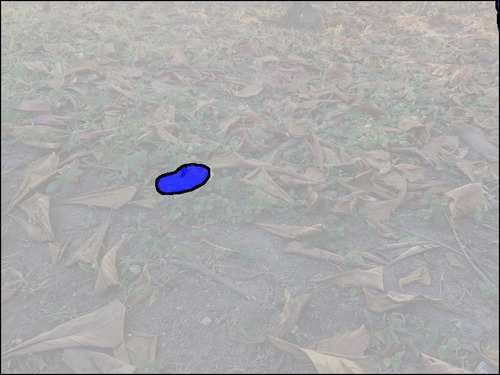}} &        
        {\hspace{-4mm}\includegraphics[width=0.16\textwidth]{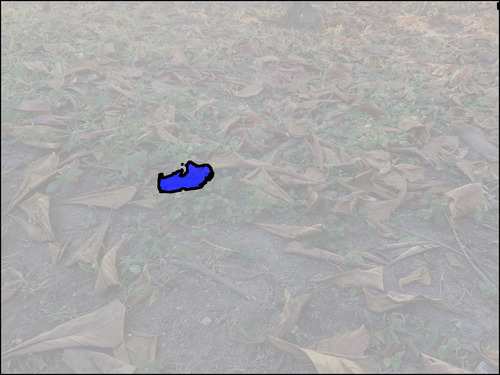}} \vspace{1mm}\\
        \includegraphics[width=0.16\textwidth]{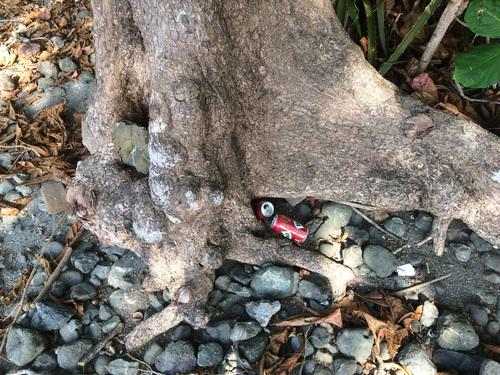} &
        {\hspace{-4mm}\includegraphics[width=0.16\textwidth]{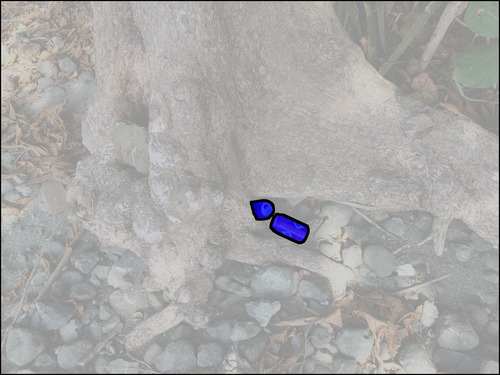}} &
        {\hspace{-4mm}\includegraphics[width=0.16\textwidth]{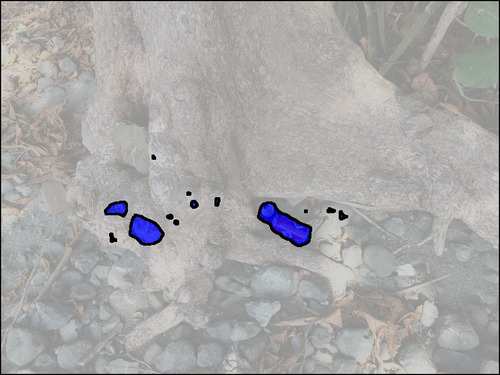}} &
        {\hspace{-4mm}\includegraphics[width=0.16\textwidth]{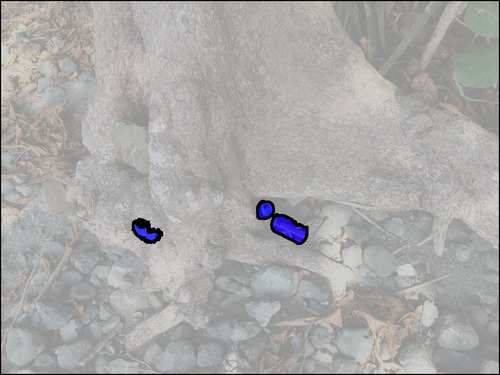}} &
        {\hspace{-4mm}\includegraphics[width=0.16\textwidth]{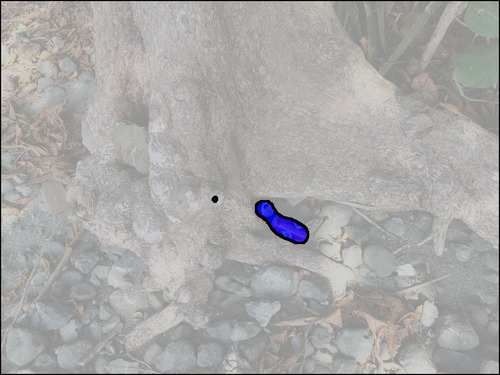}} &        
        {\hspace{-4mm}\includegraphics[width=0.16\textwidth]{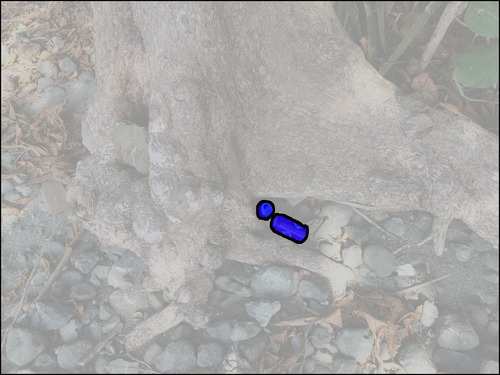}} \vspace{1mm}\\                        
        \includegraphics[width=0.16\textwidth]{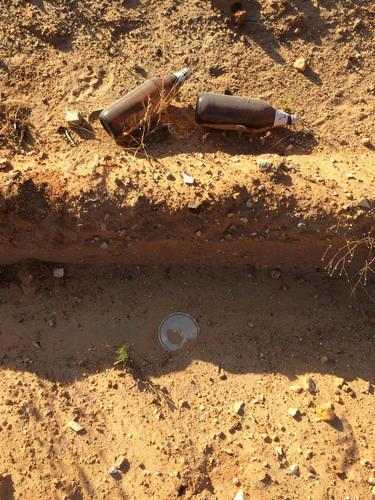} &
        {\hspace{-4mm}\includegraphics[width=0.16\textwidth]{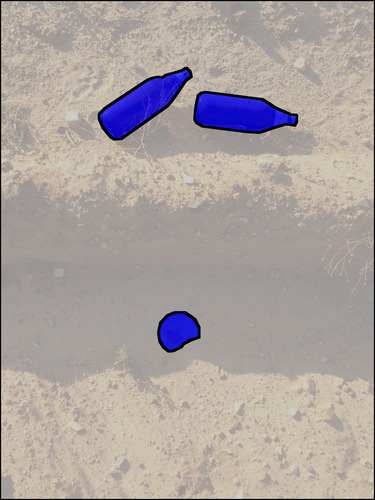}} &
        {\hspace{-4mm}\includegraphics[width=0.16\textwidth]{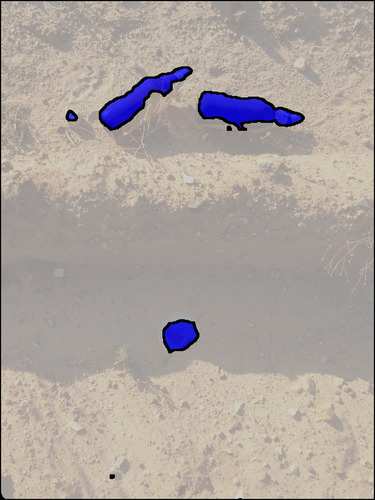}} &
        {\hspace{-4mm}\includegraphics[width=0.16\textwidth]{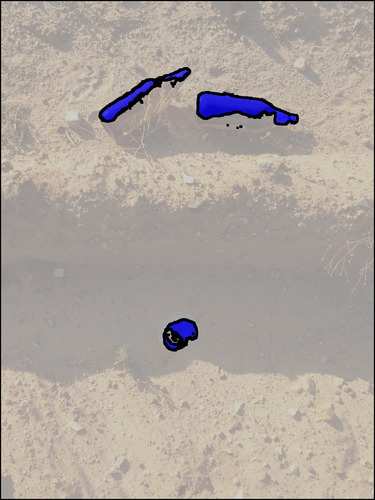}} &
        {\hspace{-4mm}\includegraphics[width=0.16\textwidth]{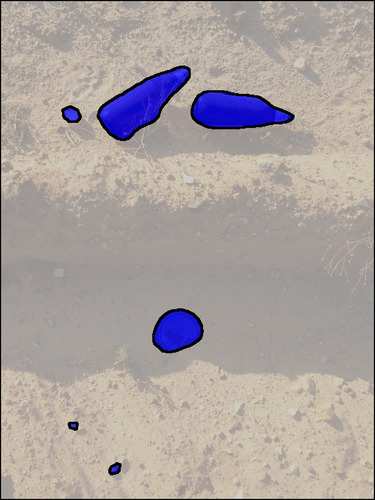}} &        
        {\hspace{-4mm}\includegraphics[width=0.16\textwidth]{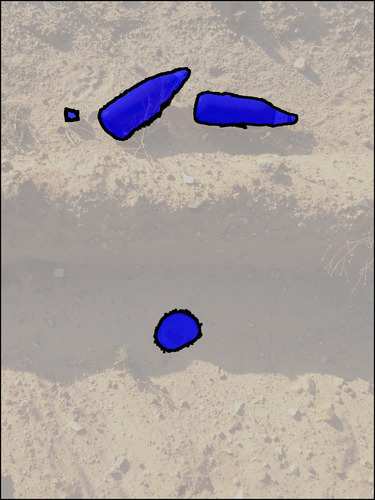}} \vspace{1mm}\\
        \includegraphics[width=0.16\textwidth]{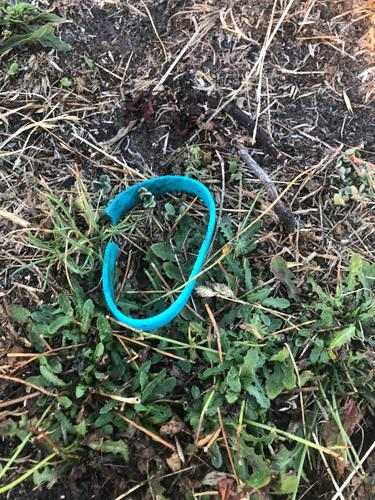} &
        {\hspace{-4mm}\includegraphics[width=0.16\textwidth]{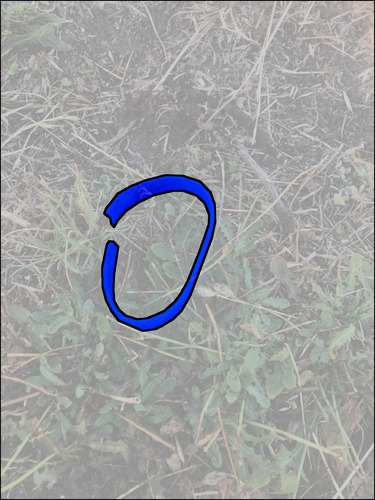}} &
        {\hspace{-4mm}\includegraphics[width=0.16\textwidth]{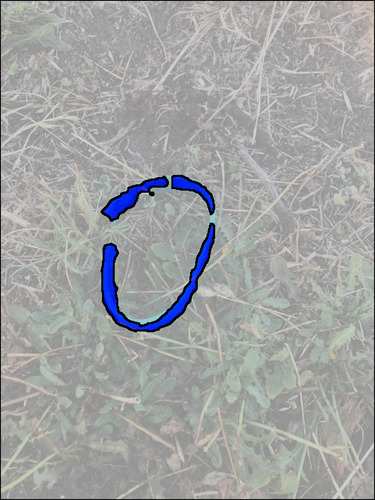}} &
        {\hspace{-4mm}\includegraphics[width=0.16\textwidth]{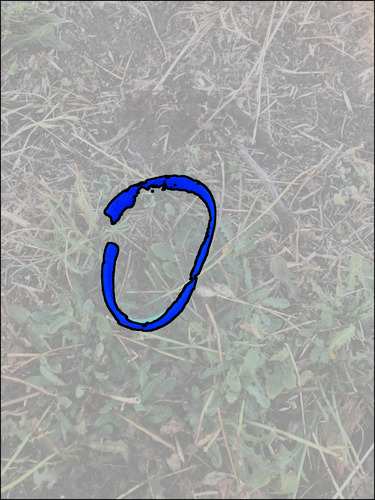}} &
        {\hspace{-4mm}\includegraphics[width=0.16\textwidth]{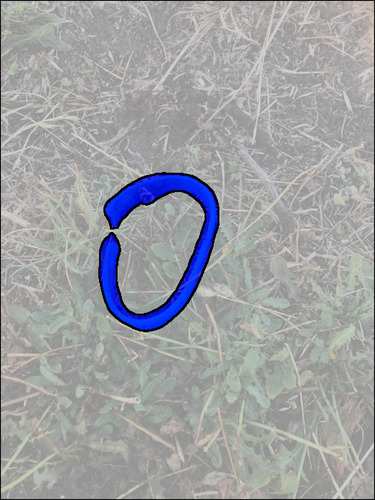}} &        
        {\hspace{-4mm}\includegraphics[width=0.16\textwidth]{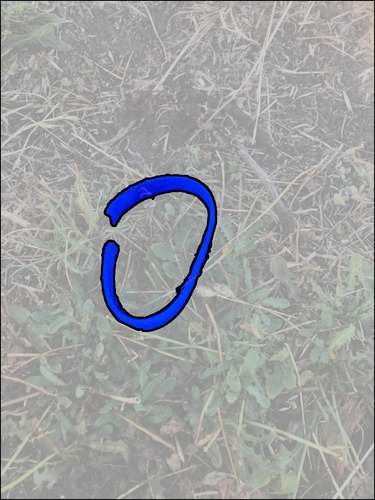}} \vspace{1mm}\\ 
      \end{tabular}
		\caption
         {
			Example segmentation results on the TACO test set. Input images and ground-truth annotations are shown
			in the first two columns. Baseline methods are FCN-8s (VGG-16) and DeepLabv3 (ResNet-101).
			Our proposed methods (FCN-8s-ML and DeepLabv3-ML) more accurately recover object boundaries.
			Best viewed electronically, zoomed in.
         }
      \label{fig:segs-taco}
        % \vspace{3mm}
\end{figure}

\begin{figure}[H]
	  \centering
      \begin{tabular}{cccccc}
        \scriptsize{Image} & \hspace{-4mm} \scriptsize{Prediction} & \hspace{-4mm} \scriptsize{Image} & \hspace{-4mm} \scriptsize{Prediction} & \hspace{-4mm} \scriptsize{Image} & \hspace{-4mm} \scriptsize{Prediction} \\
		\includegraphics[width=0.16\textwidth]{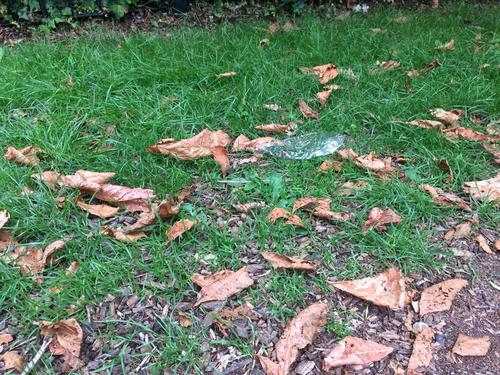} &
        {\hspace{-4mm}\includegraphics[width=0.16\textwidth]{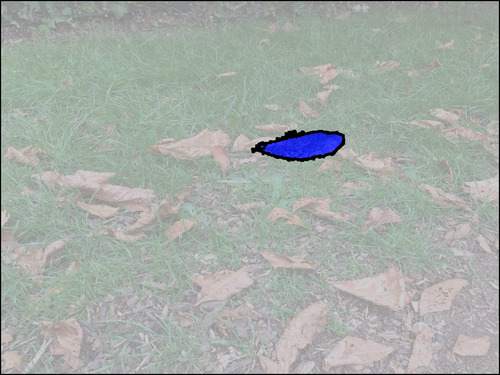}} &
        {\hspace{-4mm}\includegraphics[width=0.16\textwidth]{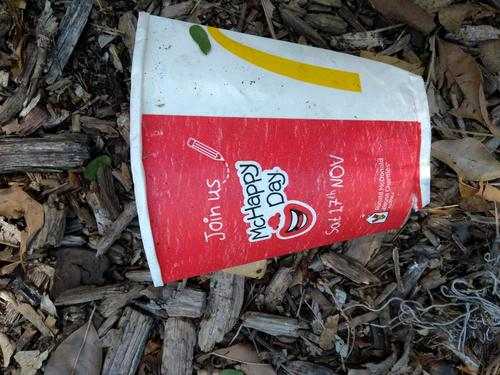}} &
        {\hspace{-4mm}\includegraphics[width=0.16\textwidth]{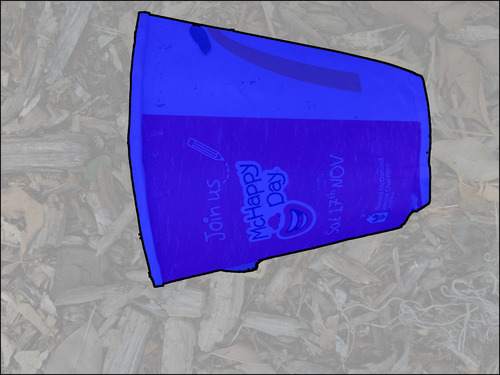}} &
        {\hspace{-4mm}\includegraphics[width=0.16\textwidth]{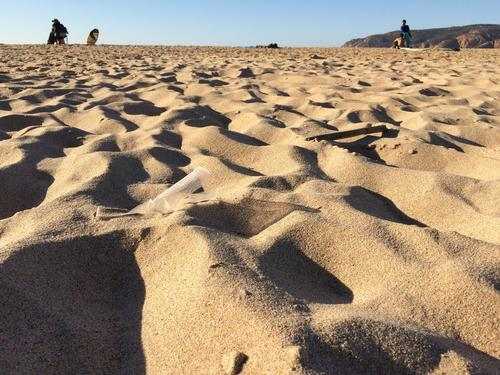}} &        
        {\hspace{-4mm}\includegraphics[width=0.16\textwidth]{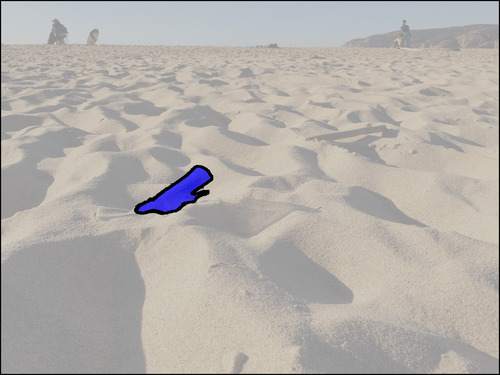}} \vspace{1mm}\\     
		\includegraphics[width=0.16\textwidth]{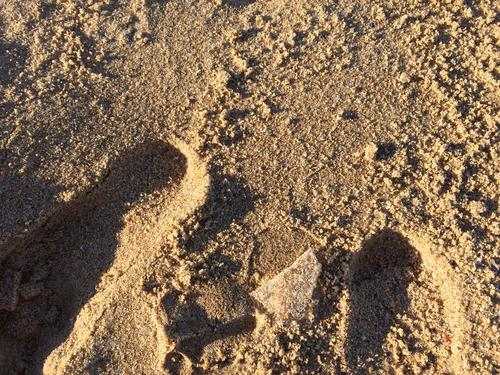} &
        {\hspace{-4mm}\includegraphics[width=0.16\textwidth]{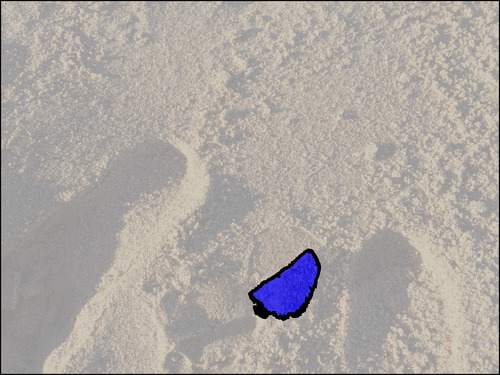}} &
        {\hspace{-4mm}\includegraphics[width=0.16\textwidth]{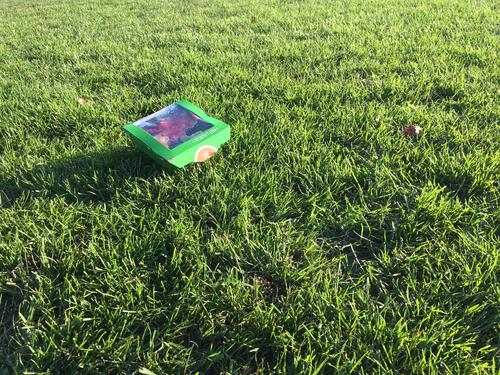}} &
        {\hspace{-4mm}\includegraphics[width=0.16\textwidth]{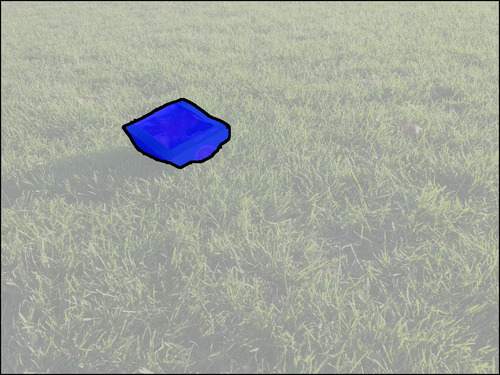}} &
        {\hspace{-4mm}\includegraphics[width=0.16\textwidth]{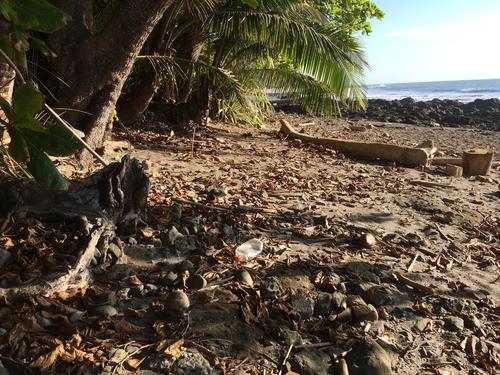}} &        
        {\hspace{-4mm}\includegraphics[width=0.16\textwidth]{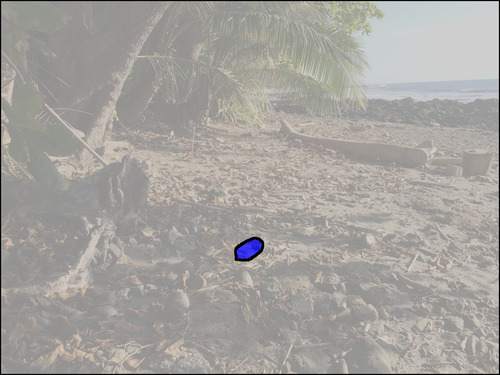}} \vspace{1mm}\\ 
		\includegraphics[width=0.16\textwidth]{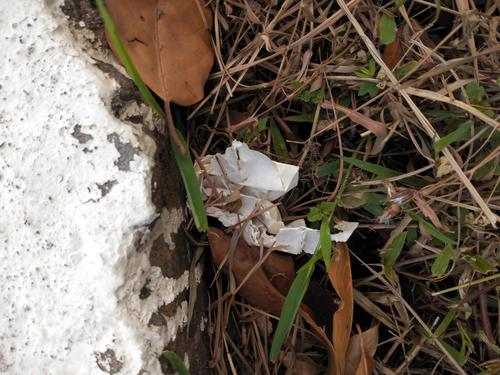} &
        {\hspace{-4mm}\includegraphics[width=0.16\textwidth]{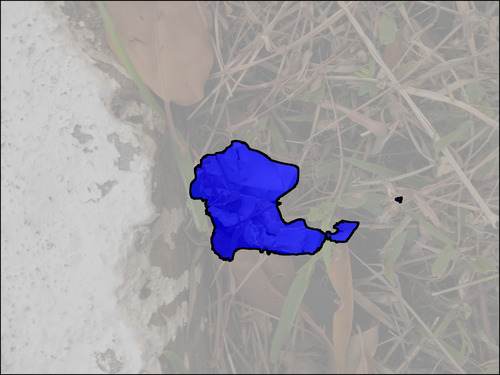}} &
        {\hspace{-4mm}\includegraphics[width=0.16\textwidth]{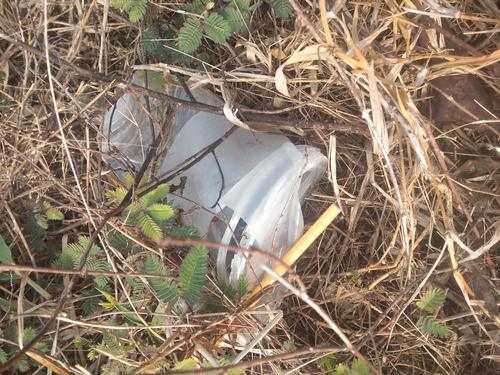}} &
        {\hspace{-4mm}\includegraphics[width=0.16\textwidth]{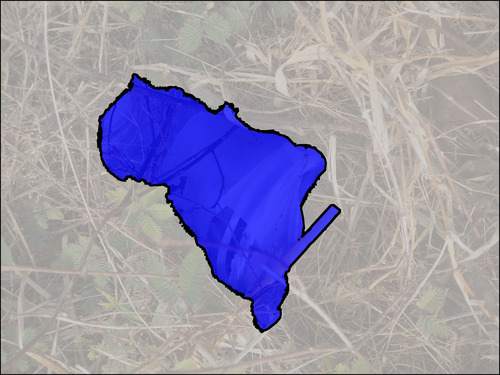}} &
        {\hspace{-4mm}\includegraphics[width=0.16\textwidth]{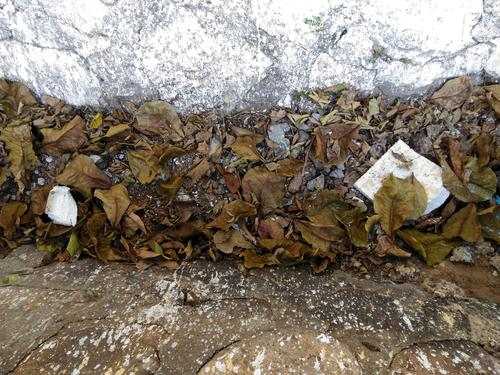}} &        
        {\hspace{-4mm}\includegraphics[width=0.16\textwidth]{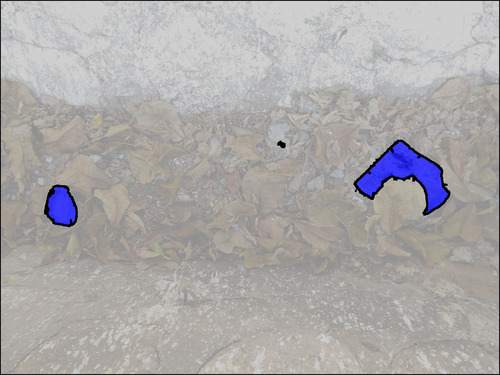}} \vspace{1mm}\\
		\includegraphics[width=0.16\textwidth]{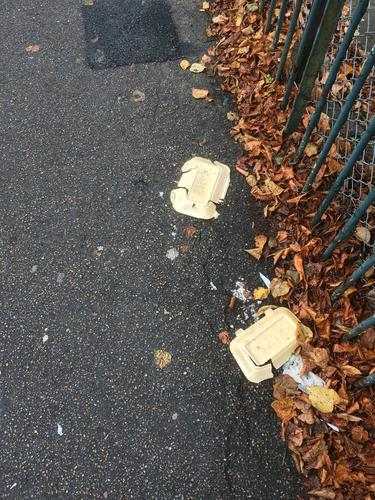} &
        {\hspace{-4mm}\includegraphics[width=0.16\textwidth]{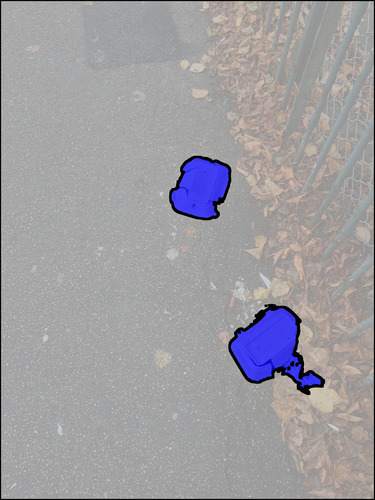}} &
        {\hspace{-4mm}\includegraphics[width=0.16\textwidth]{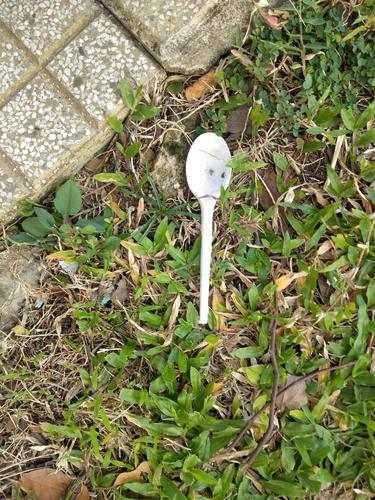}} &
        {\hspace{-4mm}\includegraphics[width=0.16\textwidth]{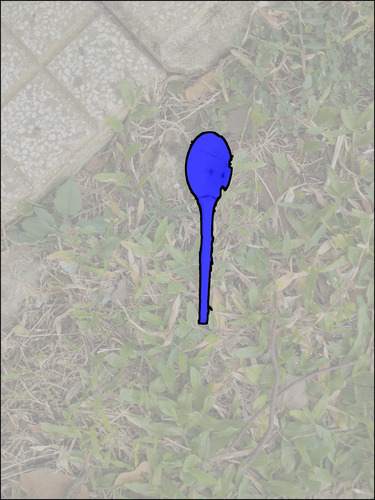}} &
        {\hspace{-4mm}\includegraphics[width=0.16\textwidth]{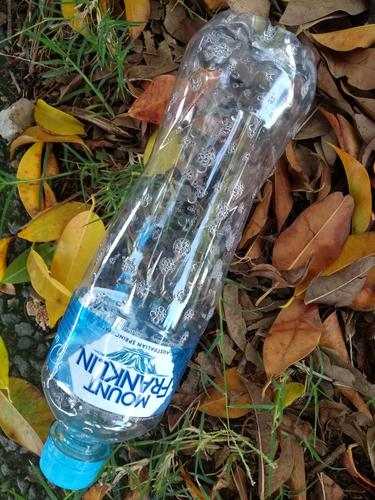}} &        
        {\hspace{-4mm}\includegraphics[width=0.16\textwidth]{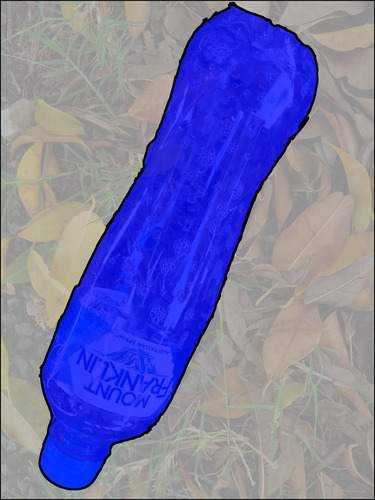}} \vspace{1mm}\\
		\includegraphics[width=0.16\textwidth]{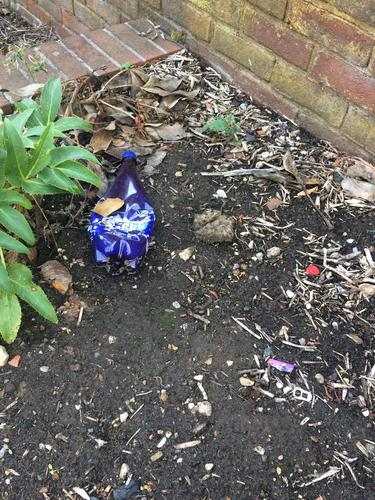} &
        {\hspace{-4mm}\includegraphics[width=0.16\textwidth]{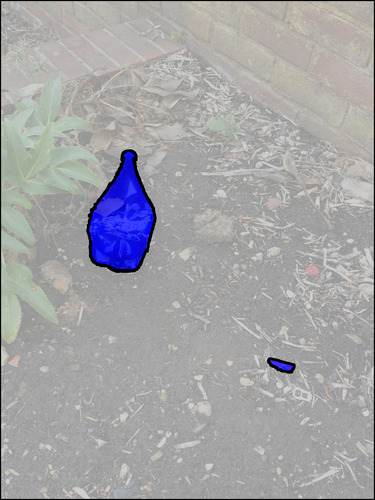}} &
        {\hspace{-4mm}\includegraphics[width=0.16\textwidth]{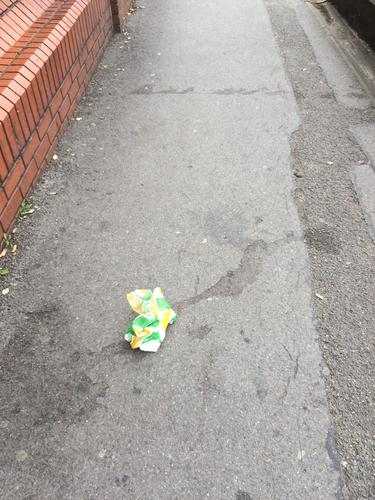}} &
        {\hspace{-4mm}\includegraphics[width=0.16\textwidth]{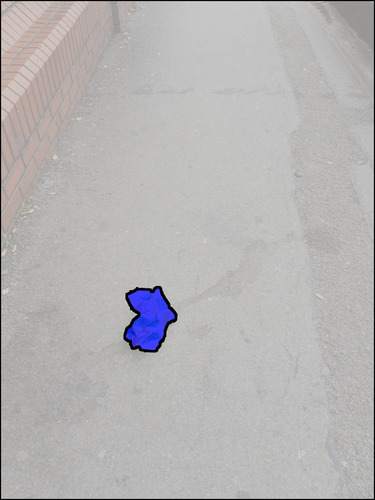}} &
        {\hspace{-4mm}\includegraphics[width=0.16\textwidth]{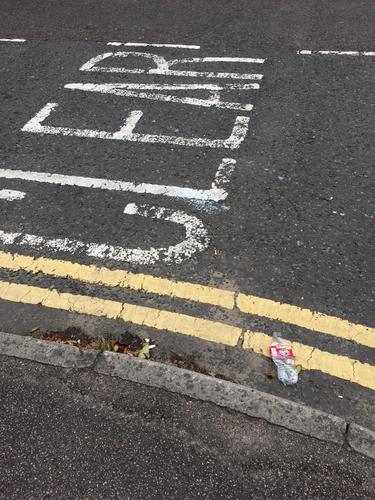}} &        
        {\hspace{-4mm}\includegraphics[width=0.16\textwidth]{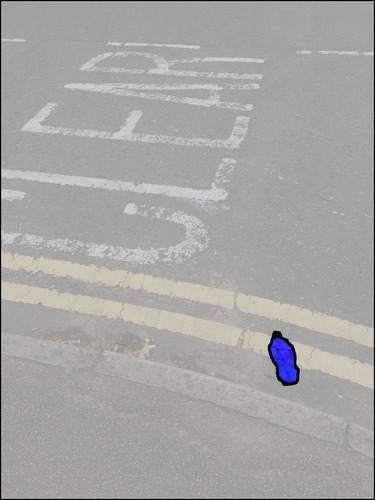}} \vspace{1mm}\\ 
		\includegraphics[width=0.16\textwidth]{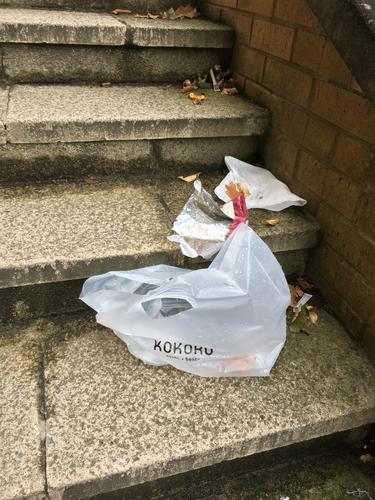} &
        {\hspace{-4mm}\includegraphics[width=0.16\textwidth]{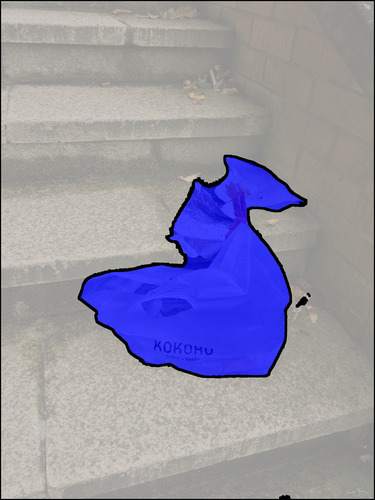}} &
        {\hspace{-4mm}\includegraphics[width=0.16\textwidth]{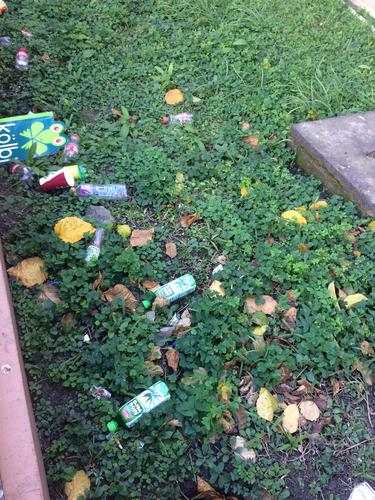}} &
        {\hspace{-4mm}\includegraphics[width=0.16\textwidth]{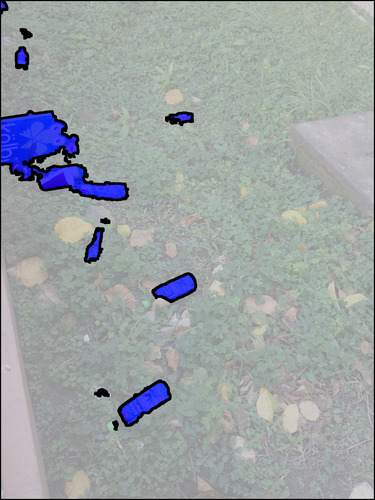}} &
        {\hspace{-4mm}\includegraphics[width=0.16\textwidth]{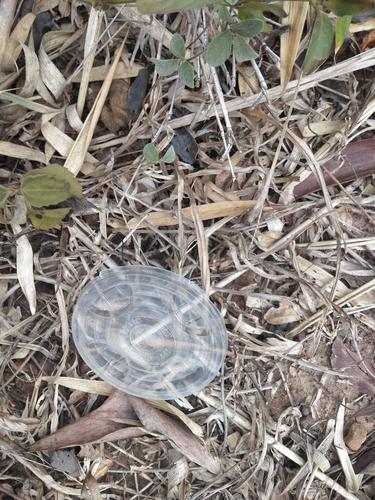}} &        
        {\hspace{-4mm}\includegraphics[width=0.16\textwidth]{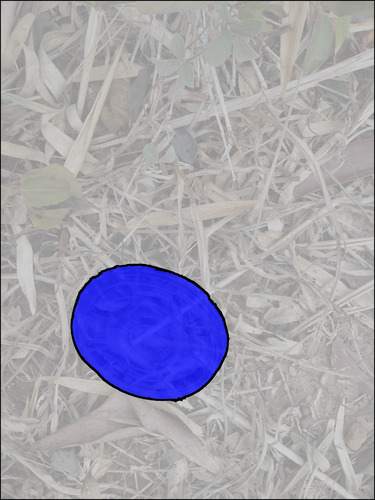}} \vspace{1mm}\\
		\includegraphics[width=0.16\textwidth]{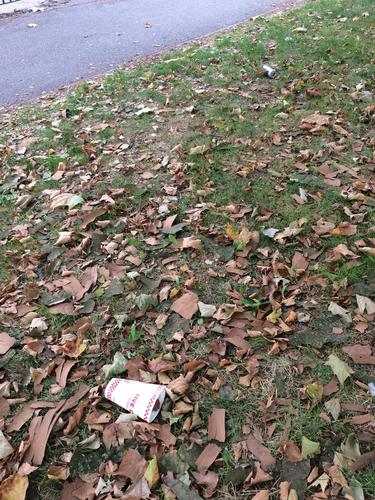} &
        {\hspace{-4mm}\includegraphics[width=0.16\textwidth]{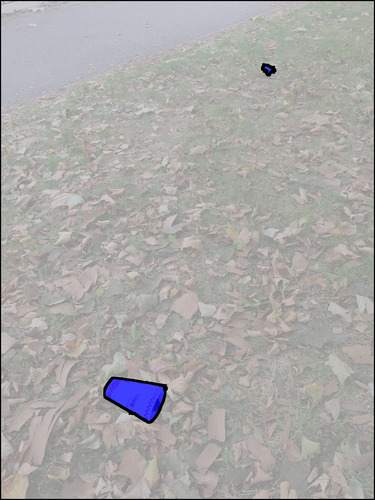}} &
        {\hspace{-4mm}\includegraphics[width=0.16\textwidth]{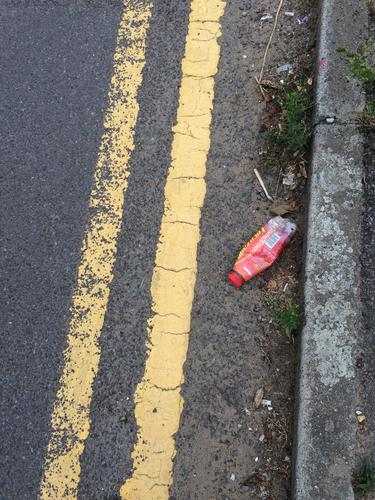}} &
        {\hspace{-4mm}\includegraphics[width=0.16\textwidth]{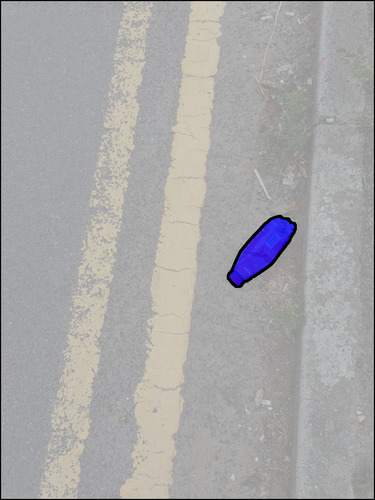}} &
        {\hspace{-4mm}\includegraphics[width=0.16\textwidth]{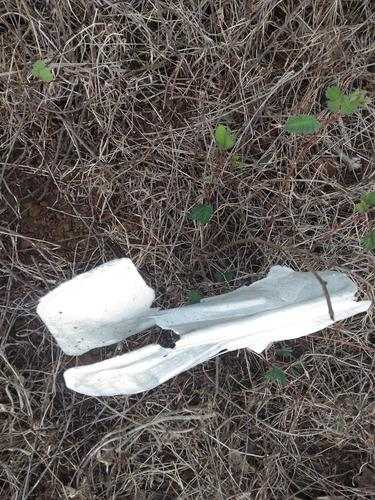}} &        
        {\hspace{-4mm}\includegraphics[width=0.16\textwidth]{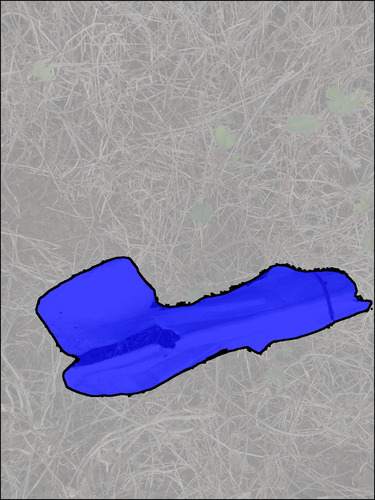}} \vspace{1mm}\\                                                                 
      \end{tabular}
		\caption
         {
			Segmentation results on TACO (test). Method is DeepLabv3-ML (ResNet-101).
         }
      \label{fig:more-taco}
\end{figure}

We additionally evaluate the performance of our method on the TACO dataset.
{TACO contains color images only, so~we exclude Equation}~(\ref{eqn:depth-pairwise}) {for training
and evaluating models on this dataset.}
This dataset presents a unique challenge for localizing waste objects ``in-the-wild''. In~general,
TACO is different to MJU-Waste in two important aspects. Firstly, multiple waste objects
with extreme scale variation are more common (see Figures~\ref{fig:segs-taco} and~\ref{fig:more-taco}
for examples). Secondly, unlike MJU-Waste the backgrounds are diverse, such~as in road, grassland and beach scenes.
Quantitative results obtained on the TACO test set are summarized in Table~\ref{tab:tacoexp}.
Specifically, we~compare our multi-level model against two baselines: FCN-8s~\cite{long2015fully} and
DeepLabv3~\cite{chen2017rethinking}. Again, in~both cases our multi-level model is able to improve the baseline
performance by a clear margin. Qualitative comparisons of the segmentation results are presented in Figure~\ref{fig:segs-taco}.
It is clear that our multi-level method is able to more closely follow object boundaries.
More example segmentation results are presented in Figure~\ref{fig:more-taco}.
{We note that the changes in illumination and orientation are generally greater on TACO than on MJU-Waste,
due to the fact that there are many outdoor images. Particularly, in~some beach images it is very challenging to
spot waste objects due to the poor illumination and the weak color contrast. Furthermore, object scale and orientation
vary greatly as a result of different camera perspectives. Again, our~model is able to detect and segment waste
objects with high accuracy in most images, demonstrating the efficacy of the proposed method.}

\begin{table}[t!]
	\centering
	  \caption
         {
            Performance comparisons on the test set of TACO. For~each method, we~report the IoU for waste objects (IoU),
            mean IoU (mIoU), pixel Precision for waste objects (Prec) and mean pixel precision (Mean). See~Section~\ref{sec:res-taco} for details.
         }
\label{tab:tacoexp}
		\begin{tabular}{ll c c c c}
	         \toprule
             \makecell[bl]{\textbf{Dataset:} \\\textbf{TACO (test)}} & \makecell[bl]{\textbf{Backbone}} & \textbf{IoU} & \textbf{mIoU} & \textbf{Prec} & \textbf{Mean} \\
             \midrule
%             \multicolumn{10}{l}{} \\[-0.9em] % Spacer
             \multicolumn{6}{l}{Baseline Approaches} \\[.1em]
             \midrule
%             \multicolumn{10}{l}{} \\[-0.9em] % Spacer
             FCN-8s~\cite{long2015fully} & \scriptsize VGG-16 & 70.43 & 84.31 & 85.50 & 92.21 \\          
             DeepLabv3~\cite{chen2017rethinking} & \scriptsize ResNet-101 & \bf{83.02} & \bf{90.99} & \bf{88.37} & \bf{94.00} \\
%             \multicolumn{10}{l}{} \\[-0.9em] % Spacer             
            \hline
            &&&&&\\[-2ex]
 %            \multicolumn{10}{l}{} \\[-0.9em] % Spacer             
             \multicolumn{6}{l}{Proposed Multi-Level (ML) Model} \\[.1em]
            \midrule
%             \multicolumn{10}{l}{} \\[-0.9em] % Spacer     
             FCN-8s-ML & \scriptsize VGG-16 & 74.21 & 86.35 & 90.36 & 94.65 \\
              & & \scriptsize(+3.78) & \scriptsize(+2.04) & \scriptsize(+4.86) & \scriptsize(+2.44) \\
             DeepLabv3-ML & \scriptsize ResNet-101 & \bf{86.58} & \bf{92.90} & \bf{92.52} & \bf{96.07} \\             
              & & \scriptsize(+3.56) & \scriptsize(+1.91) & \scriptsize(+4.15) & \scriptsize(+2.07) \\                          [.5ex]
\noalign{\hrule height 1.0pt}
         \end{tabular}
   %      \vspace{3mm}       
\end{table}

\section{Conclusion}
\label{sec:conclusion}

We presented a multi-level approach to waste object localization. Specifically, our~method integrates the appearance
and the depth information from three levels of spatial granularity: (1) A scene-level segmentation network
captures the long-range spatial contexts and produces an initial coarse segmentation. (2) Based on the coarse segmentation,
we select a few potential object regions and then perform object-level segmentation. (3) The scene and object
level results are then integrated into a pixel-level fully connected conditional random field to produce a coherent
final localization. The~superiority of our method is validated on two public datasets for waste object segmentation.
As~part of our work, we~collected the MJU-Waste dataset that is made publicly available
to facilitate future research in this area. We~hope that our method could serve as a modest attempt to induce further exploration
into vision-based perception of waste objects in complex real-world scenarios. {For example, possible future work
may explore the training of robust segmentation models that work on multiple datasets with large
object appearance and camera perspective variations.}

\vspace{5mm}
%%%%%%%%%%%%%%%%%%%%%%%%%%%%%%%%%%%%%%%%%%
\footnotesize{\noindent \textbf{Acknowledgments.}{~The authors thank Xuming He for helpful discussions, and~the anonymous reviewers for their invaluable comments.
The authors also thank students from the 2016 Class of Software Engineering, the 2019 Class of Software Engineering, and~the 2016 Class of Computer Science
at Minjiang University for their help in data collection. Finally, the~authors would like to acknowledge NVIDIA Corporation for the generous GPU donation.}}

\vspace{1mm}
%%%%%%%%%%%%%%%%%%%%%%%%%%%%%%%%%%%%%%%%%%
\footnotesize{\noindent \textbf{Funding.}{~Project sponsored by NSFC (61703195), Fujian NSF (2019J01756), Guangdong NSF (2019A1515011045),
The Education Department of Fujian Province (through the Distinguished Young Scholars Program of Fujian Universities, and JK2017039),
Fuzhou Technology Planning Program (2018-G-96, 2018-G-98), Minjiang University (MJY19021, MJY19022), and
The Key Laboratory of Cognitive Computing and Intelligent Information Processing at Wuyi University (KLCCIIP2019202).}}

%%%%%%%%%%%%%%%%%%%%%%%%%%%%%%%%%%%%%%%%%%

{\small
\bibliographystyle{ieee}
\bibliography{string,refs}
}

\end{document}